\pdfoutput=1 
\documentclass[11pt]{article}
\usepackage[preprint]{acl}
\newcommand{\ours}{\textsc{ArcANE}}

\usepackage{times}
\usepackage{latexsym}

\usepackage[T1]{fontenc}
\usepackage[utf8]{inputenc}
\usepackage{microtype}

\usepackage{inconsolata}

\usepackage{graphicx}

\usepackage{enumitem}

\usepackage{amsmath}
\usepackage{amssymb}
\usepackage{multicol}

\usepackage{pifont}
\newcommand{\cmark}{\ding{51}}
\newcommand{\xmark}{\ding{55}}
  
\usepackage[table]{xcolor}  

\usepackage{adjustbox}
\usepackage{makecell}
\usepackage{booktabs}
\usepackage{multirow}
\usepackage{dcolumn} 
\usepackage{threeparttable}
\usepackage{tabularx}
\newcolumntype{d}{D{.}{.}{-1}}
\newcolumntype{Y}{>{\centering\arraybackslash}X}

\usepackage{listings}
\lstdefinestyle{mystyle}{
  backgroundcolor=\color{gray!10},
  basicstyle=\ttfamily\small,
  frame=single, framesep=6pt, framerule=0.5pt, rulecolor=\color{black!30},
  breaklines=true, breakatwhitespace=true, breakindent=0pt,
  showstringspaces=false, columns=flexible, captionpos=b,
  numberstyle=\tiny\color{gray},
  xleftmargin=10pt, xrightmargin=10pt, tabsize=4, extendedchars=true
}
\title{\ours{}: Do Role-Playing Language Agents Stay in Character at the Right Time?}

\newcommand{\authorspacing}{\hspace{1.5mm}}

\author{
  Woojung Song$^*$ \authorspacing Nalim Kim$^*$ \authorspacing Sangjun Song \authorspacing Chaewon Heo \authorspacing Jongwon Lim \authorspacing Yohan Jo$^{\dag}$
  \\
  Graduate School of Data Science, Seoul National University\\
  \texttt{\{opusdeisong, kimnalim, yohan.jo\}@snu.ac.kr}
}

\begin{document}
\maketitle

\def\thefootnote{\fnsymbol{footnote}}
\footnotetext[1]{Equal contribution.}
\footnotetext[2]{Corresponding author.}
\def\thefootnote{\arabic{footnote}}

\begin{abstract}
Role-playing language agents (RPLAs) simulate specific characters and personas across applications such as entertainment, companionship, interactive storytelling, and education. Faithful role-play requires more than producing plausible, in-character responses: as a character’s values and behavior change over a narrative, an RPLA should reflect the character’s state at the relevant stage. However, existing benchmarks largely treat characters as fixed personas or test only what they know at a given point in the narrative. 
We introduce \ours{} (Arc-Aware Narrative Evaluation), a benchmark for evaluating whether an RPLA follows a character’s development across a narrative. 
\ours{} first builds an Arc that maps how a character’s values, motivations, or relationships change over the story. The benchmark then scores how well an RPLA’s responses fit the corresponding stages of the Arc, covering three distinct scenario types: scenes from the novel, new situations within its world, and situations outside that world.
We evaluate six models under six ways of providing narrative context. In every model, using the Arc up to the queried chapter yields the best performance, outperforming the strongest non-Arc context by 2.2–8.4 points. These results suggest that faithful role-play requires evolving character states and tracking their trajectory, rather than merely retrieving relevant episodic evidence. 
Our code is available at \url{https://github.com/holi-lab/ArcANE}.

\end{abstract}

\section{Introduction\label{sec:introduction}}

\begin{figure}[!t]
    \centering
    \includegraphics[width=\columnwidth]{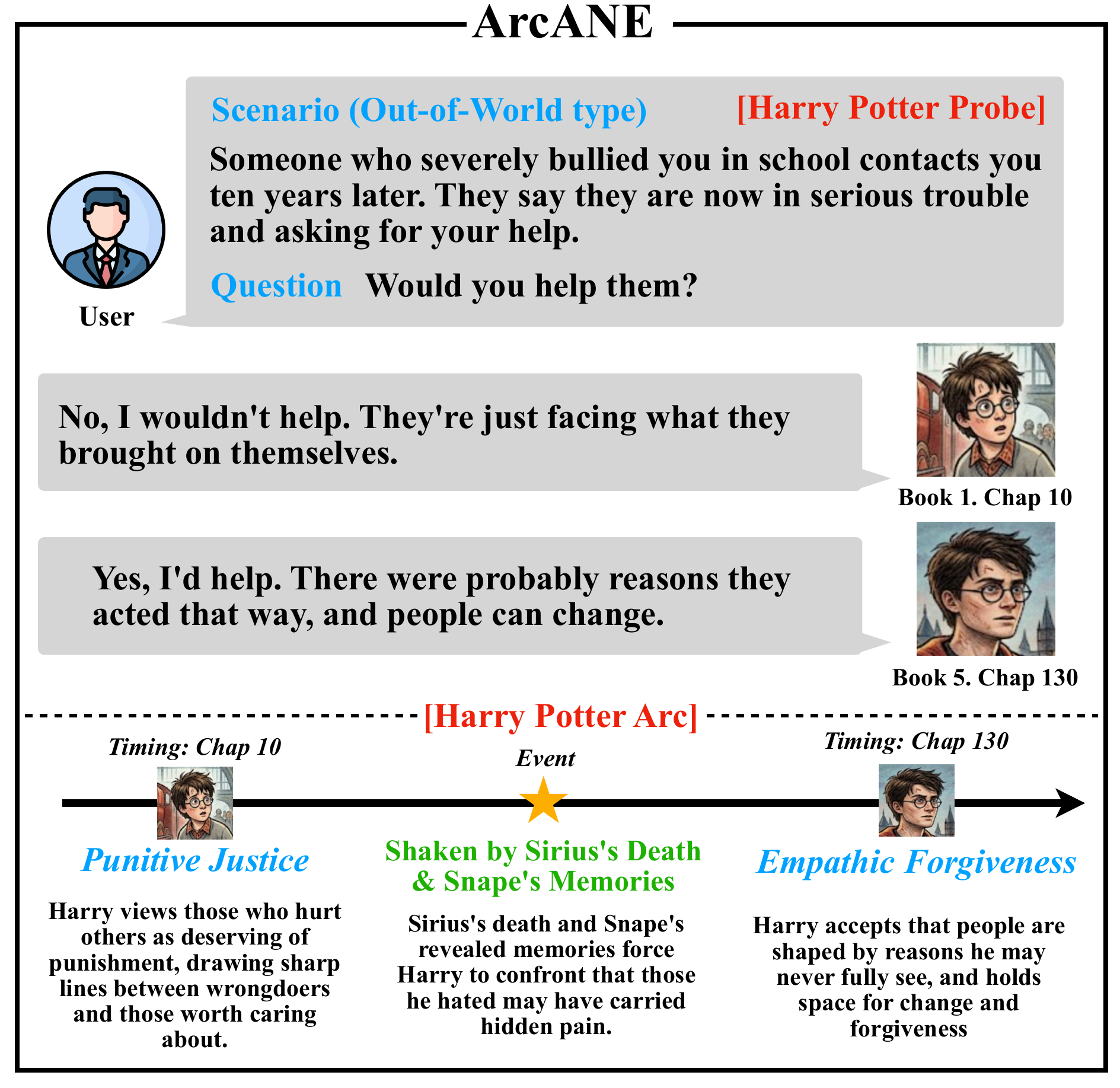}
    \caption{An example from \ours{}: an Out-of-World \emph{probe} elicits different moral responses from Harry across his \emph{Character Arc}, shifting from \textit{Punitive Justice} in Book 1 to \textit{Empathic Forgiveness} in Book 5.}
    \label{fig:introduction}
\end{figure}

Role-playing language agents (RPLAs) have become one of the most popular applications of conversational AI, enabling users to interact with fictional, historical, or persona-based characters~\citep{character_ai, chen2024personatoper}. As these agents have expanded across entertainment, companionship, interactive storytelling~\citep{park2023GenerativeAgent}, and education~\citep{Markel2023GPTeach}, user expectations have shifted from mere fluency toward authentic character portrayal~\citep{shao-etal-2023-characterllm, wang-etal-2024-incharacter}. RPLAs are thus expected to generate plausible~\citep{shanahan2023roleplaylargelanguagemodels}, in-character responses that sustain the user's sense of immersion~\citep{wu-etal-2025-towards-enhanced}.

Recent work has framed point-in-time role-playing as a task that situates a character at a specific moment in their timeline, motivated by the need to preserve narrative immersion~\citep{Ryan:2008:ICIDS, Ryan:2003:Book, ahn-etal-2024-timechara}. While much effort goes into suppressing explicit spoilers~\cite{tu-etal-2024-charactereval, sadeq-etal-2024-mitigating}, this alone is not sufficient. Beyond verifiable facts, a character's values and behavioral patterns also shift over time---as Figure~\ref{fig:introduction} illustrates, Harry Potter's moral stance differs sharply between Book~1 and Book~5. Evaluating RPLAs therefore requires assessing whether each response fits the character's state at that moment.

To address this gap, we aim to evaluate whether an RPLA's behavior shifts faithfully as a character evolves across the narrative. We focus on novels as our testbed: unlike short dialogues, scripts, or game transcripts, novels provide extended narratives in which character evolution unfolds vividly across many chapters, offering both rich internal description and explicit temporal structure~\citep{wang2025coser}. The core intuition behind our approach is simple: posing the same question at different points in the story should elicit different responses. This temporal approach goes beyond prior benchmarks such as \emph{TimeChara}~\citep{ahn-etal-2024-timechara}, which queries RPLAs at distinct points in time only to detect \emph{factual} hallucinations (i.e., whether the character knows what they should know at that moment), rather than how they would \emph{behave} in situations beyond their factual knowledge.

To this end, we introduce \ours{} (\textbf{Arc}-\textbf{A}ware \textbf{N}arrative \textbf{E}valuation), a benchmark that measures whether an RPLA's portrayed actions change in step with the source character's development across narrative phases. As illustrated in Figure~\ref{fig:introduction}, \ours{} represents this development through \emph{Character Arc}s and evaluates it using \emph{probes}. A \emph{Character Arc} organizes a character's development along a single psychological axis, such as a change in values, motivations, or relationships. It divides this trajectory into phases and grounds each phase in key narrative events. For instance, Harry's moral axis moves from \textit{Punitive Justice} (Book~1) to \textit{Empathic Forgiveness} (Book~5), with Sirius's death and Snape's revealed memories marking this change. A \emph{probe} then presents the same scenario and question at each phase of an Arc, together with phase-specific reference actions and thoughts that characterize how the character should respond. This setup tests whether an RPLA adapts its response to the character's position in the story rather than portraying the character as a fixed persona.

Our dataset contains $570$ Character Arcs and $4{,}854$ probes for $87$ principal characters across $19$ novels. The probes span three scenario types that progressively extend beyond the source text: \emph{In-Scenario} probes scenes from the novel, \emph{In-World} probes introduce new situations within its fictional world, and \emph{Out-of-World} probes place the character beyond that world. The latter two capture a central use case of RPLAs: asking how a character would respond to situations the source text never explores. Using these probes, we evaluate which form of narrative context best grounds an RPLA and which LLMs perform best as RPLAs.

To examine these two dimensions, we evaluate six LLMs under six context strategies. Across all six models, providing the Arc up to the queried chapter yields the highest Overall score, outperforming the strongest non-Arc context by $2.2$ to $8.4$ points (\S\ref{sec:main}). This advantage grows as scenarios extend further beyond the source text, suggesting that faithful role-playing requires not only retrieving narrative evidence but also tracking the character's state at the relevant phase. Beyond evaluation, we use the training split to fine-tune open-weight 8B and 32B models, producing \ours{}-8B and \ours{}-32B, which we release with the dataset. Fine-tuning further improves phase fidelity, with \ours{}-32B surpassing DeepSeek-V4-Pro on the held-out evaluation set (\S\ref{sec:additional}).

Our contributions are summarized as follows:
\begin{itemize}
    \item \textbf{Benchmark.}
    We introduce \ours{}, an automatically constructed benchmark for evaluating whether RPLAs reflect a character's development across narrative phases. It contains 570 Character Arcs and 4,854 probes across 19 novels and 87 characters (\S\ref{sec:dataset}).
    \item \textbf{Findings.} Across six models and six context strategies, providing the Arc up to the queried chapter performs best for every model, with larger gains as probes extend beyond the source text. Ablations show that these gains come from the correct trajectory information rather than the Arc format alone (\S\ref{sec:main}, \S\ref{sec:arc-ablation}). 
    \item \textbf{Training.} Using the \ours{} training split, we train and release five model variants: SFT and DPO models at 8B and 32B, and an RLVR model at 32B (\S\ref{sec:additional}, \S\ref{sec:rlvr}).

\end{itemize}

\section{Related Work\label{sec:related}}

{\renewcommand{\arraystretch}{0.95}
\setlength{\tabcolsep}{5pt}
\begin{table*}[t!]
\centering
\small
\begin{adjustbox}{width=\textwidth}
\begin{tabular}{lcccccccl}
\toprule
\textbf{Benchmark} & 
\textbf{Auto-constructed?} & 
\textbf{Free-form?} & 
\textbf{In-narrative?} & 
\textbf{Out-of-narrative?} & 
\textbf{Point-in-time?} & 
\textbf{Temporal behavioral?} & 
\textbf{Evaluated via} \\
\midrule
PsychoBench~\cite{huang2024psychobench}    & \xmark & \xmark & \xmark & \xmark & \xmark & \xmark & Trait \\
InCharacter~\cite{wang-etal-2024-incharacter}  & \cmark & \cmark & \xmark & \xmark & \xmark & \xmark & Trait \\
CharacterBox~\cite{wang-etal-2025-characterbox} & \cmark & \cmark & \cmark & \cmark & $\Delta$ & \xmark & Simulation \\
CoSER~\cite{wang2025coser} & \cmark & \cmark & \cmark & \xmark & $\Delta$ & \xmark & Simulation \\
Character-LLM~\cite{shao-etal-2023-characterllm} & \cmark & \cmark & \xmark & \cmark & \xmark & \xmark & Behavior \\
CharacterEval~\cite{tu-etal-2024-charactereval} & \cmark & \cmark & \cmark & \xmark & $\Delta$ & \xmark & Behavior \\
RoleBench~\cite{wang-etal-2024-rolellm}      & \cmark & \cmark & \xmark & \cmark & \xmark & \xmark & Behavior \\
ChatHaruhi~\cite{Li2023chatharuhi}    & \cmark & \cmark & \cmark & \xmark & \xmark & \xmark & Behavior \\
LifeChoice~\cite{xu-etal-2025-characterisdestiny} & \cmark & \xmark & \cmark & \xmark & \cmark & \xmark & Behavior \\
HPD~\cite{chen-etal-2023-hpd}                & \cmark & \cmark & \cmark & \xmark & \cmark & \cmark & Behavior \\
RoleEval~\cite{shen2024roleevalbilingualroleevaluation}      & \xmark & \xmark & \cmark & \xmark & \xmark & \xmark & Knowledge-Based Only \\
TimeCHARA~\cite{ahn-etal-2024-timechara} & \cmark & \cmark & \cmark & \xmark & \cmark & \xmark & Knowledge-Based Only \\
\midrule
\rowcolor{gray!15}
\textbf{\ours (Ours)}             & \cmark & \cmark & \cmark & \cmark & \cmark & \cmark & Behavior \\
\bottomrule
\end{tabular}
\end{adjustbox}
\caption{
Comparison of role-playing language agent evaluation benchmarks.
\textbf{Auto-constructed}: data generated primarily by LLMs.
\textbf{Free-form}: open-ended response generation.
\textbf{In-narrative}: scenario-based evaluation within source text.
\textbf{Out-of-narrative}: scenarios beyond the source text.
\textbf{Point-in-time}: evaluation anchored to specific narrative timepoints.
\textbf{Temporal behavioral}: assesses behavioral changes across narrative timepoints.
\textbf{Evaluated via}: modality through which persona is assessed.
$\Delta$: partial support.
}
\label{tab:benchmark_comparison}
\end{table*}
}

\subsection{Role-Playing Language Agents\label{sec:character_psychology}}

Role-playing language agents (RPLAs), in which LLMs simulate specific characters, have been deployed across diverse applications including interactive fiction, game NPCs, emotional companions, and personalized assistants~\cite{chen2024personatoper, wang-etal-2024-rolellm, gao-etal-2023-livechat, Xu_2022_COSPLAY, shanahan2023roleplaylargelanguagemodels, chen2025oscarsaitheatersurvey}. Beyond such practical utility, cognitive psychology research has shown that character simulation serves as an abstraction of human psychology and social experience, allowing people to explore minds, relationships, and possible selves through fictional figures~\cite{OATLEY2016618fictionsocialworlds}.

What users seek from RPLAs is consequently not mere response generation but an \emph{immersive experience of a living character}: an agent whose behavior remains coherent and responsive as the interaction unfolds, sustaining the user's narrative engagement~\cite{green2000, Ryan:2003:Book, Ryan:2008:ICIDS, Busselle23112009} with the figure being portrayed. A widely reported failure mode is that agents gradually lose persona grounding and produce generic responses even when the context remains intact~\cite{shin-etal-2025-spotting, luz-de-araujo-etal-2026-persistent}. Faithfully sustaining such a living character is therefore the central capability against which RPLA techniques must be evaluated.

\subsection{Benchmarking Role-Playing Language Agents\label{sec:rpla_bench}}

Benchmarks for RPLAs have developed around different targets of measurement: the character's trait inventory~\cite{huang2024psychobench, wang-etal-2024-incharacter}, factual knowledge and persona-driven decisions~\cite{shao-etal-2023-characterllm, xu-etal-2025-characterisdestiny, ahn-etal-2024-timechara}, surface representation such as linguistic style and conversational competence~\cite{tu-etal-2024-charactereval, Li2023chatharuhi, zhou-etal-2024-characterglm, wang-etal-2024-rolellm}, and behavioral consistency across turns~\cite{tu-etal-2024-charactereval, wang-etal-2025-characterbox}. Across these dimensions, however, the character is treated as a \emph{static target}---an identity to be reproduced rather than one whose behavior shifts as events accumulate~\cite{han2026personalityillusion, lee-etal-2025-trait, song2026humanpsychometricquestionnairesmischaracterize}.

Moving beyond this assumption, simulation-based benchmarks~\cite{wang2025coser, wang-etal-2025-characterbox} place characters in multi-turn scenarios to elicit dynamic behavior, and event-conditioned representations~\cite{park-etal-2025-charactergpt, li_behaviorchain} encode persona as snapshots or behavior chains. However, these capture variation from inter-character interaction or generic behavior chains rather than from a single character's events accumulating along the narrative. Benchmarks that consider narrative timepoints (Table~\ref{tab:benchmark_comparison}) remain limited to closed-form decisions~\cite{xu-etal-2025-characterisdestiny} or factual recall~\cite{ahn-etal-2024-timechara}---a gap we address by probing a character's behavior at specific narrative timepoints.

\begin{figure*}[t]
    \centering
    \includegraphics[width=\textwidth]{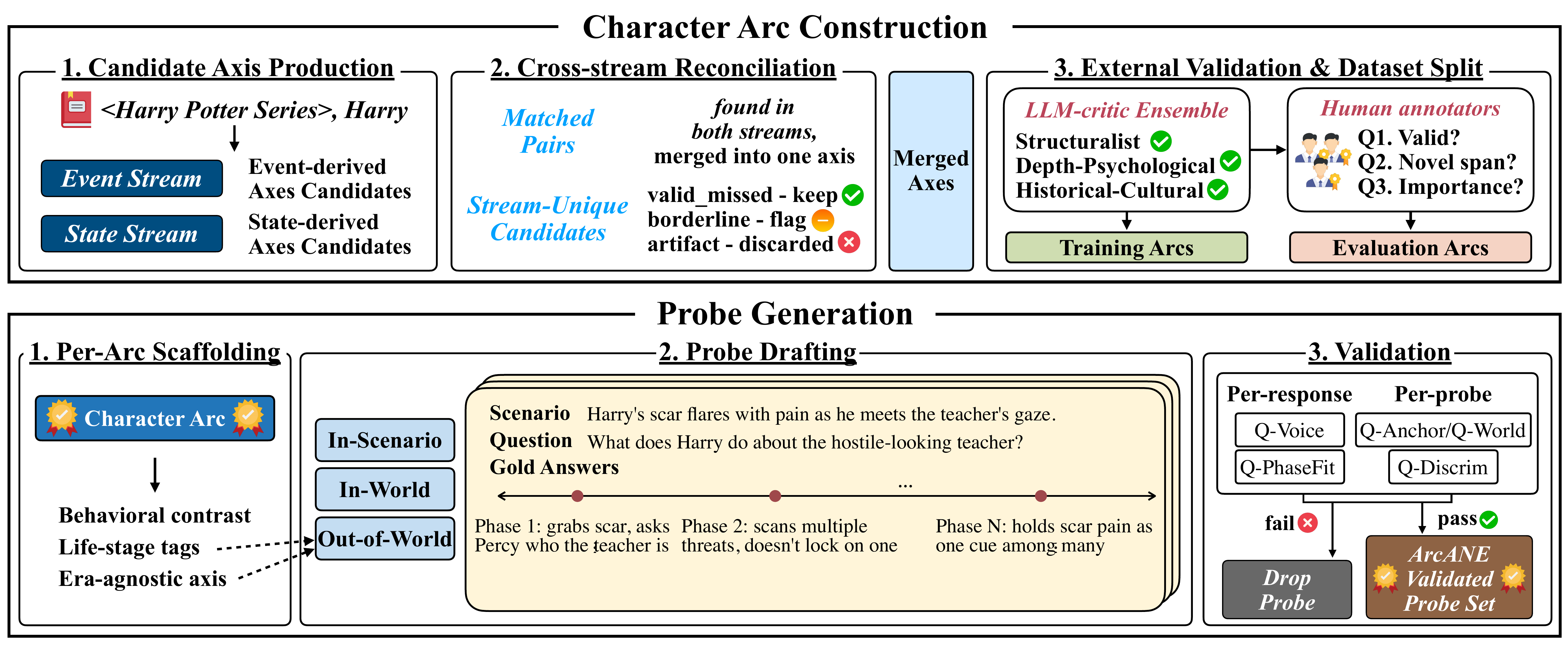}
    \caption{The \ours{} pipeline: constructing character arcs (top) and evaluation probes (bottom).}
    \label{fig:pipeline}
\end{figure*}

\section{\ours{} \label{sec:ArcANE}}

\ours{} (\textbf{Arc}-\textbf{A}ware \textbf{N}arrative \textbf{E}valuation) measures whether an RPLA's responses shift with the character's state at the specific phase, rather than matching a fixed persona. Prior RPLA benchmarks largely evaluate a character's trait inventory~\cite{huang2024psychobench, wang-etal-2024-incharacter}, which corresponds to McAdams' Layer~1~\cite{Mcadams1999, Mcadams2013}: a stable set of dispositions the character carries throughout the story. We move to Layer~2: whether the model expresses those traits at the right moment. 

The construction proceeds in two stages. First, \emph{Character Arc Construction} (\S\ref{sec:arc}) extracts what changes about each character, building one \emph{Character Arc} for every \emph{axis} the character moves along. An axis is a single psychological dimension defined by two pole descriptions. Figure~\ref{fig:introduction} shows one such arc, Harry's moral axis running from \textit{Punitive Justice} to \textit{Empathic Forgiveness} across the \emph{phases}. Second, \emph{Probe Generation} (\S\ref{sec:probe}) tests whether a model expresses the right phase at the right moment. Each arc yields probes consisting of a scenario-question pair with one phase-specific expected response per phase. Having the same scenario-question pair in every phase distinguishes a model that tracks the evolving state of the character from one that relies on a fixed persona. Figure~\ref{fig:introduction} illustrates this design with an Out-of-World probe built on the same Harry arc, where the scenario ``a former bully reaches out ten years later, asking for your help’’ draws a refusal at an early phase, hesitation in the middle, and acceptance at a late phase.

\subsection{Character Arc Construction\label{sec:arc}}

A Character Arc (Figure~\ref{fig:introduction}) aligns a character's key events with their evolving psychological states, organizing both into a phase-segmented trajectory from an initial state to a final state. A phase is a narrative span in which the character occupies a relatively stable position along the axis, with the number of phases determined by the narrative evidence rather than fixed in advance. Each phase is represented by a chapter range, a description of the character's current state, and the key moments that anchor that state. The three stages below correspond to the top panel of Figure~\ref{fig:pipeline}.

\paragraph{Stage~1: Candidate Generation.}
The novel is processed through two independent chapter-level streams: an \emph{event stream} extracts psychologically impactful events, and a \emph{state stream} emits a cross-sectional psychological profile. The split keeps event omission separable from state misreading even when both streams see the same chapters. Each stream then induces two types of candidate axes: \emph{intrapersonal} axes track internal change (beliefs, motives, coping), and \emph{relational} axes track a dyadic relation (trust, esteem, intimacy, antagonism). Every candidate is grounded in established literary or psychological scholarship~\cite{huang2025values, han-etal-2025-value} (Figure~\ref{fig:grounding}).

\paragraph{Stage~2: Reconciliation.}
An analyst LLM reconciles the candidate axes produced by the event and state streams in Stage~1. Matched pairs---axes proposed by both streams---are merged into a single axis with a direction label that records which pole each phase sits closer to. Each remaining unmatched candidate, proposed by only one stream, is then classified as a genuine missed axis (kept), an ambiguous case (flagged), or a stream-specific artifact (discarded).

\paragraph{Stage~3: Validation and Dataset split.}
The previous stage establishes internal reliability within a single LLM, not external validity. We therefore pass every axis through a critic LLM ensemble of three literary perspectives. For training novels, an axis is retained if at least two of the three critics judge it literarily grounded. For evaluation novels, the critic ratings serve only as reference: three human annotators (Appendix~\ref{app:contextmodes}) independently re-assess each axis, and only those judged valid by at least two annotators enter the evaluation set. On the validated slice, $205$ of $223$ candidate axes pass this rule (Appendix~\ref{app:arc-details}). The \textsc{Arc} context mode serves a truncated, filtered view of this arc, not the arc that defines the evaluation set (Appendix~\ref{app:contextmodes}).

\subsection{Probe Generation\label{sec:probe}}

A probe asks how a character would respond to one scenario at different phases of their arc. A single (scenario, question) is paired with $N$ reference responses, one per phase. Because the same scenario and question are presented at every phase, success requires more than knowing the character's overall personality. The model must identify which phase the character is in and respond from there. Each scenario falls into one of three categories that vary by distance from the source text. \emph{In-Scenario} lifts the scene from a verbatim passage, \emph{In-World} invents an unwritten situation inside the source's setting, and \emph{Out-of-World} transposes the scenario to a non-source era. These three categories form a difficulty gradient. In-Scenario can be answered from the source passage, In-World from the source's setting, and Out-of-World only from the arc itself. Each response pairs an action with a 1--2 sentence \emph{thought} that captures the cognitive construal when overt action is constrained~\cite{Mischel1995ACS}, plus a knowledge-cutoff chapter that blocks reference to later events. A final validation pass discards probes whose scenarios violate their category's rules. The three stages below correspond to the bottom panel of Figure~\ref{fig:pipeline}.

\paragraph{Stage~1: Per-Arc preparation.}
Before drafting any probe, we extract three reusable elements per arc. The \emph{behavioral contrast} pins down the axis as a single yes/no decision whose answer differs across phases, giving the designer the axis in concrete decisional form. A per-phase \emph{life-stage tag} from $\{$child, adolescent, young\_adult, adult, older\_adult$\}$ specifies the character's life stage at that phase. For Out-of-World only, an \emph{era-agnostic axis} specifies which era the scenario is transposed to and how the axis manifests in that setting.

\paragraph{Stage~2: Probe drafting.}
Let $N$ be the number of phases in the corresponding Character Arc. For each (target phase, category) pair, a designer LLM drafts one probe with $N$ phase-specific reference responses. One response reflects the character’s behavior at the target phase, while the remaining $N{-}1$ responses project the other phases’ behavior onto the same scenario. These counterfactual responses are grounded in the Character Arc, making them plausible for the character but incorrect for the target phase.

\paragraph{Stage~3: Validation and filtering.}
Each drafted probe is validated in two passes. The response pass applies two checks per response. Q-Voice verifies that the response stays in character, free of anachronism, and respects the knowledge cutoff. Q-PhaseFit asks a blind LLM judge which phase the response best fits, and flags any mismatch with the target phase. A failure on either check triggers one regeneration, and if it persists, the response is marked \texttt{unavailable} while the rest of the probe is kept. The probe pass applies two checks per probe. Q-Anchor (In-Scenario) or Q-World (In-World, Out-of-World) verifies that the scenario respects the setting's rules, and a failure drops the entire probe. Q-Discrim flags adjacent phase pairs whose responses are not separated along the behavioral contrast, but this check is annotation-only, since some overlap between adjacent phases is theoretically expected~\cite{Fleeson2001TowardAS}.

\subsection{\ours{} Dataset}
\label{sec:dataset}

\ours{} covers $19$ novels, $87$ principal characters, $570$ arcs, and $4{,}854$ probes in three disjoint subsets: a $12$-novel training slice, a $5$-novel human-validated evaluation slice, and a $2$-novel unvalidated low-popularity slice used as a memorization control (\S\ref{sec:additional}, Table~\ref{tab:data_stats}). Appendix~\ref{app:datapoint} gives the record schema, and Appendix~\ref{app:datasets_details} the per-stage identifiers and the overlap matrix.

\subsection{Training Arc-aware RPLAs}
\label{sec:training_rpla}

Although \ours{} is introduced primarily as an evaluation framework, its structure lends itself to a training formulation. Because each probe is answered across multiple phases of a character's arc, the dataset naturally yields contrastive pairs that could teach a model to distinguish subtle behavioral shifts between adjacent developmental stages.

Standard role-playing supervision encourages character-consistent responses, but does not explicitly teach models to distinguish how the same character should respond at different narrative phases. \ours{} provides this missing signal by pairing the same scenario and question with phase-specific responses. The model receives the character identity, query chapter, and context, and generates a free-form in-character response. We fine-tune Qwen3-8B and Qwen3-32B~\citep{yang2025qwen3} through a two-stage pipeline. In the SFT stage, the model learns the response format of \ours{}. In the subsequent DPO stage, it learns to distinguish ground-truth character behavior from plausible but temporally displaced alternatives, namely actions or speech that fit the character but belong to a different narrative phase. Full data construction and training details are provided in Appendix~\ref{app:training_details}.

\section{Experiments}
\label{sec:experiments}

\begin{table*}[!htbp]
\centering
\scriptsize
\setlength{\tabcolsep}{5.5pt}
\renewcommand{\arraystretch}{0.85}
\setlength{\aboverulesep}{0.2ex}
\setlength{\belowrulesep}{0.2ex}
\begin{tabular}{lllccccccccccccc}
\toprule
Family & Size & Mode & \multicolumn{4}{c}{In-Scenario} & \multicolumn{4}{c}{In-World} & \multicolumn{4}{c}{Out-of-World} & Overall \\
\cmidrule(lr){4-7} \cmidrule(lr){8-11} \cmidrule(lr){12-15} \cmidrule(lr){16-16}
 & & & APF & RPF & RAE & PTF & APF & RPF & RAE & PTF & APF & RPF & RAE & PTF & Overall \\
\midrule
\multirow[t]{12}{*}{DeepSeek-V4} & \multirow[t]{6}{*}{Flash} & Vanilla & 59.8 & 59.3 & 52.1 & 53.8 & 56.7 & 56.9 & 48.8 & 45.4 & 54.8 & 55.4 & 46.8 & 44.5 & 52.9 \\
 &  & Summary & 64.1 & 63.0 & 56.4 & 59.6 & 57.8 & 58.1 & 49.4 & 49.4 & 57.2 & 57.6 & 49.0 & 47.7 & 55.8 \\
 &  & RAG & 66.3 & 64.7 & 59.1 & 60.9 & 56.6 & 57.1 & 48.0 & 47.8 & 54.0 & 54.8 & 45.2 & 44.2 & 54.9 \\
 &  & LifeChoice & 66.3 & 64.7 & 59.1 & 61.7 & 57.4 & 57.8 & 48.6 & 49.5 & 56.8 & 57.0 & 48.2 & 46.5 & 56.1 \\
 &  & TimeChara & 60.8 & 60.2 & 53.2 & 55.6 & 56.2 & 56.6 & 48.4 & 46.8 & 54.5 & 55.2 & 46.3 & 43.5 & 53.1 \\
 &  & \cellcolor{yellow!15} Arc & \cellcolor{yellow!15} \textbf{66.8} & \cellcolor{yellow!15} \textbf{65.8} & \cellcolor{yellow!15} \textbf{59.2} & \cellcolor{yellow!15} \textbf{62.1} & \cellcolor{yellow!15} \textbf{61.8} & \cellcolor{yellow!15} \textbf{61.9} & \cellcolor{yellow!15} \textbf{53.8} & \cellcolor{yellow!15} \textbf{55.3} & \cellcolor{yellow!15} \textbf{62.2} & \cellcolor{yellow!15} \textbf{62.0} & \cellcolor{yellow!15} \textbf{54.5} & \cellcolor{yellow!15} \textbf{51.4} & \cellcolor{yellow!15} \textbf{59.7} \\
\cmidrule(l){2-16}
 & \multirow[t]{6}{*}{Pro} & Vanilla & 61.3 & 60.2 & 53.4 & 56.1 & 54.2 & 54.9 & 46.1 & 45.5 & 53.4 & 54.1 & 45.2 & 44.0 & 52.4 \\
 &  & Summary & 67.1 & 65.6 & 60.2 & 61.4 & 59.7 & 59.5 & 51.8 & 51.2 & 57.1 & 57.4 & 48.8 & 46.5 & 57.2 \\
 &  & RAG & 68.1 & 66.1 & 61.8 & 63.0 & 57.6 & 57.6 & 49.2 & 50.5 & 55.7 & 56.1 & 47.6 & 47.0 & 56.7 \\
 &  & LifeChoice & 69.1 & 66.9 & \textbf{63.0} & 64.0 & 59.1 & 58.9 & 51.1 & 50.9 & 56.7 & 56.8 & 48.7 & 47.6 & 57.7 \\
 &  & TimeChara & 60.4 & 59.5 & 53.1 & 54.8 & 54.9 & 55.4 & 46.9 & 46.7 & 53.0 & 53.7 & 44.8 & 42.5 & 52.1 \\
 &  & \cellcolor{yellow!15} Arc & \cellcolor{yellow!15} \textbf{69.5} & \cellcolor{yellow!15} \textbf{67.7} & \cellcolor{yellow!15} 62.8 & \cellcolor{yellow!15} \textbf{65.1} & \cellcolor{yellow!15} \textbf{64.2} & \cellcolor{yellow!15} \textbf{63.6} & \cellcolor{yellow!15} \textbf{57.0} & \cellcolor{yellow!15} \textbf{58.2} & \cellcolor{yellow!15} \textbf{64.1} & \cellcolor{yellow!15} \textbf{63.7} & \cellcolor{yellow!15} \textbf{57.0} & \cellcolor{yellow!15} \textbf{55.8} & \cellcolor{yellow!15} \textbf{62.4} \\
\midrule
\multirow[t]{12}{*}{Qwen3} & \multirow[t]{6}{*}{8B} & Vanilla & 43.0 & 42.8 & 32.7 & 38.2 & 40.5 & 41.0 & 30.8 & 30.6 & 41.4 & 42.4 & 32.1 & 30.6 & 37.2 \\
 &  & Summary & 47.2 & 45.4 & 36.5 & 41.7 & 42.1 & 41.9 & 32.1 & 31.6 & 41.7 & 41.9 & 31.7 & 30.3 & 38.7 \\
 &  & RAG & \textbf{53.0} & \textbf{51.0} & \textbf{43.2} & \textbf{44.6} & 43.0 & 43.1 & 33.1 & 31.6 & 42.2 & 42.7 & 32.8 & 30.3 & 40.9 \\
 &  & LifeChoice & 50.9 & 48.9 & 40.8 & 42.4 & 42.6 & 42.3 & 32.1 & 31.4 & 42.3 & 42.2 & 32.1 & 30.0 & 39.8 \\
 &  & TimeChara & 43.1 & 43.1 & 33.3 & 36.5 & 41.6 & 42.2 & 32.0 & 30.6 & 41.8 & 42.8 & 32.7 & 29.4 & 37.4 \\
 &  & \cellcolor{yellow!15} Arc & \cellcolor{yellow!15} 50.1 & \cellcolor{yellow!15} 48.5 & \cellcolor{yellow!15} 39.2 & \cellcolor{yellow!15} 42.1 & \cellcolor{yellow!15} \textbf{47.5} & \cellcolor{yellow!15} \textbf{47.3} & \cellcolor{yellow!15} \textbf{37.3} & \cellcolor{yellow!15} \textbf{36.6} & \cellcolor{yellow!15} \textbf{47.4} & \cellcolor{yellow!15} \textbf{47.5} & \cellcolor{yellow!15} \textbf{37.4} & \cellcolor{yellow!15} \textbf{36.0} & \cellcolor{yellow!15} \textbf{43.1} \\
\cmidrule(l){2-16}
 & \multirow[t]{6}{*}{32B} & Vanilla & 49.7 & 48.6 & 39.4 & 43.4 & 48.6 & 48.6 & 38.8 & 35.8 & 48.6 & 48.6 & 39.2 & 35.0 & 43.7 \\
 &  & Summary & 54.8 & 53.6 & 45.2 & 46.7 & 49.5 & 49.3 & 39.2 & 36.0 & 48.5 & 48.7 & 38.9 & 36.6 & 45.6 \\
 &  & RAG & \textbf{59.4} & \textbf{57.3} & \textbf{50.8} & 49.2 & 49.6 & 49.9 & 40.0 & 38.6 & 48.3 & 48.7 & 39.0 & 36.1 & 47.2 \\
 &  & LifeChoice & 58.3 & 56.6 & 49.1 & 50.5 & 50.6 & 50.5 & 40.9 & 40.3 & 48.4 & 48.6 & 38.8 & 36.3 & 47.4 \\
 &  & TimeChara & 51.3 & 50.4 & 41.3 & 45.4 & 48.9 & 49.1 & 39.5 & 36.4 & 48.4 & 48.9 & 39.3 & 34.4 & 44.4 \\
 &  & \cellcolor{yellow!15} Arc & \cellcolor{yellow!15} 57.4 & \cellcolor{yellow!15} 55.6 & \cellcolor{yellow!15} 47.4 & \cellcolor{yellow!15} \textbf{51.6} & \cellcolor{yellow!15} \textbf{54.1} & \cellcolor{yellow!15} \textbf{53.7} & \cellcolor{yellow!15} \textbf{44.3} & \cellcolor{yellow!15} \textbf{41.0} & \cellcolor{yellow!15} \textbf{54.8} & \cellcolor{yellow!15} \textbf{54.3} & \cellcolor{yellow!15} \textbf{45.4} & \cellcolor{yellow!15} \textbf{41.7} & \cellcolor{yellow!15} \textbf{50.1} \\
\midrule
\multirow[t]{12}{*}{\ours-DPO} & \multirow[t]{6}{*}{8B} & Vanilla & 47.8 & 46.9 & 38.3 & 41.9 & 48.2 & 47.5 & 39.0 & 37.4 & 47.4 & 46.8 & 38.5 & 36.3 & 43.0 \\
 &  & Summary & 56.5 & 55.0 & 47.7 & 49.7 & 49.4 & 48.8 & 40.0 & 40.3 & 50.3 & 49.6 & 41.2 & 38.3 & 47.2 \\
 &  & RAG & \textbf{61.4} & \textbf{59.5} & \textbf{54.4} & 53.0 & 48.8 & 48.8 & 40.0 & 39.6 & 49.6 & 49.4 & 41.2 & 36.3 & 48.5 \\
 &  & LifeChoice & 58.2 & 56.2 & 50.1 & 50.5 & 49.1 & 48.6 & 40.1 & 39.3 & 50.3 & 49.9 & 41.7 & 37.9 & 47.6 \\
 &  & TimeChara & 49.3 & 48.2 & 40.2 & 43.0 & 47.8 & 47.3 & 38.8 & 38.9 & 47.7 & 47.4 & 39.3 & 36.5 & 43.7 \\
 &  & \cellcolor{yellow!15} Arc & \cellcolor{yellow!15} 59.5 & \cellcolor{yellow!15} 57.7 & \cellcolor{yellow!15} 50.6 & \cellcolor{yellow!15} \textbf{56.0} & \cellcolor{yellow!15} \textbf{60.9} & \cellcolor{yellow!15} \textbf{59.3} & \cellcolor{yellow!15} \textbf{52.3} & \cellcolor{yellow!15} \textbf{54.6} & \cellcolor{yellow!15} \textbf{62.5} & \cellcolor{yellow!15} \textbf{60.5} & \cellcolor{yellow!15} \textbf{54.5} & \cellcolor{yellow!15} \textbf{54.6} & \cellcolor{yellow!15} \textbf{56.9} \\
\cmidrule(l){2-16}
 & \multirow[t]{6}{*}{32B} & Vanilla & 53.6 & 53.2 & 44.9 & 45.0 & 55.1 & 55.1 & 46.7 & 41.7 & 56.5 & 56.2 & 48.6 & 41.5 & 49.9 \\
 &  & Summary & 57.0 & 55.8 & 48.4 & 49.5 & 53.2 & 52.6 & 44.0 & 44.3 & 51.9 & 51.6 & 42.6 & 40.6 & 49.3 \\
 &  & RAG & \textbf{65.8} & \textbf{63.5} & \textbf{59.2} & \textbf{55.8} & 52.9 & 52.9 & 43.8 & 41.9 & 51.6 & 51.8 & 42.9 & 41.6 & 52.0 \\
 &  & LifeChoice & 62.1 & 60.2 & 54.6 & 53.6 & 54.9 & 54.2 & 45.8 & 44.3 & 54.4 & 53.9 & 45.3 & 41.1 & 52.0 \\
 &  & TimeChara & 53.7 & 53.2 & 44.8 & 46.9 & 55.6 & 55.0 & 47.5 & 41.8 & 56.4 & 56.1 & 48.7 & 39.6 & 49.9 \\
 &  & \cellcolor{yellow!15} Arc & \cellcolor{yellow!15} 59.4 & \cellcolor{yellow!15} 58.0 & \cellcolor{yellow!15} 50.9 & \cellcolor{yellow!15} 53.4 & \cellcolor{yellow!15} \textbf{66.2} & \cellcolor{yellow!15} \textbf{64.2} & \cellcolor{yellow!15} \textbf{58.8} & \cellcolor{yellow!15} \textbf{58.5} & \cellcolor{yellow!15} \textbf{68.0} & \cellcolor{yellow!15} \textbf{65.7} & \cellcolor{yellow!15} \textbf{61.6} & \cellcolor{yellow!15} \textbf{59.8} & \cellcolor{yellow!15} \textbf{60.4} \\
\bottomrule
\end{tabular}
\caption{Main results on the validated slice (5 novels). Probe categories and metrics are defined in \S\ref{sec:probe} and \S\ref{sec:evalproto}. Best mode per column per model is bolded.}
\label{tab:main_results}
\end{table*}

\subsection{Experimental Setup\label{sec:setup}}

We evaluate six models under six in-context strategies on the validated evaluation slice of \S\ref{sec:dataset}: five novels (\emph{Harry Potter}, \emph{Anna Karenina}, \emph{Don Quixote}, \emph{The Count of Monte Cristo}, and \emph{The Autobiography of Benjamin Franklin}) covering 25 principal characters, 205 arcs, and 1{,}754 probes.

\paragraph{Models.}
We compare four open-weight baselines, Qwen3-8B, Qwen3-32B~\cite{yang2025qwen3}, DeepSeek-V4-Flash, and DeepSeek-V4-Pro~\cite{deepseekai2026deepseekv4}, and our post-trained models, \ours{}-8B-DPO and \ours{}-32B-DPO.

\paragraph{Context modes.}
\emph{Vanilla} supplies only the character's identity and the query chapter. \emph{Summary} adds the most recent five chapter summaries, while \emph{RAG} adds the top-6 retrieved source-text chunks. In both cases, context is truncated at the query chapter. \emph{LifeChoice}~\cite{xu-etal-2025-characterisdestiny} and \emph{TimeCHARA}~\cite{ahn-etal-2024-timechara} reuse the corresponding original context formats. \emph{\textsc{Arc}} (ours) supplies the character arc automatically constructed from the test novel by the pipeline in \S\ref{sec:dataset}, truncated at the phase of the query chapter. Full context-mode details are in Appendix~\ref{app:contextmodes}; all prompts are in Appendix~\ref{app:prompts}.

\subsection{Evaluation Protocol\label{sec:evalproto}}

The role-playing language agent receives the probe (scenario, question, and the chapter-truncated context produced by the current mode) and generates a free-form response along with its reasoning. An LLM judge (DeepSeek-V4-Flash; full setup, prompts, and human-annotator plausibility validation in Appendix~\ref{app:judge}) scores the response against the reference attached to the probe phase on a $1$--$100$ scale, at two granularities.

\textbf{Per-phase metrics.}
Each (probe, phase) pair receives three metric scores. \textbf{APF (Action Phase-Fidelity)} grades the response's overt action against \texttt{ref\_action} at three levels (strategy, valence, target). \textbf{RPF (Reasoning Phase-Fidelity)} parses both \texttt{ref\_thought} and the response's reasoning into four mechanism slots (trigger, appraisal, goal, strategy) and aggregates the slot-wise matches. \textbf{RAE (Reasoning--Action Entailment)} fixes \texttt{ref\_thought} as the operative reasoning and asks whether the response's action is one this reasoning would license, catching responses that are internally coherent yet reach an action \texttt{ref\_thought} would not.

\textbf{Trajectory metric.}
Per-phase metrics do not penalize a model that scores well on each phase in isolation yet collapses adjacent phases or moves along the axis in the wrong direction. \textbf{PTF (Phase Trajectory Fidelity)} addresses this: a single judge call sees all $N$ phase-keyed (reference, response) pairs together and returns the mean of three sub-scores (alignment, direction, shape).

\textbf{Aggregation.}
Scores are pooled within each novel, novels are averaged with equal weight, and Overall is the mean of the twelve probe-type by metric cells (Appendix~\ref{app:judge}).

\subsection{Main Results\label{sec:main}}

Across the six models and six context modes from \S\ref{sec:setup}, conditioning on the Character Arc gives the strongest Overall score on every model. Table~\ref{tab:main_results} reports Overall and the four per-category metrics for each (model, mode) cell. The Arc row leads the Overall column for all six models, with a gap of $2.2$ to $8.4$ points over the same model's strongest non-Arc row. At the high-capacity end, DeepSeek-V4-Pro reaches $62.4$ under Arc against $57.7$ for its best non-Arc mode (LifeChoice). At the small open-weight end, Qwen3-8B reaches $43.1$ under Arc against $40.9$ for its best non-Arc mode (RAG). The pattern survives a per-novel split: across the $30$ (model, novel) Overall cells produced by evaluating each of the six models on each of the five validated novels, Arc is the top mode on $29$ (per-novel breakdown in Appendix~\ref{app:memorization}). Resampling the $199$ contributing (novel, character, axis) clusters within novel, all $30$ paired $95\%$ intervals exclude zero (Appendix~\ref{app:uncertainty}).

The lift is not uniform across the three probe categories. The Arc-vs-best-non-Arc gap on Overall in the DeepSeek-V4-Pro row is $+0.5$ on In-Scenario, $+5.2$ on In-World, and $+7.7$ on Out-of-World. The same ordering recurs in the other models, and the probe construction explains it. An In-Scenario probe takes its scenario from a verbatim source passage, so source-text retrieval (RAG, LifeChoice) already returns the relevant scene and Arc adds little. In-World and Out-of-World probes have no such passage, and only Arc supplies the phase of the query chapter.

The Arc lift is consistently larger on the trajectory metric (PTF) than on the per-phase metrics (APF, RPF, RAE). APF, RPF, and RAE judge each (probe, phase) pair in isolation, so a non-Arc mode matching plausible per-phase content can score well phase by phase without shifting along the arc. PTF grades all $N$ phase responses to a probe as one sequence (\S\ref{sec:evalproto}) and rewards only sequences whose alignment, direction, and shape track the reference across phases. For DeepSeek-V4-Pro, the Arc-vs-best-non-Arc PTF gap reaches $+1.1$ on In-Scenario, $+7.0$ on In-World, and $+8.2$ on Out-of-World, each at least as large as the matched gap averaged over the three per-phase metrics ($+0.3$, $+4.6$, $+7.2$). The same ordering holds for the other models. Arc is the only mode that supplies the per-phase trajectory that PTF scores against.

\subsection{Additional Results\label{sec:additional}}

We extend \S\ref{sec:main} in two directions: low-popularity novels outside the validated subset, and role-playing models beyond those in Table~\ref{tab:main_results}. We evaluate the same six models on two low-popularity titles, \emph{He Knew He Was Right} and \emph{The Odd Women}, using probes constructed with the same pipeline.\footnote{Validated public-domain titles, downloads/month: Monte Cristo $80{,}951$, Anna Karenina $17{,}785$, Don Quixote $12{,}669$, Benjamin Franklin $2{,}187$. Harry Potter is non-Gutenberg. Low-popularity titles, downloads/month: He Knew He Was Right $1{,}069$, The Odd Women $1{,}219$.} Arc is again the top Overall mode for all six models ($+3.81$ to $+12.54$ over the strongest non-Arc mode, leading in all $12$ model-by-novel comparisons), and the category ordering of \S\ref{sec:main} holds (Appendix~\ref{app:extra3-results}). Character-class results are in Appendix~\ref{app:character-class}.

Figure~\ref{fig:added_models} carries the \S\ref{sec:main} Arc-lift pattern to other role-playing models and isolates the SFT and DPO contributions inside \ours{}. It plots the per-category Arc gap to best baseline for HER-32B~\cite{du2026herhumanlikereasoningreinforcement}, CoSER-8B and CoSER-70B~\cite{wang2025coser}, and the SFT and DPO variants of \ours{} at both sizes (full numbers in Appendix~\ref{app:added-results}). Every entry has a non-negative lift on In-World and Out-of-World, and In-Scenario stays mixed: the asymmetry generalizes beyond Table~\ref{tab:main_results}. Inside \ours{}, SFT broadly raises Arc Overall from $50.1$ to $58.4$ but its Arc lift on In-World and Out-of-World averages only $+7.3$ at \ours{}-32B-SFT. DPO reaches further into those categories, growing it to $+12.5$ at \ours{}-32B-DPO, at the cost of a smaller In-Scenario lift from $-2.2$ to $-5.6$. We analyze this pattern in detail in~\S\ref{sec:compare_sft_dpo}.

\begin{figure}[!t]
\centering
\includegraphics[width=\linewidth]{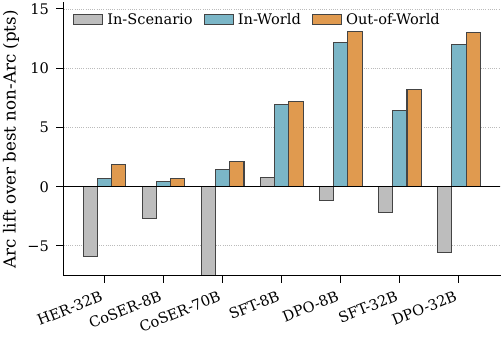}
\caption{Per-category Arc lift (Arc minus best non-Arc) for added RPLA baselines and the \ours{} models. SFT/DPO at each size are adjacent. Tables~\ref{tab:main_results} and~\ref{tab:main_results_added}.}
\label{fig:added_models}
\end{figure}

\section{Analysis\label{sec:analysis}}

\subsection{What Information in the Arc Matters?\label{sec:arc-ablation}}

We use two controlled variants to ask whether the Arc format alone is sufficient and how much of the correct Arc content a model needs. \textsc{MixedArc} preserves the Arc schema and truncation rule but replaces the content with another character's arc from the same novel. \textsc{ArcHint} retains only the name of each axis and the character's current phase number and label, removing the phase descriptions, pole definitions, and supporting evidence. This reduces the context by approximately $40\times$. We evaluate both variants on approximately 120 validated probes per model for DeepSeek-V4-Flash, Qwen3-32B, and \ours{}-32B-DPO (Figure~\ref{fig:leakage_bars}; details in Appendix~\ref{app:arc-ablation-extra}).

\textsc{MixedArc} shows that the Arc format alone is not sufficient. It fails to outperform Vanilla on Qwen3-32B and \ours{}-32B-DPO, the two models for which the correct Arc produces a large gain. On \ours{}-32B-DPO, it falls $4.7$ points in Per-Phase Average and $6.5$ in PTF below Vanilla. The degradation from incorrect content is expected; the relevant result is that preserving the same schema and truncation rule does not preserve the Arc gain.
\textsc{ArcHint} then tests how much of the correct content is needed. On DeepSeek-V4-Flash and Qwen3-32B, it remains within $\pm2.6$ points of the full Arc on every metric, showing that compact axis and phase cues provide nearly the full prompting benefit. Howevere, on \ours{}-32B-DPO it recovers only about half of the Arc-over-Vanilla gain, falling $5.1$ points below the full Arc in Per-Phase Average. \ours{}-32B-DPO therefore benefits from the detailed phase descriptions, whereas the other two models obtain most of the Arc gain from compact phase cues alone.

\begin{figure}[!t]
\centering
\includegraphics[width=\columnwidth]{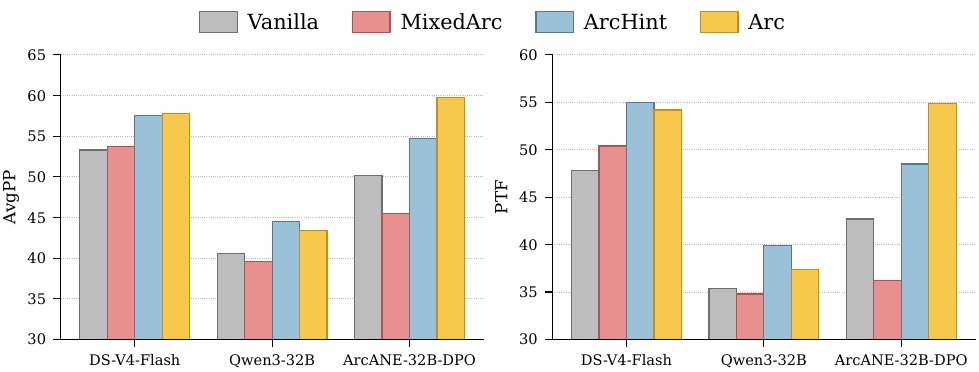}
\caption{Arc source-of-effect ablation across three models, validated subset. DS refers to DeepSeek.}
\label{fig:leakage_bars}
\end{figure}

\subsection{Evaluation validation\label{sec:judge-agreement}}

\paragraph{Human-anchored judge plausibility.} Three annotators rate the DeepSeek-V4-Flash judge verdict on $70$ probes drawn across $7$ novels and all $6$ context modes. Majority plausibility is $87.1\%$, with all three categories falling within $10$ points (In-Scenario $84.6\%$, In-World $82.3\%$, Out-of-World $95.7\%$). On a $50$-cell subset the annotators also re-scored APF/RPF/RAE on the same $1$--$100$ scale, anchored to the judge's verdict; the human-average Overall tracks the judge at Pearson $r{=}0.96$, so where humans adjusted the judge they shifted it by only a few points. Per-dimension correlations, agreement coefficients, and the full re-score table are in Appendix~\ref{app:judge-crossval-human}.

\paragraph{LLM cross-judge replication.} On an independent $300$-cell sample stratified across $5$ novels and $4$ (model, mode) configs, we re-score with three different judges (Claude Sonnet 4.5, Opus 4.5, and GPT-5.5). DeepSeek-V4-Flash sits inside the three-judge consensus, and all four judges rank \ours{}-32B-DPO under Arc first, and the full ranking table are in Appendix~\ref{app:judge-crossval-cross}.

\paragraph{PTF metric sensitivity.} Perturbing the probe response sequences (response-shuffle and response-reverse, with per-dimension drops reported in Appendix~\ref{app:ptf-validity}) drops PTF Overall on \ours{}-32B-DPO by $-8.8$ to $-23.0$ points against $-1.3$ to $-6.3$ on DeepSeek-V4-Pro: PTF responds to phase-order corruption, and the larger drop on the model with stronger per-phase anchoring reads as a behavioral signature of phase tracking rather than a per-phase re-summary. A third perturbation, block-shuffle, is a diagnostic only. Its effect disappears under a matched order-agnostic instruction, and every scored trajectory is rendered from \texttt{phase\_idx}-sorted blocks (Appendix~\ref{app:ptf-validity}).

\subsection{Training-effect analysis\label{sec:error-analysis}}
\ours{}-32B-DPO beats Qwen3-32B on $1{,}198/1{,}750$ validated subset probes (ties $84$, losses $468$, mean Overall $\Delta{=}{+}9.49$), with the category asymmetry of \S\ref{sec:main}. Side-by-side inspection of the wins surfaces one stylistic difference between the two models: \ours{} writes in the first-person, present-tense register the judges prefer, while Qwen3-32B defaults to third-person narration. A hostile reading is that the gain is a register artifact rather than evidence of arc tracking. We test this with a POV control. On a stratified $150$-probe sample we append one register instruction to Qwen3-32B's prompt (\emph{``respond IN CHARACTER, first person, present tense''}) and rescore the regenerated responses under the unchanged Stage-iii rubric. Forcing the register \emph{lowers} Qwen's Overall to $50.0$ (from $53.8$), while \ours{} holds at $56.7$. If register carried the gain, the instruction would close it. A proposition-level decomposition on the same $150$ probes attributes the surviving gap to category-conditional canon discipline, not surface register (Appendix~\ref{app:pairwise-coding}, Table~\ref{tab:propositions}).

\paragraph{What does DPO add to SFT?}
\label{sec:compare_sft_dpo}

DPO beats SFT on $940/1{,}750$ probes, with the same category trade-off: SFT wins In-Scenario, whereas DPO widens the gains on In-World and Out-of-World. This difference may arise from DPO's contrastive supervision. SFT sees only target responses, so characteristics shared across phases can dominate the learning signal. DPO instead compares each target with a plausible response from an adjacent phase; because both responses are in character, their contrast highlights the subtler change between phases. For example, both responses for Don Quixote retain his chivalric manner, but only the target for his repentant phase softens the knightly pose and turns toward family. DPO directly learns this distinction, whereas SFT must infer it from the target alone.

\subsection{Reinforcement Learning with Verifiable Rewards\label{sec:rlvr}}

\paragraph{Why rubric-based RL?} SFT and DPO train from fixed target responses and pairwise preferences. However, \ours{} also provides a per-phase rubric that can score newly generated responses, raising a further question: can this rubric provide direct feedback for improving phase fidelity without additional reward annotations? To test this, we continue training \ours{}-32B-DPO with GRPO~\cite{shao2024deepseekmathpushinglimitsmathematical}. The reward is $r=\mathrm{mean}(\textrm{APF},\textrm{RPF},\textrm{RAE})/100$, scored by Qwen3.6-27B against the corresponding phase reference. Training prompts come from the DPO pool, while evaluation is conducted on the disjoint validated slice under \textsc{Arc} context using a separate judge, DeepSeek-V4-Flash (Appendix~\ref{app:rlvr}).

\begin{figure}[!t]
\centering
\includegraphics[width=\columnwidth]{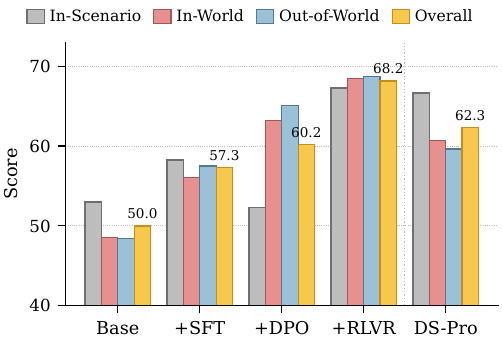}
\caption{Performance across \ours{}-32B training stages by probe category under \textsc{Arc} context on the validated slice.}
\label{fig:rlvr_stages}
\end{figure}

\paragraph{Results.}
As Figure~\ref{fig:rlvr_stages} shows, the resulting model reaches $68.2$ Overall, compared with $60.2$ for DPO, $57.3$ for SFT, and $50.0$ for the base model. It improves all three probe categories, recovering DPO's loss on In-Scenario while preserving its gains on In-World and Out-of-World. The directly rewarded metrics improve consistently across all three judges ($+8.8$, $+9.3$, and $+8.8$ over DPO; Table~\ref{tab:rlvr-crossjudge}). PTF is less consistent ($+5.7$, $-5.0$, and $-5.8$), as each rollout and reward covers only one phase. Thus, the rubric provides an effective reward for per-phase fidelity, while learning cross-phase progression will require a trajectory-level reward.

\section{Conclusion\label{sec:conclusion}}
We study a dimension of faithful role-play overlooked by existing benchmarks: whether an RPLA reflects a character's changing state at the appropriate point in a narrative. We introduce \ours{}, an automatically constructed benchmark with 570 Character Arcs and 4,854 phase-specific probes spanning scenarios within and beyond the source text. Across models, Arc-grounded context performs best, especially beyond the source text where retrieval alone is insufficient. Training on \ours{} further improves phase fidelity. By making phase fidelity measurable, \ours{} broadens character-fidelity evaluation from asking whether a response is in character to whether it is in character at that point in the story.

\section*{Limitations\label{sec:limitations}}
While \ours{} offers a new perspective for evaluating role-playing language agents, several limitations remain. \ours{} asks whether a response matches one validated account of a character at a point in the story, and that account is not the only defensible reading. The Arc context and the phase references come from the same LLM pipeline, so only readers can check one against the other. Extending our blind study beyond the one character it covers is the task we leave open. \ours{} also looks at one character at a time, so relational axes record how a character stands toward someone else rather than live dialogue between two agents. Adapting to multi-turn dialogue requires evaluating whether a model maintains its phase across turns or appropriately shifts when challenged by the user, which demands new annotations for counterfactual scenarios the novel never explores. We leave user influence and the choice of axis to that setting. We order phases by chapter rather than by story time, so strongly nonlinear novels would need story-time alignment first. Finally, we work from English editions of novels written mostly in the 19th and early 20th centuries, and models trained on them may repeat the social attitudes of those periods. Stronger agents also make impersonation easier. We therefore release our artifacts for research only, and ask that applications disclose when a character is AI-generated.

\section*{Acknowledgments}
This work was supported by the National Research Foundation of Korea (NRF) grants funded by the Korean government (MSIT) (RS-2024-00333484, RS-2024-00414981), and by the Institute of Information \& Communications Technology Planning \& Evaluation (IITP) grant (No. RS-2026-25551943, Military-Industry-Academia Cooperation Center (Seoul (Yongsan))) funded by the Korean government (MND). This work was also supported by the National Supercomputing Center with supercomputing resources including technical support (KSC-2025-CRE-0514).

\bibliography{custom}

\clearpage
\appendix

\section{Character Arc Construction Details\label{app:arc-details}}

\paragraph{Why two streams.}
A single end-to-end pass conflates two distinct failure modes: missing a psychologically impactful event, and misreading the character's state given the same events. Running the event and state streams in parallel keeps the two failures separable in any downstream audit, and partial convergence between the streams serves as an internal reliability check. Each chapter is processed independently, with a recurrent summary that carries the prior chapter's running state forward.

\begin{figure}[t]
    \centering
    \includegraphics[width=0.8\columnwidth]{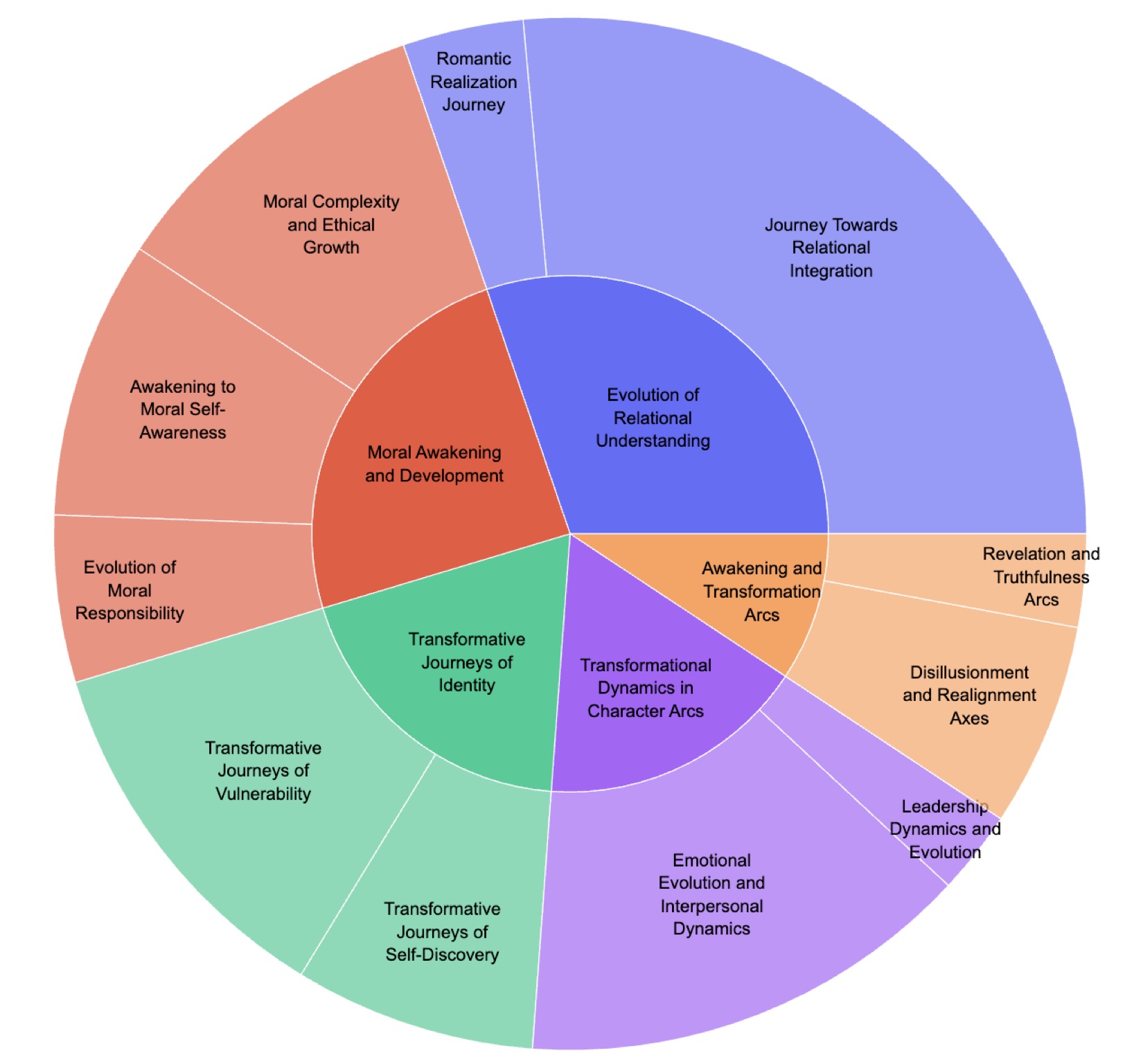}
    \caption{Induced axes are clustered and grounded against established literary or psychological scholarship, following the construct-grounding criterion of \emph{Values in the Wild}~\cite{huang2025values}.}
    \label{fig:grounding}
\end{figure}

\paragraph{Axis induction constraints.}
Each character receives 3--5 intrapersonal axes plus one relational axis per target (a relational target is admitted only when the text exhibits non-trivial trackable change). The first intrapersonal axis is anchored to the agency--communion dimension~\cite{Mcadams1999, Mcadams2013} as a sanity coordinate, and its absence in the induced set is a strong drift signal. Every induced axis must ground in established literary or psychological scholarship, following the construct-grounding criterion of \emph{Values in the Wild}~\cite{huang2025values} (Figure~\ref{fig:grounding}).

\paragraph{Reconciliation tag set.}
Stream-unique candidates receive one of three tags, each tied to a distinct post-processing outcome (Figure~\ref{fig:pipeline}, top middle). \texttt{valid\_missed} (\emph{kept}) marks a candidate that is plausible on review but absent from the other stream's set, and is added to the merged axis set. \texttt{borderline} (\emph{flagged}) marks candidates with insufficient evidence to commit either way, which are retained but tagged for downstream filtering. \texttt{artifact} (\emph{discarded}) marks candidates that are products of how the evidence was scoped (for example, a transient event-stream salience). Matched pairs receive an arc-direction label that extends McAdams' redemptive / contaminating dichotomy to non-linear trajectories such as oscillating or U-shaped arcs.

\paragraph{LLM-critic ensemble.}
The Stage~iii critics adopt three literary perspectives: a structuralist / narratologist critic that reads the axis against the novel's plot architecture, a depth-psychological critic that reads it against psychoanalytic and developmental frameworks, and a historical / cultural critic that reads it against the novel's social-historical context. Each critic attaches scholarly citations as candidate evidence, not as ground truth. The citations are passed to human annotators alongside the axis for their own use.

\begin{figure}[!t]
\centering
\includegraphics[width=\linewidth]{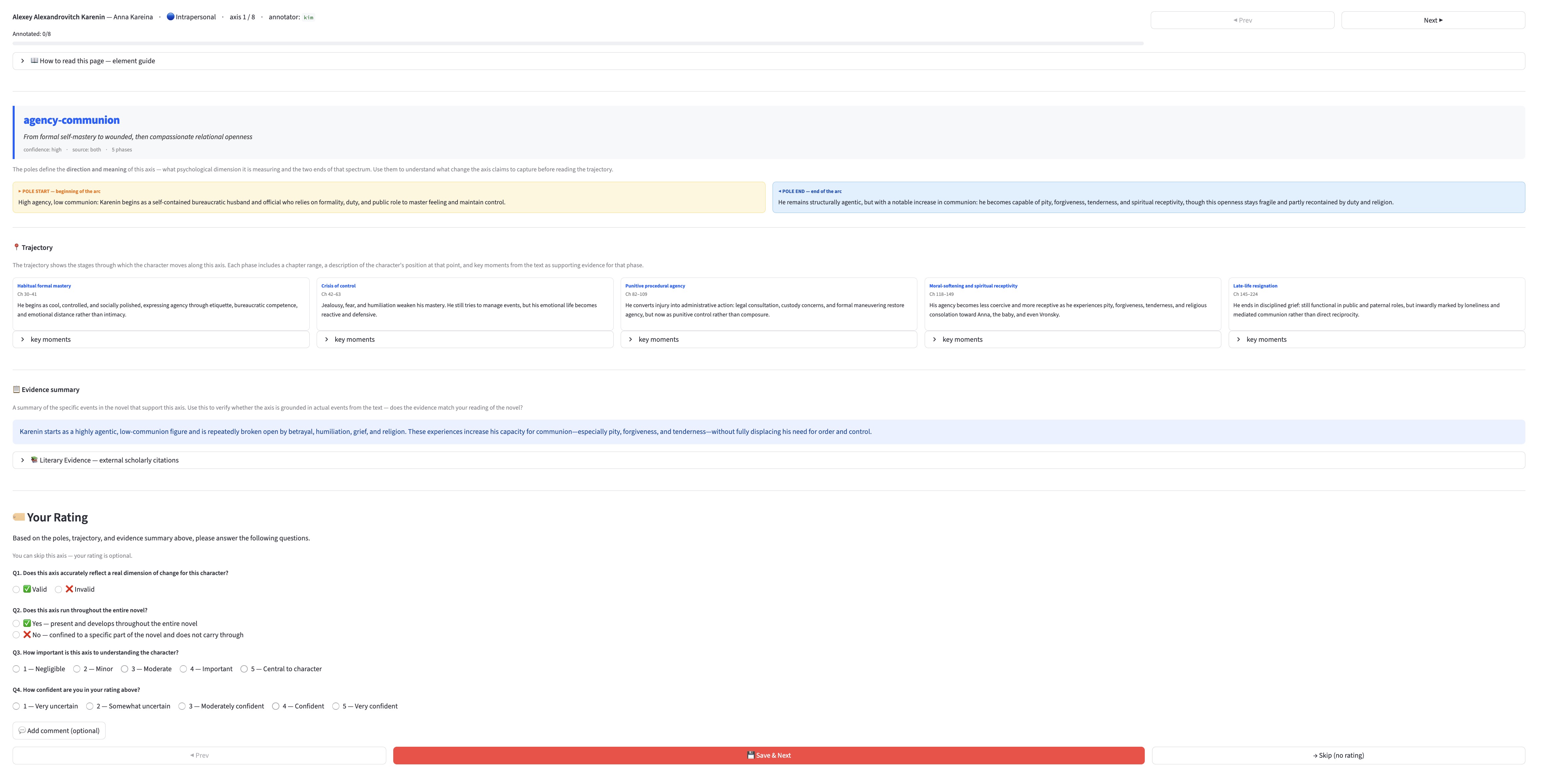}
\caption{Human Annotation Page for axis rating.}
\label{fig:annotation_1}
\end{figure}

\paragraph{Human-annotator ratings.}
Annotators were volunteers who self-reported having read the target novel and were compensated at a fair rate. All participation was on the basis of informed consent. (see Figure~\ref{fig:annotation_1}) Three annotators independently rate each axis on three substantive dimensions (Figure~\ref{fig:pipeline}, top right): \emph{validity} (does the axis describe a real psychological or relational change in the source text?), \emph{novel span} (the proportion of the novel over which the change unfolds), and \emph{importance} (the centrality of the change to the character's portrayal). Each rating is accompanied by a \emph{confidence} self-rating that records the annotator's certainty in their three substantive judgments. Confidence is a meta-rating rather than a property of the axis. Only validity gates evaluation-set entry, with a 2-of-3 majority rule. The remaining ratings are stored as per-axis metadata so analyses can stratify, for example reporting fidelity separately on high- vs.\ low-importance axes. The axis-validity annotation did not constitute human subjects research under our institution's IRB definitions. The task asked annotators to judge whether each extracted character-arc axis is faithful to the source novel. No personally identifiable information, sensitive data, or behavioral data about the annotators themselves was collected.

\paragraph{Validation outcome.}
Axis validation covers all $25$ characters of the validated slice. Three annotators were assigned to each of the $223$ candidate axes, and after excluding unsubmitted ratings each axis carries two or three ratings, for $653$ judgments in total. Annotators saw the character, the proposed poles, the phase trajectory with its key moments, the evidence summary, and the critic panel, and rated validity, novel span, and importance on a $1$--$5$ scale together with a confidence self-rating, with a free-text field for factual errors or alternative readings. $205$ of $223$ axes ($91.9\%$) were retained: $147$ ($65.9\%$) unanimously, $58$ ($26.0\%$) over one dissenting vote, and $18$ ($8.1\%$) rejected. Dissent is one-sided rather than polarized, since axes retained over one dissent outnumber axes rejected with a lone supporting vote by more than three to one and exactly one axis was rejected unanimously. A rejected example is the relational axis from Fernand Mondego to Edmond Dant\`es, which no annotator passed because its description wrongly stated that the two never meet directly.

On the binary validity decision, observed agreement is $0.770$, Fleiss' $\kappa$ is $+0.038$, Gwet's AC1 is $+0.697$, and Brennan--Prediger is $+0.540$. The label pool holds $562$ valid and $91$ invalid ratings, an $86.1\%$ prevalence of the positive class, at which chance agreement under $\kappa$ is $0.761$. $\kappa$ therefore collapses toward zero despite high observed agreement, which reflects the prevalence paradox rather than absent signal. We read the four coefficients together as moderate-to-substantial agreement on an imbalanced task.

\section{Probe Generation Details\label{app:probe-details}}

\paragraph{Out-of-World era menu.}
The era menu has seven slots: \emph{modern urban}, \emph{industrial early 20th century}, \emph{mid-century}, \emph{pre-industrial agrarian}, \emph{speculative near-future}, \emph{pre-modern imperial}, and \emph{diasporic contemporary}. Each slot is broad enough to generalize across source novels yet specific enough to commit the designer to one setting (suppressing a default contemporary-urban collapse seen in early pilots). An (axis, phase) pair is assigned an era by a cyclic walk through the menu starting from a SHA-1-derived offset of the axis id, which guarantees that all $N$ phases of one arc receive distinct eras when $N$ does not exceed the menu size.

\paragraph{Per-(scene phase, probe type) structure.}
For an arc of $N$ phases, the designer is called $3N$ times: once per (scene phase, probe type) cell. The scene phase is the trajectory phase whose lived experience grounds the scenario, and the other $N-1$ phase responses in the same probe are counterfactual projections of those phases' policies onto that scenario. This cell is also the unit at which Stage~iii validators act: per-phase validators score each of the $N$ responses in a cell, and per-probe validators score the cell as a whole.

\paragraph{Q-PhaseFit verdict mapping.}
Q-PhaseFit returns the trajectory phase index $\hat{k}$ that the response is most diagnostic of, plus a confidence label. The wrapper compares $\hat{k}$ to the target phase $k^\star$ and emits one of three verdicts: \texttt{pass} ($\hat{k}=k^\star$), \texttt{adjacent} ($|\hat{k}-k^\star|=1$), or \texttt{off\_phase} ($|\hat{k}-k^\star|\geq 2$). Only \texttt{off\_phase} triggers the per-phase regeneration path. \texttt{adjacent} is recorded but not corrected, since adjacent overlap is theoretically expected when adjacent phases sit close in the state distribution~\cite{Fleeson2001TowardAS}.

\paragraph{Drop and retry policy.}
Post-processing is asymmetric. A per-phase failure (Q-Voice \texttt{fail}, or Q-PhaseFit verdict \texttt{off\_phase}) triggers one regeneration of that response. If the regenerated response also fails, it is marked \texttt{unavailable} and the rest of the probe is kept. A per-probe failure drops the whole probe: Q-Anchor failure for \emph{In-Scenario}, Q-World failure for \emph{In-World} and \emph{Out-of-World}. Q-Discrim is annotation-only and never drops a probe, so adjacent-pair similarity surfaces as a Stage~iii covariate rather than as a filter.

\section{Datapoint information\label{app:datapoint}}

The \ours{} corpus serializes into two record types: a \emph{Character Arc} record per (character, axis), and a \emph{probe} record per (axis, anchor phase, probe category). Probe records carry $N$ \emph{phase responses} as nested objects, one per phase of the parent arc. Both record types live in JSON and share a common key schema across the training and evaluation subsets of Table~\ref{tab:data_stats}; the evaluation subset arcs additionally carry the Stage~iii critic verdicts and the three-annotator validity votes (\S\ref{sec:arc}).

\paragraph{Character Arc record.}
Each arc fixes one axis for one character and is the unit indexed by \texttt{axis\_id}. The header fields are \texttt{character}, \texttt{axis\_type} (\texttt{intrapersonal} or \texttt{relational}), \texttt{dimension\_label} (the McAdams construct family, e.g.\ \emph{Agency--Communion}), \texttt{axis\_name}, \texttt{pole\_start} and \texttt{pole\_end} (one-sentence descriptions of the two poles), and \texttt{arc\_direction} (\emph{redemptive}, \emph{contaminating}, \emph{ambivalent}, etc.; see Appendix~\ref{app:arc-details}). \texttt{source} marks whether the candidate was emitted by the event stream, the state stream, or both (\S\ref{sec:arc}). The \texttt{trajectory} field is an ordered list of $N$ phase objects (\texttt{phase}, \texttt{chapter\_range}, \texttt{position\_description}, \texttt{key\_moments}). For relational arcs, \texttt{target\_character} names the second party. Evaluation subset arcs additionally carry \texttt{literary\_validation}, a triplet of critic verdicts (structuralist, depth-psychological, historical/cultural) each paired with reasoning and candidate citations, and \texttt{annotation}, a per-annotator vote record over \emph{validity}, \emph{novel span}, and \emph{importance}.

\paragraph{Probe record.}
Each probe pairs one (scenario, question) with $N$ phase-keyed reference responses. The header fields are \texttt{probe\_id}, \texttt{probe\_type} ($\in\{$\emph{In-Scenario}, \emph{In-World}, \emph{Out-of-World}$\}$), \texttt{anchor\_phase\_idx} (the trajectory phase whose lived experience grounds the scenario), \texttt{anchor\_query\_chapter} (the chapter at which the anchor response is set, which also caps the knowledge cutoff), \texttt{scenario}, and \texttt{question}. \emph{In-Scenario} probes carry an \texttt{anchor\_source} pointer back to the verbatim passage in the source novel; \emph{Out-of-World} probes carry \texttt{era\_label} together with the \texttt{abstract\_axis} and \texttt{abstract\_phases} reformulations that let the axis transpose into a non-source era (Appendix~\ref{app:probe-details}). The \texttt{phase\_responses} list contains $N$ objects, one per phase, each with \texttt{phase\_idx}, \texttt{phase\_label}, \texttt{query\_chapter}, \texttt{gt\_action} (the overt act), \texttt{gt\_speech} (the in-character utterance, when one is licensed), \texttt{gt\_thought} (one to two sentences of cognitive construal), and \texttt{gt\_typicality} ($\in\{$\emph{typical}, \emph{plausible\_tail}$\}$). Stage~iii validator outputs are carried alongside the responses: per-phase verdicts from Q-Voice and Q-PhaseFit sit inside each phase response under \texttt{quality}, with \texttt{retry\_count} and \texttt{unavailable} recording the drop-and-retry trace (Appendix~\ref{app:probe-details}); per-probe Q-Anchor / Q-World and Q-Discrim verdicts sit at the probe level under \texttt{probe\_check}, and Q-Discrim-flagged adjacent pairs are mirrored on the parent arc under \texttt{weak\_pairs}.

\paragraph{Example.}
Figure~\ref{fig:datapoint_example} shows a single \emph{Out-of-World} probe drawn from Anna Karenina's \emph{Agency--Communion} arc, abbreviated to the first two of $N{=}5$ phase responses. The shared scenario is asked at every phase, and the phase-keyed reference responses encode how the same character would handle that scene at successive points on the axis.

\begin{figure}[t]
\centering
\fbox{\begin{minipage}{0.94\columnwidth}
\scriptsize
\setlength{\parindent}{0pt}
\setlength{\parskip}{2pt}
\texttt{probe\_id}: \texttt{anna\_karenina\_final\_intra\_01\_oow\_a0}

\texttt{probe\_type}: \texttt{out\_of\_world} \quad \texttt{anchor\_phase\_idx}: 0 \quad \texttt{anchor\_query\_chapter}: 20

\texttt{era\_label}: \texttt{pre\_industrial\_agrarian}

\texttt{abstract\_axis}: ``how self-assertion shifts under relational strain from prosocial attunement toward coercive control and ultimately self-negating withdrawal''

\texttt{scenario}: ``At the harvest supper, Anna sees two households quarrel over a promised strip of meadow, and the man whose favor matters most to her stands waiting to see whether she will speak for peace or for herself. The room falls quiet around her.''

\texttt{question}: ``How does Anna answer the silence in the room?''

\hrule\vspace{2pt}

\texttt{phase\_responses[0]} (\emph{Socially fused mediator}, ch.~20)

\quad \texttt{gt\_action}: ``She steps between the benches and proposes that the meadow be shared until the elders measure it after harvest.''

\quad \texttt{gt\_thought}: ``She feels the strain in every face and moves to ease it; if she can give each side a little dignity, the whole gathering may hold together.''

\texttt{phase\_responses[1]} (\emph{Divided self and concealed attachment}, ch.~36)

\quad \texttt{gt\_action}: ``She lowers her eyes and offers a mild compromise, careful not to meet the favored man's gaze for long.''

\quad \texttt{gt\_thought}: ``She wants peace, yet her own hidden leaning makes every word feel stained; if she speaks too warmly, others may hear more in it than the dispute itself.''

\textit{(phases 2--4 omitted for space)}
\end{minipage}}
\caption{One \emph{Out-of-World} probe from Anna Karenina's \emph{Agency--Communion} arc, truncated to the first two of $N{=}5$ phase responses. The same (scenario, question) is asked at every phase, with the references shifting along the axis.}
\label{fig:datapoint_example}
\end{figure}

\begin{table}[t]
\centering
\scriptsize
\setlength{\tabcolsep}{4pt}
\renewcommand{\arraystretch}{0.9}
\setlength{\aboverulesep}{0.2ex}
\setlength{\belowrulesep}{0.2ex}
\begin{tabular}{lrrrr}
\toprule
Split & \#Novels & \#Chars & \#Arcs & \#Probes \\
\midrule
Train (SFT and DPO)    & 12 & 55 & 339 & 2{,}847 \\
Test (validated)            &  5 & 25 &  205 & 1{,}754 \\
Test (unval., low-pop) & 2 & 7 & 26 & 253 \\
\midrule
Total                       & 19 & 87 & 570 & 4{,}854 \\
\bottomrule
\end{tabular}
\caption{Dataset statistics. \emph{Train} uses ungrounded auto arcs. \emph{Test (validated)} uses human-validated arcs (\S\ref{sec:arc}). \emph{Test (unval., low-pop)} uses auto arcs on two low-popularity unvalidated novels as a memorization control~\cite{han-etal-2026-quantifying} (\S\ref{sec:additional}). SFT and DPO data are generated only for Train. Of the $205$ validated arcs, $199$ are probe-bearing and contribute to evaluation.}
\label{tab:data_stats}
\end{table}

\paragraph{Dataset distributions.}
On the validated slice, arcs carry $2$ to $6$ phases (mean $3.51$; counts $13$, $102$, $69$, $15$, $6$ for $2$ to $6$ phases) and split into $93$ intrapersonal and $112$ relational axes, with trajectory directions $124$ positive, $44$ negative, $21$ ambivalent, $12$ flat, and $4$ disillusionment. Of $6{,}567$ phase slots, $637$ ($9.7\%$) are marked \texttt{unavailable} and excluded without imputation. Among available references, final Q-PhaseFit verdicts are $5{,}436$ \texttt{pass} ($91.7\%$) and $494$ \texttt{adjacent} ($8.3\%$), with no \texttt{off\_phase} reference entering evaluation, since such references are either regenerated to \texttt{pass} or \texttt{adjacent} status or marked \texttt{unavailable}. Q-Discrim flags at least one weak adjacent pair on $39.4\%$ of probes, corresponding to $1{,}068$ of $4{,}813$ adjacent pairs ($22.2\%$) by probe type the weak-pair rate is In-Scenario $68.5\%$, In-World $35.4\%$, and Out-of-World $22.8\%$. Dataset-level counts cover the $205$ validated arcs, while response-level statistics use the $199$ probe-bearing axes.

\paragraph{Stratification by Q-Discrim status.}
Q-Discrim is annotation-only, so weak-pair and clean probes carry equal weight in Table~\ref{tab:main_results}. Stratifying the Arc-versus-strongest-non-Arc gap by weak-pair status leaves all $24$ contrasts positive. On the mean of APF, RPF, and RAE the gap runs $+3.7$ to $+11.1$ points on clean probes and $+1.0$ to $+4.3$ on weak-pair probes. The matched PTF gaps run $+3.7$ to $+14.6$ and $+0.3$ to $+8.1$. The smaller margins on weak-pair probes follow from their lower adjacent-phase separability.

\paragraph{Artifact availability and licenses.}
The corpus mixes public-domain and copyrighted source material. All titles other than \emph{Harry Potter} come from Project Gutenberg and are in the U.S. public domain; that source material is identified separately in the release and is not presented as content licensed by us. Our license applies only to annotations we authored: the released \ours{} annotations (character arcs, probes, validator outputs) are distributed under CC-BY-4.0, and the accompanying code under MIT. \emph{Harry Potter} is not a Gutenberg text, so its public export carries no source text, verbatim passage, source quotation, or recoverable excerpt. We therefore omit all \emph{Harry Potter} artifacts from the public release rather than place them under our data license, so the released evaluation slice covers four of the five validated novels. The benchmark remains reproducible on the remaining titles, and the released code allows the \emph{Harry Potter} portion to be rebuilt from a legally obtained copy. The open-weight models used for inference and fine-tuning (e.g., Qwen3, DeepSeek-V4 family) are used under their respective publishers' licenses, and the proprietary models accessed via API (\S\ref{sec:setup}, Appendix~\ref{app:judge}) are used under the providers' terms of service. The released \ours{} artifacts are intended for non-commercial research on context-grounded role-playing language agents, consistent with the original access conditions of the source materials.

\section{Experimental Details}
\label{app:training_details}

\subsection{Datasets}
\label{app:datasets_details}

This section provides details for the datasets. Both SFT and DPO datasets are derived from the \ours{} corpus and share the same underlying hierarchy, differing only in how they are formulated.

\paragraph{SFT dataset.}
SFT data is generated on the 12 training novels, covering $55$ characters and $339$ axes. For each (probe, phase) pair under \textsc{Arc} context, we sample three completions from \texttt{gpt-5.4-mini} (OpenAI) with $T=0.9$ and a 800-token cap. The teacher conditions on the phase reference privately, and the stored row keeps only the character system prompt, the scenario--question user prompt, and the answer, so no reference text leaks into the training input.

\paragraph{DPO dataset.}
Preference pairs are constructed automatically, spanning 55 characters across 12 novels, organized along $339$ axes and 2{,}516 unique probes in \ours{} corpus. For each probe, we designate the anchor phase response as chosen and a response from an adjacent phase as rejected, with both sides sharing the same scenario and question. This ensures that the only contrast between chosen and rejected is the character's behavioral disposition at different developmental stages, rather than differences in scenario or context. The resulting training set contains 14{,}671 preference pairs across 2{,}516 probes.

\paragraph{Training and evaluation overlap.}
Table~\ref{tab:overlap-matrix} states the audited overlap between each evaluation slice and the training pool. The validated slice that carries the main comparison of Table~\ref{tab:main_results} is disjoint from it. \emph{East Lynne} and \emph{The Underdogs} are themselves training novels, so the \ours{} results on them are in-distribution rather than held out and are retained for reference only (Appendix~\ref{app:extra3-results}). The low-popularity comparison of \S\ref{sec:additional} runs instead on two novels with no training overlap. Per-stage novel, character, arc, and probe identifiers are released with the data.

\begin{table}[!t]
\centering
\scriptsize
\setlength{\tabcolsep}{5pt}
\renewcommand{\arraystretch}{1.05}
\begin{tabular}{lcl}
\toprule
Slice & Training-pool overlap & Status \\
\midrule
Validated ($5$ novels) & $0/0/0/0$ & Held out \\
Low-pop, replacement & $0/0/0/0$ & Held out \\
\emph{East Lynne}, \emph{Underdogs} & $2/7/31/301$ & Training novels \\
\bottomrule
\end{tabular}
\caption{Overlap with the training pool as novels, characters, arcs, and evaluable probes. The SFT and DPO stages use the same twelve training novels.}
\label{tab:overlap-matrix}
\end{table}

\subsection{Two-Stage Pipeline}
\label{app:training_pipeline}
This section provides details for the two-stage training pipeline described in~\S\ref{sec:training_rpla}, applied to both \ours{}-8B and \ours{}-32B. The two stages are designed to be complementary. SFT instills the form and texture of \ours{} corpus, and the DPO stage sharpens the model's sensitivity to developmental trajectory by contrasting anchor phase behavior against plausible adjacent phase alternatives.

\paragraph{Stage 1: Supervised fine-tuning.} In the SFT stage, the model learns to produce responses that conform to the format and behavioral conventions of \ours{}, which differ substantially from standard instruction-following. Generations from \texttt{gpt-5.4-mini} serve as imitation targets.
 
\paragraph{Stage 2: Direct preference optimization.} In the DPO stage, the model learns to distinguish ground-truth character behavior from plausible but temporally displaced alternatives, namely, actions or thoughts that fit the character but belong to a different narrative phase.

\subsection{Training Configuration}
\label{app:training_model}

This section provides the training configuration. Table~\ref{tab:training_config} summarizes the training configurations for both \ours{}-8B and \ours{}-32B models across the two-stage pipeline.

\begin{table}[h]
\centering
\small
\begin{tabularx}{\columnwidth}{lcc}
\toprule
& \textbf{\ours{}-8B} & \textbf{\ours{}-32B} \\
\midrule
Base model          & Qwen3-8B  & Qwen3-32B \\
Fine-tuning         & Full      & LoRA \\
\midrule
\multicolumn{3}{@{}c@{}}{\makebox[\columnwidth][c]{\textit{SFT}}} \\
\midrule
Epochs & 1 & 1  \\
Learning rate & 1e-5 & 1e-4 \\
Effective batch size & 64 & 32 \\
Max length & 8192 & 8192 \\
Training time & 2h 38m & 9h 49m \\
\midrule
\multicolumn{3}{@{}c@{}}{\makebox[\columnwidth][c]{\textit{DPO}}} \\
\midrule
Epochs & 1 & 1 \\
Learning rate & 5e-6 & 1e-5 \\
Batch size & 64 & 64 \\
Max length & 8192 & 8192 \\
Training time & 3h 44m & 14h 3m  \\
\midrule
\multicolumn{3}{@{}c@{}}{\makebox[\columnwidth][c]{\textit{RLVR (GRPO)}}} \\
\midrule
Steps / epochs & --- & 86 / 2 \\
Learning rate & --- & 1e-5 \\
KL coef.\ $\beta$ / clip $\varepsilon$ & --- & 0.001 / 0.2 \\
Train / mini batch & --- & 64 / 32 \\
Rollouts per prompt $G$ & --- & 8 \\
Prompt / response len & --- & 8192 / 2048 \\
\midrule
\multicolumn{3}{@{}c@{}}{\makebox[\columnwidth][c]{\textit{LoRA}}} \\
\midrule
Rank ($r$)          & --- & 64  \\
Alpha ($\alpha$)    & --- & 128  \\
\midrule
GPU                 & \multicolumn{2}{c}{1 $\times$ NVIDIA B200 (192GB)} \\
\bottomrule
\end{tabularx}
\caption{Training configuration for \ours{}-8B and \ours{}-32B, following the official Qwen3 sampling recommendations for non-thinking mode. The RLVR stage is applied to \ours{}-32B only, initialized from its DPO checkpoint.}
\label{tab:training_config}
\end{table}

\subsection{RLVR Stage\label{app:rlvr}}
This section details the reinforcement learning with verifiable-reward stage of \S\ref{sec:rlvr}, applied to \ours{}-32B only; there is no $8$B counterpart. We run GRPO~\citep{shao2024deepseekmathpushinglimitsmathematical} with \texttt{verl}~\citep{sheng2025hybridflow} and \texttt{vLLM} rollouts, optimizing with AdamW, on two B200 GPUs for the actor and rollout plus one B200 serving the Qwen3.6-27B reward judge. Hyperparameters are in the RLVR block of Table~\ref{tab:training_config}.

\paragraph{Prompts.} We reuse the SFT and DPO prompt distribution, keep the prompts where the character sits at the probe's anchor phase, deduplicate on the (system, user) message pair, and drop the assistant turn so GRPO generates fresh rollouts. This yields $2{,}812$ training and $757$ validation prompts over the same $12$ novels as the DPO preference pool, so the overlap audit of Appendix~\ref{app:datasets_details} applies unchanged and the validated slice remains disjoint from the RLVR pool.

\paragraph{Reward.} The reward judge (Qwen3.6-27B) is deliberately different from the evaluation judge (DeepSeek-V4-Flash), and it scores rollouts under the unchanged Appendix~\ref{app:judge} prompt.

\paragraph{Scoring.} Evaluation reuses the validated slice under \textsc{Arc} context only, so the comparison isolates the training stage. The five systems are scored on the (novel, character, probe, phase) records they share, aggregated by the rule of Appendix~\ref{app:judge}. Under the evaluation judge the base and DeepSeek-V4-Pro sweeps lack \emph{Benjamin Franklin}, so that judge covers four of the five validated novels; the two cross-judges cover all five, which is why the RLVR-versus-DPO comparison of \S\ref{sec:rlvr} spans $14$ rather than $15$ novel-by-judge cells.

\begin{table}[!t]
\centering
\small
\setlength{\tabcolsep}{4pt}
\renewcommand{\arraystretch}{1.0}
\begin{tabular}{llccc}
\toprule
 & & \multicolumn{3}{c}{Judge} \\
\cmidrule(lr){3-5}
 & & \makecell{DS-V4-Flash\\(eval)} & \makecell{Qwen3.5} & \makecell{Qwen3.6\\(reward)} \\
\midrule
\multirow{3}{*}{Overall} & DPO & $60.2$ & $50.3$ & $45.5$ \\
  & RLVR & $68.2$ & $56.0$ & $50.6$ \\
  & $\Delta$ & $\mathbf{+8.0}$ & $\mathbf{+5.7}$ & $\mathbf{+5.2}$ \\
\midrule
\multirow{3}{*}{Rewarded} & DPO & $62.3$ & $51.3$ & $49.6$ \\
  & RLVR & $71.1$ & $60.6$ & $58.4$ \\
  & $\Delta$ & $\mathbf{+8.8}$ & $\mathbf{+9.3}$ & $\mathbf{+8.8}$ \\
\midrule
\multirow{3}{*}{PTF} & DPO & $53.7$ & $47.1$ & $33.1$ \\
  & RLVR & $59.4$ & $42.2$ & $27.3$ \\
  & $\Delta$ & $\mathbf{+5.7}$ & $\mathbf{-5.0}$ & $\mathbf{-5.8}$ \\
\bottomrule
\end{tabular}
\caption{DPO to RLVR under three judges. \emph{Rewarded} is the mean of APF, RPF, and RAE, the three metrics the reward optimizes; PTF is held out of it. The rewarded metrics rise by a similar amount under every judge, including the one that supplied the reward, while PTF rises only under the evaluation judge. Absolute levels differ across judges, so values are comparable down a column but not across columns.}
\label{tab:rlvr-crossjudge}
\end{table}

\paragraph{Why the category asymmetry.} DPO sees only a binary contrast between the anchor phase and an adjacent one, so it teaches separation between neighboring phases. Adjacent phases legitimately overlap most on In-Scenario, where a source scene constrains the available action and $68.5\%$ of probes carry a Q-Discrim weak pair (Appendix~\ref{app:datapoint}), so pushing the phases apart there costs fidelity. A graded per-phase reward instead scores each response against its own phase reference, which lets it credit the right answer where two phases genuinely coincide.

\paragraph{\ours{} as an RL environment.} The same rubric that scores the benchmark supplies a dense reward and the probe supplies the state, so any arc-grounded corpus built with the \S\ref{sec:ArcANE} pipeline yields an RL environment at no extra annotation cost.

\subsection{Inference Details}
\label{app:inference_details}

This section describes the inference setup used for all models evaluated in \S\ref{sec:experiments}.

\paragraph{Inference setup and cost.}
Open-weight models (Qwen3-8B, Qwen3-32B, \ours{}-8B, and \ours{}-32B) are served locally via \texttt{vLLM} on a single NVIDIA B200 (192GB) GPU, following the official Qwen3 sampling recommendations for non-thinking mode. Generation is capped at 8{,}192 tokens, set at the 99th percentile of observed response lengths, to guard against runaway generations. Covering all novels and modes, inference takes up to 6 hours per model.

Open-weight large models (DeepSeek-V4-Flash and DeepSeek-V4-Pro) are called through \texttt{OpenRouter} with the provider pinned to a fixed pair (\texttt{DeepSeek,Alibaba}, \texttt{allow\_fallbacks=false}) so that role-playing calls cannot silently re-route to a different quantization. 

Role-play generation passes no explicit decoding parameters; each backend's default sampling applies (effectively temperature~$1.0$ with no top-$p$ or repetition-penalty restriction across vLLM, OpenAI, and OpenRouter). One sample is drawn per (probe, mode, model). Two subcalls override this for determinism: the TimeCHARA two-stage hint extractor and the LLM judge (\S\ref{sec:evalproto}) both run at $T=0$.

\paragraph{Construction cost.}
The cost that matters for extending the pipeline to new novels is construction rather than training. Saved artifacts support auditable API-call counts but not token or dollar totals, since historical logs do not stamp per-call usage. Arc construction over the $17$ novels of the original corpus used at least $14{,}096$ calls to \texttt{gpt-5.4-mini}, a lower bound from persisted intermediate directories and critic-stage candidates that excludes unlogged retries. The probe pipeline is fully logged for two novels, where $14{,}452$ calls to \texttt{gpt-5.4} produced $764$ retained probes across generation, reference construction, retries, and validation. At that rate of $18.9$ calls per retained probe, the probe stage over the $4{,}601$ probes of that corpus comes to roughly $87{,}034$ calls.

\section{Context-Mode Templates\label{app:contextmodes}}

The six context modes evaluated in \S\ref{sec:setup} share the same (scenario, question, query chapter) prompt body and differ only in the system-side context block.

\paragraph{Vanilla.}
The system message states the character's identity and the query chapter, and no further context is supplied. Vanilla is the reference for how much the LLM can do from its own world knowledge.

\paragraph{Summary.}
Vanilla plus per-chapter summaries for the most recent \texttt{SUMMARY\_MAX\_CHAPTERS}~$=5$ chapters up to and including the query chapter. The summaries are produced once per novel by the \texttt{summary} pipeline of \texttt{experiments/}.

\paragraph{RAG.}
Vanilla plus the top-$k$ source-text chunks retrieved with the (scenario, question) string as the embedding query. Chunks are 1.5k-character windows with 300-character overlap, embedded with \texttt{text-embedding-3-small}, and retrieval is masked to chunks up to the query chapter, with $k=6$.

\paragraph{LifeChoice.}
A port of CHARMAP~\cite{xu-etal-2025-characterisdestiny}. The \emph{description} is the chapter-summary stack (the same source as Summary), and the \emph{memory} is a top-$k$ retrieval over the RAG index using the description as the embedding query (CHARMAP's contribution over plain RAG).

\paragraph{TimeCHARA.}
A port of the narrative-experts pipeline of \citet{ahn-etal-2024-timechara}: a first call predicts the query chapter and the character's presence in it, and the second call uses those predictions as inline hints to the role-play model. No system-side context block is added.

\paragraph{\textsc{Arc} (ours).}
Vanilla plus the curated character arc up to the query chapter. The payload is the final-stage axis JSON with phases beyond the query chapter hidden. \texttt{literary\_validation} and \texttt{evidence\_summary} are always stripped (the former is the human-rater bibliography, the latter summarizes the whole arc and would leak future state). When at least one later phase is hidden, \texttt{pole\_end} and \texttt{arc\_direction} are also removed so the truncated payload cannot reveal where the trajectory is heading.

\section{Judge Setup and Rubrics\label{app:judge}}

\paragraph{Judge model and serving.}
The default judge is \texttt{deepseek/deepseek-v4-flash} served through OpenRouter, with the provider pinned to a fixed pair (\texttt{DeepSeek,Alibaba}) and \texttt{allow\_fallbacks=false}, so judge calls cannot silently re-route to a different provider or quantization. The reasoning pass is disabled (\texttt{EXP\_JUDGE\_DISABLE\_REASONING=true}) because each verdict is a small JSON object and the thinking pass otherwise dominates the bill ($\sim$57$\times$ the static prompt). Each run uses a 200-thread judge pool, with per-score scale $1$--$100$. The judge sees both \texttt{ref\_action} and \texttt{ref\_thought} by default. A no-thought variant (\texttt{INCLUDE\_THOUGHT=false}) instructs the judge to infer the reference reasoning from \texttt{ref\_action} and \texttt{ref\_speech} alone, used as an ablation of the rubric's dependence on \texttt{ref\_thought}.

\paragraph{Per-phase rubric (APF / RPF / RAE).}
Each (probe, phase) response is scored on three sub-dimensions. \emph{APF} judges whether the response's action is mechanism-equivalent to \texttt{ref\_action} at the target phase, with three salvage levels (A strict, B and C giving partial credit when A fails). \emph{RPF} parses \texttt{ref\_thought} and the response's reasoning into four mechanism slots (appraisal, goal, strategy, trigger) and judges each slot, then aggregates. \emph{RAE} fixes \texttt{ref\_thought} as the operative reasoning and asks whether the response's action is one this reasoning would license. \texttt{gt\_entailment} is the dominant sub-check, so a response with internally consistent but \texttt{ref\_thought}-incompatible reasoning still scores low.

\paragraph{Trajectory rubric (PTF).}
The trajectory judge sees the scenario, the question, and $N$ phase blocks in chronological order. Each block carries the reference fields and the model's response at that phase. PTF aggregates three sub-scores (alignment, direction, shape), each on the $1$--$100$ scale, designed so a model that scores well on every phase in isolation can still earn a low PTF by collapsing adjacent phases, traversing the axis in the wrong direction, or compressing the shape of the change. \texttt{unavailable} phase responses are dropped from their blocks, but the original phase indices are preserved so the judge can see the gap.

\paragraph{Aggregation rule.}
APF, RPF, and RAE are pooled over the available (probe, phase) records within each novel, and PTF is scored once per eligible trajectory. Novel means are then averaged with equal weight, and Overall is the simple mean of the twelve probe-type by metric cells. Missing, unparsable, and \texttt{unavailable} records are dropped without imputation, and a trajectory needs at least two available phases to be scored. Q-Discrim is annotation-only, so probes with weakly separated adjacent pairs carry the same weight as the rest. The judge treats each reference as an anchor rather than a target string, granting partial credit to responses that reach the reference behavior by a different route.

The judge-side no-thought ablation swaps the two system prompts for variants that drop the \texttt{ref\_thought} mention and instruct the judge to infer the reference reasoning from \texttt{ref\_action} and \texttt{ref\_speech} alone. The sub-check definitions are unchanged.

\section{Reward-Hacking Falsification\label{app:pairwise-coding}}

The reward-hacking probe in \S\ref{sec:error-analysis} runs two structured pair-trace procedures on the same stratified $150$-probe sample: a four-proposition coding test and a POV-control regeneration. The four propositions were not chosen \emph{a priori}. An author close-reading of six paired trajectories surfaced them first. We describe the close-reading, then the proposition test, then the POV control, then the joint result.

\paragraph{Coder configuration.} The proposition test uses \texttt{deepseek/deepseek-v4-pro} on OpenRouter (provider pinned to \texttt{DeepSeek}, \texttt{allow\_fallbacks=false}, temperature $0$, \texttt{json\_object} response format, $16$--$32$-thread \texttt{ThreadPoolExecutor}). We use a pro-tier coder rather than the \emph{flash}-tier judge that scores APF/RPF/RAE/PTF (Appendix~\ref{app:judge}) because each call reads two full per-phase response sets ($\sim$$2$--$5$ phases per side, $\sim$$5$--$15$K input tokens) and emits a structured verdict.

\paragraph{Sampling.}
The pair-trace pool is the validated slice ($1{,}750$ probes, $4$ dropped for missing judge scores). For each probe we compute per-probe Overall under Arc as the mean over \{\textsc{apf}, \textsc{rpf}, \textsc{rae}, \textsc{ptf}\}, with \textsc{apf}/\textsc{rpf}/\textsc{rae} averaged across the probe's phase responses and \textsc{ptf} the mean of the three trajectory sub-scores. Per-probe deltas are tagged \emph{positive} ($\Delta > 1$), \emph{tie} ($|\Delta| \leq 1$), or \emph{negative} ($\Delta < -1$) on the $0$--$100$ scale, and ties are dropped from coding. We stratify by (probe-type, sign) into six cells and draw $25$ per cell ($150$ probes total, Qwen3-32B vs.\ \ours{}-DPO) for both the proposition test and the POV control.

\paragraph{Close-reading six paired trajectories.}
We close-read six probes spanning all three categories and both delta signs to surface candidate patterns: Dulcinea silent under courtly mockery (In-Scenario, relational), Don Quixote's volunteer-patrol arc (Out-of-World, intrapersonal), Sans\'{o}n Carrasco listening to Sancho (Out-of-World, relational), Darya Oblonskaya through a domestic-crisis cascade (In-World, intrapersonal), Snape after the Lily slur (In-Scenario, relational), and Karenin handling a clerk's wife in public (In-World, intrapersonal). Four patterns recurred across all six.

\begin{itemize}
\setlength{\itemsep}{1pt}
\item \textbf{P1, Phase distinctness.} Per-phase content is substantively different across phases versus paraphrase. \ours{} writes different content per phase. Qwen3-32B often paraphrases: on Carrasco--Sancho it opens \emph{every} phase with ``\emph{Ah, Sancho Panza, you speak with the boldness of a man who has long since learned}\ldots''. \ours{} moves him from amused engagement (``\emph{Ha, Sancho, you speak as if you had a warrant in your sleeve!}'') to practical incorporation (``\emph{I took it as a thing to be weighed, not mocked}'').
\item \textbf{P2, Canonical specificity.} The response invokes named secondary characters, named places, or source-specific scenes, versus generic trait language. \ours{}'s Snape phase 1 routes through Dumbledore and the memory transfer. Qwen3-32B stays at a general lament (``\emph{I would have chosen you, Lily}'').
\item \textbf{P3, Negative-space fidelity.} When canon calls for silence, proxy, or absence, the response preserves it. Both Qwen3-32B and \ours{} pass Dulcinea: Qwen3-32B by third-person narration, \ours{} by routing through canon's Merlin pageant. POV-Qwen (the register-controlled Qwen run, below) fails this case, inventing speech the canonical scene refuses.
\item \textbf{P4, Phase register switch.} Voice and register shift to match each phase's psychological description. \ours{}'s Darya opens phase 0 (``Emotional collapse'') with ``\emph{I feel my heart give a violent jump}'' and closes phase 3 (``Late steady endurance'') with ``\emph{first care, then order, then the business of the day}''. Qwen3-32B keeps one composed-narrator register throughout.
\end{itemize}

\paragraph{Proposition test.}
For each of $150$ probes $\times$ $3$ model conditions ($450$ calls) the coder reads axis, scenario, question, reference fields (\texttt{ref\_action}, \texttt{ref\_thought}), and one model's per-phase response set, and returns yes/no on each of P1--P4 with a one-sentence reason. P3 is tagged \texttt{na} when canon does not call for silence or proxy. The three conditions are: Qwen3-32B with the original system prompt, Qwen3-32B with one POV instruction appended (POV-Qwen, defined next), and \ours{}-32B-DPO.

\paragraph{POV-control regeneration.}
For each (probe, phase) in the $150$-probe sample we load the byte-identical \texttt{prompt\_system} and \texttt{prompt\_user} from the main paper's arc run and append one paragraph:

\begin{quote}
\emph{Respond IN CHARACTER, in the first person, in the present tense. Speak AS this character, not ABOUT them. Do NOT narrate from the outside; do not write `X would respond' or `X's manner is one of'; speak as the character. Use period-appropriate diction natural to the source novel. Do not mention `axes', `phases', or any meta-analytic framing.}
\end{quote}

Calls go to \texttt{qwen/qwen3-32b} on OpenRouter pinned to the Alibaba provider with thinking disabled, the same backend used in the main runs. The resulting $509/510$ phase responses are scored with the unchanged Stage-iii APF/RPF/RAE judge (\texttt{deepseek/deepseek-v4-flash}), prompts and provider settings identical to Appendix~\ref{app:judge}.

\paragraph{Results.}
Table~\ref{tab:propositions} reports proposition yes-rates by category. The Overall row collapses the three categories. Per-category rows are diagnostic. Bold marks the highest \emph{when the proposition is category-appropriate}. For P2, ``yes'' on In-World and Out-of-World by POV-Qwen is largely context-violating canon-drop (e.g.\ Voldemort name-dropped into a city transit hub), not honest specificity, so we do not bold those cells. P3 is applicable on only $2$ probes in this stratified sample (the silence-or-proxy structure is rare on the validated slice), so we report it qualitatively above.

\begin{table}[!t]
\centering
\small
\setlength{\tabcolsep}{4pt}
\resizebox{\columnwidth}{!}{%
\begin{tabular}{lrrrr}
\toprule
Proposition & Category & Qwen3-32B & POV-Qwen & \ours{}-DPO \\
\midrule
\multirow{4}{*}{P1 Phase distinctness}
        & In-Scenario   &  4.5 & 61.4 & 53.3 \\
        & In-World      & 22.0 & 70.0 & \textbf{74.0} \\
        & Out-of-World  &  4.1 & 55.1 & \textbf{67.3} \\
        & Overall       & 10.5 & 62.2 & \textbf{65.3} \\
\midrule
\multirow{4}{*}{P2 Canonical specificity}
        & In-Scenario   & 20.0 & 68.0 & \textbf{34.0} \\
        & In-World      &  8.0 & 60.0 &  2.0 \\
        & Out-of-World  &  4.0 & 54.0 &  2.0 \\
        & Overall       & 10.7 & 60.7 & 12.7 \\
\midrule
\multirow{4}{*}{P4 Phase register switch}
        & In-Scenario   &  6.8 & 61.4 & 53.3 \\
        & In-World      & 22.0 & 72.0 & \textbf{76.0} \\
        & Out-of-World  &  4.1 & 57.1 & \textbf{65.3} \\
        & Overall       & 11.2 & 63.6 & \textbf{65.3} \\
\bottomrule
\end{tabular}}
\caption{Proposition yes-rates ($\%$, excluding \texttt{na}) on $n{=}150$ probes per cell, judge \texttt{deepseek-v4-pro}. Per-category cells $n{\approx}50$.}
\label{tab:propositions}
\end{table}

\paragraph{Rubric scores and a representative case.}
On the same $150$ probes, the Stage-iii APF/RPF/RAE rubric judge scores Qwen3-32B $53.8$, POV-Qwen $50.0$, and \ours{}-32B-DPO $56.7$ (per-probe Overall, $0$--$100$). The POV instruction lowers the rubric score by penalizing context-violating canon-drop. The Dulcinea phase-1 probe (``\emph{Theatrical prop and staged manipulation}'') is representative. Qwen3-32B writes ``\emph{Dulcinea\ldots does not speak, does not interject}'' in third person. \ours{} routes through canon's Merlin pageant (``\emph{Dulcinea answered in the person of another\ldots a squire in a mantle of green}''). POV-Qwen, forced into first person, invents a speech canon refuses (``\emph{`Beauty, that fleeting muse of mortal eyes\ldots'}''). Rubric scores: $80$, $40$, $22$ respectively. The same pattern, POV-Qwen losing on fabrication while \ours{}-32B-DPO stays anchored, recurs across the high-drop tail.

\paragraph{Note on a discarded discovery pass.}
Before the close-reading above, we ran a free-form LLM coding pass on a $72$-probe sample ($8$ per (probe-type, sign) cell from Qwen3-32B vs.\ \ours{}-DPO, $4$ per cell from \ours{}-SFT vs.\ \ours{}-DPO). The coder returned per-probe \texttt{summary}, \texttt{trajectory\_shape}, $2$--$4$ \texttt{concrete\_tells}, a \texttt{substance\_or\_style} tag, and a $4$--$10$ word \texttt{category\_label}. A second LLM call clustered the observations into themes. The themes confounded POV style with substantive arc movement, so we do not use them in the main paper. The author close-reading above replaced the failed clustering.

\section{SFT vs DPO Close-Coding\label{app:sft-vs-dpo}}

What does DPO add over SFT (\S\ref{sec:error-analysis})? The \ours{}-SFT vs.\ \ours{}-DPO pair-trace covers $1{,}750$ probes (mean $\Delta{=}{+}1.89$, $N_{+}{=}940$, $N_{-}{=}702$). We close-code $180$ of them ($30$ per (probe-type, sign) cell) using the same free-form LLM coder as the discarded discovery pass in Appendix~\ref{app:pairwise-coding}. On the SFT$\to$DPO pair the resulting themes are coherent rather than confounded, dominated by \emph{phase-driven arc vs.\ static register}: \ours{}-DPO sharpens what \ours{}-SFT already started.

\paragraph{Two illustrative cases.}
On Don Quixote's interdependence axis, \ours{}-SFT keeps a single knightly tag (``\emph{Stand aside, good people, I go in service of order}'') through every phase. \ours{}-DPO drops the knightly voice in the repentant phase and turns toward family (``\emph{You need not fear for me\ldots come home, and let us settle what ought to be settled in peace}''). On Voldemort, \ours{}-SFT keeps him stoic across phases (``\emph{I do not falter}''). \ours{}-DPO finds the fracture (``\emph{`No!' I snarled, and the word itself seemed to recoil from me}'').

\paragraph{One regression.}
On Hagrid the \ours{}-DPO compression hurts. Hagrid's idiom is communal, dramatically expansive, and multi-speaker, and it sits more naturally in Qwen3-32B's third-person scene narration than in \ours{}-DPO's first-person interior monologue. This is the only character in the $180$-probe sample where the first-person register \ours{} installs is a net loss.

\section{Additional Novels: Full Results\label{app:extra3-results}}

Table~\ref{tab:main_results_lowpop} reports the low-popularity slice used in \S\ref{sec:additional}, namely \emph{He Knew He Was Right} and \emph{The Odd Women}, covering $7$ characters, $26$ axes, and $253$ probes with no overlap with the SFT or DPO pools. Cells pool observations within a novel and then weight the two novels equally; Overall is the mean of the twelve probe-type by metric cells. Table~\ref{tab:main_results_extra3} reports \emph{The Underdogs} and \emph{East Lynne}, which are training novels (Appendix~\ref{app:datasets_details}) and are retained for reference only. In both tables, probes are filtered automatically by the same Stage-iii validators as the main slice but are not annotator-audited.

\begin{table}[!t]
\centering
\scriptsize
\setlength{\tabcolsep}{2.6pt}
\renewcommand{\arraystretch}{1.0}
\begin{tabular}{lrrrrrrr}
\toprule
\multicolumn{8}{c}{Overall by context mode} \\
\midrule
Model & Van. & Sum. & RAG & LifeC. & TimeC. & Arc & Gap \\
\midrule
DeepSeek-V4-Flash & 51.59 & 52.91 & 52.50 & 53.70 & 51.86 & \textbf{60.32} & $+6.62$ \\
DeepSeek-V4-Pro   & 49.61 & 53.15 & 51.14 & 53.27 & 50.26 & \textbf{60.56} & $+7.29$ \\
Qwen3-8B          & 35.02 & 39.30 & 38.78 & 39.68 & 35.75 & \textbf{43.49} & $+3.81$ \\
Qwen3-32B         & 39.47 & 47.14 & 46.14 & 47.62 & 40.30 & \textbf{51.50} & $+3.87$ \\
\ours{}-8B-DPO    & 40.82 & 45.48 & 46.69 & 46.62 & 40.05 & \textbf{59.23} & $+12.54$ \\
\ours{}-32B-DPO   & 46.69 & 49.43 & 48.42 & 52.88 & 46.17 & \textbf{65.21} & $+12.33$ \\
\midrule
\multicolumn{8}{c}{Arc lift by probe type} \\
\midrule
Model & \multicolumn{2}{c}{In-Scen.} & \multicolumn{2}{c}{In-World} & \multicolumn{3}{c}{Out-of-World} \\
\midrule
DeepSeek-V4-Flash & \multicolumn{2}{c}{$+2.58$} & \multicolumn{2}{c}{$+8.64$} & \multicolumn{3}{c}{$+6.41$} \\
DeepSeek-V4-Pro   & \multicolumn{2}{c}{$+3.61$} & \multicolumn{2}{c}{$+8.78$} & \multicolumn{3}{c}{$+7.44$} \\
Qwen3-8B          & \multicolumn{2}{c}{$+1.40$} & \multicolumn{2}{c}{$+3.77$} & \multicolumn{3}{c}{$+4.88$} \\
Qwen3-32B         & \multicolumn{2}{c}{$-1.15$} & \multicolumn{2}{c}{$+4.62$} & \multicolumn{3}{c}{$+7.09$} \\
\ours{}-8B-DPO    & \multicolumn{2}{c}{$+1.66$} & \multicolumn{2}{c}{$+15.52$} & \multicolumn{3}{c}{$+16.54$} \\
\ours{}-32B-DPO   & \multicolumn{2}{c}{$-1.04$} & \multicolumn{2}{c}{$+16.30$} & \multicolumn{3}{c}{$+19.56$} \\
\bottomrule
\end{tabular}
\caption{Low-popularity slice with no training overlap (\emph{He Knew He Was Right}, \emph{The Odd Women}). Gap and lift are Arc minus the strongest non-Arc mode, taken over Overall in the top block and within each probe type in the bottom block, with the baseline reselected per cell.}
\label{tab:main_results_lowpop}
\end{table}

\begin{table*}[t]
\centering
\scriptsize
\setlength{\tabcolsep}{5pt}
\renewcommand{\arraystretch}{0.85}
\setlength{\aboverulesep}{0.2ex}
\setlength{\belowrulesep}{0.2ex}
\begin{tabular}{lllccccccccccccc}
\toprule
Family & Size & Mode & \multicolumn{4}{c}{In-Scenario} & \multicolumn{4}{c}{In-world} & \multicolumn{4}{c}{Out-of-world} & Overall \\
\cmidrule(lr){4-7} \cmidrule(lr){8-11} \cmidrule(lr){12-15} \cmidrule(lr){16-16}
 & & & APF & RPF & RAE & PTF & APF & RPF & RAE & PTF & APF & RPF & RAE & PTF & Overall \\
\midrule
\multirow[t]{12}{*}{DeepSeek-V4} & \multirow[t]{6}{*}{Flash} & Vanilla & 56.9 & 57.7 & 47.9 & 51.3 & 55.6 & 56.6 & 47.9 & 47.1 & 52.4 & 53.4 & 43.7 & 47.8 & 51.5 \\
 &  & Summary & 60.6 & 60.2 & 52.8 & 56.9 & 54.5 & 55.5 & 45.9 & 48.6 & 51.9 & 52.9 & 42.4 & 46.4 & 52.4 \\
 &  & RAG & 62.7 & 61.9 & \textbf{55.6} & 59.7 & 54.0 & 55.4 & 44.9 & 47.5 & 53.2 & 53.8 & 43.5 & 48.7 & 53.4 \\
 &  & LifeChoice & 61.0 & 60.5 & 53.3 & 52.8 & 55.7 & 57.0 & 46.9 & 48.4 & 51.9 & 53.2 & 42.7 & 46.8 & 52.5 \\
 &  & TimeChara & 56.6 & 57.3 & 48.6 & 51.8 & 53.1 & 54.2 & 45.1 & 43.1 & 53.8 & 54.8 & 45.2 & 45.8 & 50.8 \\
 &  & \cellcolor{yellow!15} Arc & \cellcolor{yellow!15} \textbf{63.4} & \cellcolor{yellow!15} \textbf{63.0} & \cellcolor{yellow!15} 55.3 & \cellcolor{yellow!15} \textbf{60.9} & \cellcolor{yellow!15} \textbf{62.8} & \cellcolor{yellow!15} \textbf{62.4} & \cellcolor{yellow!15} \textbf{54.9} & \cellcolor{yellow!15} \textbf{57.0} & \cellcolor{yellow!15} \textbf{62.4} & \cellcolor{yellow!15} \textbf{62.3} & \cellcolor{yellow!15} \textbf{54.9} & \cellcolor{yellow!15} \textbf{53.6} & \cellcolor{yellow!15} \textbf{59.4} \\
\cmidrule(l){2-16}
 & \multirow[t]{6}{*}{Pro} & Vanilla & 54.3 & 55.2 & 45.9 & 51.1 & 53.0 & 54.2 & 45.6 & 45.3 & 52.5 & 53.5 & 44.0 & 43.0 & 49.8 \\
 &  & Summary & 60.0 & 60.2 & 52.3 & 55.7 & 55.3 & 56.4 & 47.6 & 50.5 & 55.9 & 56.3 & 47.9 & 49.3 & 54.0 \\
 &  & RAG & 64.1 & 63.1 & 57.6 & 58.4 & 55.8 & 56.6 & 48.6 & 46.3 & 52.9 & 53.7 & 44.3 & 47.0 & 54.0 \\
 &  & LifeChoice & 64.4 & \textbf{63.8} & \textbf{57.8} & 58.7 & 57.6 & 57.9 & 49.7 & 54.6 & 56.0 & 56.7 & 48.2 & 48.5 & 56.2 \\
 &  & TimeChara & 54.9 & 55.6 & 46.8 & 50.4 & 51.4 & 52.4 & 43.8 & 43.0 & 50.5 & 51.6 & 42.1 & 44.3 & 48.9 \\
 &  & \cellcolor{yellow!15} Arc & \cellcolor{yellow!15} \textbf{64.7} & \cellcolor{yellow!15} 63.6 & \cellcolor{yellow!15} 57.1 & \cellcolor{yellow!15} \textbf{60.5} & \cellcolor{yellow!15} \textbf{64.3} & \cellcolor{yellow!15} \textbf{64.1} & \cellcolor{yellow!15} \textbf{58.8} & \cellcolor{yellow!15} \textbf{57.7} & \cellcolor{yellow!15} \textbf{63.9} & \cellcolor{yellow!15} \textbf{64.0} & \cellcolor{yellow!15} \textbf{56.9} & \cellcolor{yellow!15} \textbf{56.2} & \cellcolor{yellow!15} \textbf{61.0} \\
\midrule
\multirow[t]{12}{*}{Qwen3} & \multirow[t]{6}{*}{8B} & Vanilla & 35.5 & 36.7 & 26.1 & 27.8 & 35.6 & 36.8 & 26.7 & 25.9 & 36.0 & 37.2 & 26.6 & 28.6 & 31.6 \\
 &  & Summary & 43.3 & 42.3 & 33.0 & 33.5 & 39.4 & 39.7 & 29.8 & 32.9 & 36.5 & 36.8 & 26.4 & 28.1 & 35.1 \\
 &  & RAG & \textbf{49.2} & \textbf{48.0} & \textbf{39.8} & 37.7 & 38.5 & 38.7 & 28.4 & 30.3 & 35.6 & 36.1 & 25.8 & 27.7 & 36.3 \\
 &  & LifeChoice & 47.2 & 45.6 & 37.0 & \textbf{42.0} & 39.5 & 39.4 & 29.1 & 30.2 & 37.9 & 38.0 & 27.8 & 29.0 & 36.9 \\
 &  & TimeChara & 38.7 & 40.3 & 29.5 & 31.0 & 38.2 & 39.3 & 29.1 & 29.0 & 37.4 & 38.4 & 27.8 & 27.7 & 33.9 \\
 &  & \cellcolor{yellow!15} Arc & \cellcolor{yellow!15} 46.7 & \cellcolor{yellow!15} 46.6 & \cellcolor{yellow!15} 36.1 & \cellcolor{yellow!15} 37.5 & \cellcolor{yellow!15} \textbf{44.6} & \cellcolor{yellow!15} \textbf{45.2} & \cellcolor{yellow!15} \textbf{33.9} & \cellcolor{yellow!15} \textbf{37.6} & \cellcolor{yellow!15} \textbf{46.2} & \cellcolor{yellow!15} \textbf{46.3} & \cellcolor{yellow!15} \textbf{35.9} & \cellcolor{yellow!15} \textbf{35.3} & \cellcolor{yellow!15} \textbf{41.0} \\
\cmidrule(l){2-16}
 & \multirow[t]{6}{*}{32B} & Vanilla & 40.1 & 40.4 & 30.0 & 33.2 & 40.8 & 41.8 & 31.8 & 30.2 & 41.4 & 42.8 & 31.9 & 31.6 & 36.3 \\
 &  & Summary & 50.5 & 50.0 & 40.3 & 42.7 & 46.4 & 46.9 & 36.8 & 36.5 & 44.3 & 45.0 & 34.3 & 36.0 & 42.5 \\
 &  & RAG & \textbf{55.2} & 53.2 & 45.4 & 41.3 & 46.2 & 47.0 & 36.4 & 36.7 & 42.9 & 42.9 & 32.9 & 33.0 & 42.8 \\
 &  & LifeChoice & 54.8 & \textbf{54.1} & \textbf{46.0} & 45.9 & 46.8 & 47.8 & 37.2 & 35.1 & 43.8 & 44.4 & 34.2 & 37.3 & 44.0 \\
 &  & TimeChara & 42.6 & 42.9 & 32.7 & 34.8 & 41.0 & 42.0 & 31.7 & 29.6 & 42.3 & 43.1 & 32.4 & 32.0 & 37.3 \\
 &  & \cellcolor{yellow!15} Arc & \cellcolor{yellow!15} 55.1 & \cellcolor{yellow!15} 53.5 & \cellcolor{yellow!15} 44.0 & \cellcolor{yellow!15} \textbf{50.7} & \cellcolor{yellow!15} \textbf{55.1} & \cellcolor{yellow!15} \textbf{54.7} & \cellcolor{yellow!15} \textbf{45.6} & \cellcolor{yellow!15} \textbf{44.2} & \cellcolor{yellow!15} \textbf{53.1} & \cellcolor{yellow!15} \textbf{52.6} & \cellcolor{yellow!15} \textbf{43.3} & \cellcolor{yellow!15} \textbf{42.3} & \cellcolor{yellow!15} \textbf{49.5} \\
\midrule
\multirow[t]{12}{*}{\ours{}} & \multirow[t]{6}{*}{8B} & Vanilla & 44.2 & 43.6 & 34.9 & 38.3 & 46.1 & 46.2 & 37.2 & 34.6 & 43.1 & 43.0 & 34.0 & 35.0 & 40.0 \\
 &  & Summary & 52.8 & 51.4 & 44.5 & 46.1 & 45.2 & 45.3 & 36.2 & 40.3 & 44.5 & 44.7 & 35.7 & 36.0 & 43.6 \\
 &  & RAG & \textbf{59.1} & \textbf{57.6} & \textbf{51.9} & 51.2 & 48.5 & 48.6 & 39.7 & 38.5 & 45.4 & 45.5 & 36.8 & 35.1 & 46.5 \\
 &  & LifeChoice & 53.7 & 52.6 & 45.0 & 49.0 & 44.9 & 45.2 & 35.8 & 38.3 & 44.7 & 45.0 & 36.4 & 37.3 & 44.0 \\
 &  & TimeChara & 44.4 & 45.1 & 36.0 & 38.6 & 45.8 & 46.3 & 37.2 & 34.8 & 43.7 & 43.9 & 35.1 & 35.7 & 40.6 \\
 &  & \cellcolor{yellow!15} Arc & \cellcolor{yellow!15} 58.5 & \cellcolor{yellow!15} 57.2 & \cellcolor{yellow!15} 50.4 & \cellcolor{yellow!15} \textbf{51.6} & \cellcolor{yellow!15} \textbf{60.8} & \cellcolor{yellow!15} \textbf{58.8} & \cellcolor{yellow!15} \textbf{52.1} & \cellcolor{yellow!15} \textbf{55.4} & \cellcolor{yellow!15} \textbf{62.0} & \cellcolor{yellow!15} \textbf{59.7} & \cellcolor{yellow!15} \textbf{54.4} & \cellcolor{yellow!15} \textbf{51.6} & \cellcolor{yellow!15} \textbf{56.0} \\
\cmidrule(l){2-16}
 & \multirow[t]{6}{*}{32B} & Vanilla & 48.9 & 49.3 & 39.9 & 41.6 & 55.3 & 55.5 & 47.0 & 41.9 & 55.9 & 55.8 & 47.5 & 42.1 & 48.4 \\
 &  & Summary & 55.3 & 54.3 & 46.3 & 52.8 & 51.4 & 51.0 & 41.8 & 41.8 & 48.5 & 48.8 & 39.2 & 40.9 & 47.7 \\
 &  & RAG & \textbf{61.7} & \textbf{60.0} & \textbf{53.9} & 53.8 & 51.1 & 51.1 & 42.2 & 42.9 & 48.5 & 48.6 & 39.5 & 38.9 & 49.3 \\
 &  & LifeChoice & 57.8 & 55.9 & 49.8 & 53.2 & 53.8 & 53.0 & 44.4 & 45.1 & 50.8 & 50.6 & 42.0 & 45.6 & 50.2 \\
 &  & TimeChara & 47.4 & 47.9 & 38.6 & 41.6 & 55.8 & 55.5 & 47.2 & 44.0 & 56.5 & 56.2 & 49.0 & 41.4 & 48.4 \\
 &  & \cellcolor{yellow!15} Arc & \cellcolor{yellow!15} 60.0 & \cellcolor{yellow!15} 58.6 & \cellcolor{yellow!15} 52.1 & \cellcolor{yellow!15} \textbf{56.5} & \cellcolor{yellow!15} \textbf{72.5} & \cellcolor{yellow!15} \textbf{70.1} & \cellcolor{yellow!15} \textbf{66.8} & \cellcolor{yellow!15} \textbf{66.8} & \cellcolor{yellow!15} \textbf{73.5} & \cellcolor{yellow!15} \textbf{70.7} & \cellcolor{yellow!15} \textbf{68.1} & \cellcolor{yellow!15} \textbf{70.0} & \cellcolor{yellow!15} \textbf{65.5} \\
\bottomrule
\end{tabular}
\caption{Per-(model, mode) results on \emph{The Underdogs} and \emph{East Lynne}. Probe categories and metrics as in Table~\ref{tab:main_results}. Best mode per column per model bolded. Both novels are training novels, so all four \ours{} blocks are in-distribution rather than held out. The four remaining models are unaffected.}
\label{tab:main_results_extra3}
\end{table*}

\section{Additional Models: Full Results\label{app:added-results}}

We report per-(model, mode) results on the human-validated slice for three further model families: HER-32B~\cite{du2026herhumanlikereasoningreinforcement}, CoSER-8B and CoSER-70B~\cite{wang2025coser}, and the SFT-only ablation of our model (\ours{}-SFT-8B and \ours{}-SFT-32B). Compare against Table~\ref{tab:main_results} for the DPO-tuned \ours{} models used in the main comparison.

\begin{table*}[!htbp]
\centering
\scriptsize
\setlength{\tabcolsep}{5pt}
\renewcommand{\arraystretch}{0.85}
\setlength{\aboverulesep}{0.2ex}
\setlength{\belowrulesep}{0.2ex}
\begin{tabular}{lllccccccccccccc}
\toprule
Family & Size & Mode & \multicolumn{4}{c}{In-Scenario} & \multicolumn{4}{c}{In-world} & \multicolumn{4}{c}{Out-of-world} & Overall \\
\cmidrule(lr){4-7} \cmidrule(lr){8-11} \cmidrule(lr){12-15} \cmidrule(lr){16-16}
 & & & APF & RPF & RAE & PTF & APF & RPF & RAE & PTF & APF & RPF & RAE & PTF & Overall \\
\midrule
\multirow[t]{6}{*}{HER} & \multirow[t]{6}{*}{32B} & Vanilla & 51.1 & 49.9 & 42.5 & 40.9 & 48.5 & 47.9 & 39.9 & 31.3 & 45.9 & 46.0 & 37.6 & 30.0 & 42.6 \\
 &  & Summary & 57.9 & 55.7 & 49.6 & 42.4 & 50.1 & 49.1 & 41.5 & 31.3 & 48.1 & 47.4 & 39.2 & 31.2 & 45.3 \\
 &  & RAG & \textbf{63.0} & \textbf{60.2} & \textbf{55.9} & \textbf{47.1} & 49.0 & 48.0 & 40.2 & 32.7 & 48.6 & 47.8 & 39.7 & 32.2 & \textbf{47.0} \\
 &  & LifeChoice & 61.3 & 58.9 & 53.9 & 46.3 & 50.2 & 49.4 & 41.6 & \textbf{34.9} & 47.6 & 47.5 & 38.9 & 31.4 & 46.8 \\
 &  & TimeChara & 52.0 & 50.7 & 44.0 & 35.9 & 48.5 & 48.0 & 40.2 & 31.9 & 46.0 & 46.2 & 37.6 & 30.8 & 42.6 \\
 &  & \cellcolor{yellow!15} Arc & \cellcolor{yellow!15} 56.0 & \cellcolor{yellow!15} 54.6 & \cellcolor{yellow!15} 47.7 & \cellcolor{yellow!15} 44.2 & \cellcolor{yellow!15} \textbf{51.1} & \cellcolor{yellow!15} \textbf{50.4} & \cellcolor{yellow!15} \textbf{42.7} & \cellcolor{yellow!15} 34.6 & \cellcolor{yellow!15} \textbf{49.7} & \cellcolor{yellow!15} \textbf{49.5} & \cellcolor{yellow!15} \textbf{41.3} & \cellcolor{yellow!15} \textbf{35.2} & \cellcolor{yellow!15} 46.4 \\
\midrule
\multirow[t]{12}{*}{CoSER} & \multirow[t]{6}{*}{8B} & Vanilla & 43.0 & 44.2 & 35.1 & 36.6 & 38.8 & 41.6 & 31.5 & \textbf{32.5} & 39.7 & 42.5 & 32.6 & 29.1 & 37.3 \\
 &  & Summary & 45.8 & 46.7 & 38.2 & 37.9 & 38.7 & 41.5 & 31.3 & 32.0 & 38.4 & 41.4 & 30.9 & 30.1 & 37.7 \\
 &  & RAG & \textbf{49.9} & \textbf{50.4} & \textbf{43.5} & 39.4 & 36.7 & 39.8 & 29.3 & 29.9 & 38.2 & 40.9 & 30.8 & 28.9 & 38.1 \\
 &  & LifeChoice & 46.0 & 46.5 & 38.6 & 38.3 & 38.2 & 40.9 & 30.7 & 31.4 & 38.7 & 41.5 & 31.0 & 29.3 & 37.6 \\
 &  & TimeChara & 43.4 & 45.1 & 35.9 & 36.4 & 38.0 & 40.9 & 30.7 & 30.0 & 39.9 & 42.9 & 32.9 & 30.0 & 37.2 \\
 &  & \cellcolor{yellow!15} Arc & \cellcolor{yellow!15} 46.4 & \cellcolor{yellow!15} 47.3 & \cellcolor{yellow!15} 38.7 & \cellcolor{yellow!15} \textbf{39.8} & \cellcolor{yellow!15} \textbf{39.9} & \cellcolor{yellow!15} \textbf{42.6} & \cellcolor{yellow!15} \textbf{32.3} & \cellcolor{yellow!15} 31.1 & \cellcolor{yellow!15} \textbf{40.7} & \cellcolor{yellow!15} \textbf{43.3} & \cellcolor{yellow!15} \textbf{33.3} & \cellcolor{yellow!15} \textbf{31.1} & \cellcolor{yellow!15} \textbf{38.9} \\
\cmidrule(l){2-16}
 & \multirow[t]{6}{*}{70B} & Vanilla & 45.2 & 45.7 & 37.2 & 40.6 & 40.2 & 42.1 & 32.2 & 34.0 & 40.5 & 42.1 & 32.5 & 31.9 & 38.7 \\
 &  & Summary & 48.7 & 48.6 & 40.4 & 44.9 & 40.6 & 42.0 & 32.5 & \textbf{34.7} & 40.5 & 42.0 & 32.2 & 33.7 & 40.1 \\
 &  & RAG & \textbf{59.1} & \textbf{57.8} & \textbf{53.5} & \textbf{47.1} & 38.0 & 39.8 & 29.9 & 32.1 & 40.2 & 41.6 & 32.1 & 32.7 & \textbf{42.0} \\
 &  & LifeChoice & 53.2 & 52.4 & 46.4 & 45.5 & 40.0 & 41.5 & 32.2 & 32.8 & 40.4 & 42.0 & 32.1 & 33.5 & 41.0 \\
 &  & TimeChara & 45.6 & 45.8 & 37.5 & 40.6 & 40.7 & 42.3 & 32.8 & 33.7 & 40.7 & 42.1 & 32.7 & 32.8 & 38.9 \\
 &  & \cellcolor{yellow!15} Arc & \cellcolor{yellow!15} 48.4 & \cellcolor{yellow!15} 48.4 & \cellcolor{yellow!15} 39.8 & \cellcolor{yellow!15} 43.9 & \cellcolor{yellow!15} \textbf{42.7} & \cellcolor{yellow!15} \textbf{43.6} & \cellcolor{yellow!15} \textbf{34.4} & \cellcolor{yellow!15} 34.6 & \cellcolor{yellow!15} \textbf{42.8} & \cellcolor{yellow!15} \textbf{43.9} & \cellcolor{yellow!15} \textbf{34.3} & \cellcolor{yellow!15} \textbf{35.7} & \cellcolor{yellow!15} 41.0 \\
\midrule
\multirow[t]{12}{*}{\ours{}-SFT} & \multirow[t]{6}{*}{8B} & Vanilla & 47.1 & 46.0 & 37.6 & 40.0 & 46.2 & 46.0 & 36.8 & 35.4 & 47.0 & 46.5 & 37.9 & 35.2 & 41.8 \\
 &  & Summary & 53.5 & 52.8 & 44.8 & 47.9 & 49.1 & 49.1 & 40.7 & 39.8 & 49.7 & 49.8 & 41.7 & 38.0 & 46.4 \\
 &  & RAG & 57.2 & 55.7 & \textbf{49.6} & 48.4 & 47.9 & 47.7 & 38.9 & 37.7 & 48.6 & 48.7 & 40.6 & 36.2 & 46.4 \\
 &  & LifeChoice & 54.4 & 53.2 & 46.2 & 47.0 & 49.0 & 48.8 & 40.2 & 38.4 & 48.5 & 48.5 & 40.3 & 37.2 & 46.0 \\
 &  & TimeChara & 47.1 & 46.3 & 37.7 & 40.1 & 46.9 & 46.2 & 37.5 & 35.9 & 46.6 & 46.5 & 38.0 & 36.3 & 42.1 \\
 &  & \cellcolor{yellow!15} Arc & \cellcolor{yellow!15} \textbf{57.3} & \cellcolor{yellow!15} \textbf{55.9} & \cellcolor{yellow!15} 49.2 & \cellcolor{yellow!15} \textbf{51.6} & \cellcolor{yellow!15} \textbf{56.5} & \cellcolor{yellow!15} \textbf{56.0} & \cellcolor{yellow!15} \textbf{48.6} & \cellcolor{yellow!15} \textbf{45.1} & \cellcolor{yellow!15} \textbf{56.7} & \cellcolor{yellow!15} \textbf{55.8} & \cellcolor{yellow!15} \textbf{48.8} & \cellcolor{yellow!15} \textbf{46.7} & \cellcolor{yellow!15} \textbf{52.3} \\
\cmidrule(l){2-16}
 & \multirow[t]{6}{*}{32B} & Vanilla & 56.0 & 55.4 & 47.7 & 48.9 & 54.4 & 54.4 & 47.0 & 40.5 & 55.2 & 55.1 & 48.7 & 41.9 & 50.4 \\
 &  & Summary & 60.5 & 59.1 & 52.7 & 52.8 & 55.3 & 55.2 & 47.7 & 43.4 & 55.6 & 55.3 & 48.4 & 42.9 & 52.4 \\
 &  & RAG & \textbf{65.7} & \textbf{63.7} & \textbf{59.8} & 56.2 & 54.5 & 54.4 & 47.0 & 44.5 & 54.8 & 54.8 & 47.7 & 40.9 & 53.7 \\
 &  & LifeChoice & 62.6 & 61.1 & 55.8 & 55.1 & 55.3 & 55.3 & 48.0 & 44.2 & 55.9 & 55.6 & 48.9 & 42.4 & 53.3 \\
 &  & TimeChara & 55.5 & 54.5 & 47.5 & 48.7 & 54.8 & 54.6 & 47.4 & 41.5 & 54.8 & 54.9 & 48.0 & 40.7 & 50.3 \\
 &  & \cellcolor{yellow!15} Arc & \cellcolor{yellow!15} 62.6 & \cellcolor{yellow!15} 61.3 & \cellcolor{yellow!15} 55.9 & \cellcolor{yellow!15} \textbf{56.7} & \cellcolor{yellow!15} \textbf{61.4} & \cellcolor{yellow!15} \textbf{60.6} & \cellcolor{yellow!15} \textbf{54.7} & \cellcolor{yellow!15} \textbf{51.8} & \cellcolor{yellow!15} \textbf{63.4} & \cellcolor{yellow!15} \textbf{62.3} & \cellcolor{yellow!15} \textbf{57.9} & \cellcolor{yellow!15} \textbf{52.2} & \cellcolor{yellow!15} \textbf{58.4} \\
\bottomrule
\end{tabular}
\caption{Added baselines (HER, CoSER) and the SFT-only \ours{} ablation on the validated slice. Setup matches Table~\ref{tab:main_results}.}
\label{tab:main_results_added}
\end{table*}

\section{Paired Confidence Intervals\label{app:uncertainty}}

Table~\ref{tab:main_results} draws one generation per (probe, mode, model) cell, and observations are nested within arcs, characters, and novels. We therefore resample the $199$ contributing (novel, character, axis) clusters within novel over $10{,}000$ bootstrap replicates, recomputing the equal-novel Overall score on each replicate. Because the main text compares Arc against a baseline selected after inspecting the results, Table~\ref{tab:clustered-ci} reports all $30$ paired contrasts rather than only the selected one. All $30$ intervals exclude zero, and all $30$ leave-one-novel-out differences stay positive. These intervals cover benchmark-sample uncertainty and not decoding variance, which one generation per cell cannot estimate.

\begin{table*}[!t]
\centering
\scriptsize
\setlength{\tabcolsep}{3pt}
\renewcommand{\arraystretch}{1.05}
\resizebox{\textwidth}{!}{%
\begin{tabular}{lcccccc}
\toprule
Model & vs.\ Vanilla & vs.\ Summary & vs.\ RAG & vs.\ LifeChoice & vs.\ TimeCHARA & Min.\ LOO \\
\midrule
DeepSeek-V4-Flash & $+6.87$ [$6.06$, $7.70$] & $+3.95$ [$2.97$, $4.92$] & $+4.85$ [$4.04$, $5.60$] & $+3.60$ [$2.79$, $4.38$] & $+6.61$ [$5.80$, $7.40$] & $+3.41$ \\
DeepSeek-V4-Pro   & $+10.02$ [$9.20$, $10.84$] & $+5.20$ [$4.34$, $6.07$] & $+5.69$ [$4.77$, $6.61$] & $+4.66$ [$3.80$, $5.51$] & $+10.25$ [$9.45$, $11.06$] & $+4.34$ \\
Qwen3-8B          & $+5.90$ [$5.01$, $6.77$] & $+4.41$ [$3.58$, $5.26$] & $+2.18$ [$1.36$, $3.04$] & $+3.24$ [$2.44$, $4.10$] & $+5.65$ [$4.85$, $6.50$] & $+1.79$ \\
Qwen3-32B         & $+6.43$ [$5.68$, $7.19$] & $+4.54$ [$3.76$, $5.37$] & $+2.89$ [$2.01$, $3.78$] & $+2.71$ [$1.93$, $3.50$] & $+5.69$ [$4.90$, $6.48$] & $+2.33$ \\
\ours{}-8B-DPO    & $+13.91$ [$12.64$, $15.16$] & $+9.69$ [$8.40$, $11.07$] & $+8.41$ [$7.15$, $9.71$] & $+9.26$ [$8.13$, $10.46$] & $+13.21$ [$12.07$, $14.36$] & $+7.96$ \\
\ours{}-32B-DPO   & $+10.52$ [$9.05$, $12.01$] & $+11.10$ [$9.65$, $12.60$] & $+8.41$ [$6.93$, $9.89$] & $+8.35$ [$6.91$, $9.86$] & $+10.44$ [$8.96$, $11.95$] & $+7.30$ \\
\bottomrule
\end{tabular}}
\caption{Paired Arc-versus-baseline differences in Overall on the validated slice, with arc-clustered $95\%$ bootstrap intervals. Min.\ LOO is the smallest difference against the reselected strongest baseline after omitting each novel in turn.}
\label{tab:clustered-ci}
\end{table*}

\section{Source-of-Effect Ablation: Setup and Extended Reading\label{app:arc-ablation-extra}}

This appendix documents the ablations summarized in \S\ref{sec:arc-ablation}.

\paragraph{\textsc{MixedArc} construction.}
For each probe, the donor character is chosen deterministically from the same novel by hashing the probe id and indexing into the sorted list of characters with a constructed arc, excluding the queried character. The donor's full arc JSON is loaded (axis blocks, per-phase trajectory, pole descriptions), then the top-level \texttt{character} field is rewritten to the queried character's name so the swap is not announced in the prompt. The arc is truncated at the queried chapter under the same rule used for the unaltered Arc context (\S\ref{sec:arc}), so the surface length and structure match Arc closely.

\paragraph{\textsc{ArcHint} construction.}
The arc body is stripped to one line per (intrapersonal or relational) axis of the form \texttt{Axis: <axis\_name> / Phase: $k$ of $N$ (label: <phase\_label>)}, where $k$ is determined by the queried chapter and the phase-segmented trajectory. No per-phase prose, pole descriptions, or evidence summaries are exposed. Typical \textsc{ArcHint} contexts are $\sim$300 characters per probe, against $\sim$10K characters for the full Arc JSON.

\paragraph{Probe sample.}
We draw $\sim 120$ probes per model from the validated slice: $15$ characters (three per novel), $8$ probes per character, stratified across the three probe categories (In-Scenario, In-World, Out-of-World). Paired deltas in Figure~\ref{fig:leakage_bars} are computed within model on whichever probes have all four mode rows for that model. The Vanilla and Arc rows reuse the existing main-results responses for the corresponding probes. All four modes are scored by the same Stage~iii judge (\S\ref{sec:evalproto}).

\paragraph{Three-layer reading.}
The two ablations partition the pipeline's contribution into three layers. (i) Construction (\S\ref{sec:arc}) extracts a (character, axis-set, phase-trajectory) structure from the chapter text. (ii) Inference: on the untrained models (DeepSeek-V4-Flash and Qwen3-32B) the highest-level part of that structure (axis labels plus the current phase index) already carries the full Arc benefit, so the per-phase prose is redundant for prompting these models. (iii) Training: the per-phase prose is the supervision substrate \ours{}-DPO conditions on, so \textsc{ArcHint} recovers only roughly half of the Arc lift on \ours{}-32B-DPO (Figure~\ref{fig:leakage_bars}), localizing the $+10.3$ lift over Qwen3-32B in \S\ref{sec:main} to the prose conditioning. A practical implication is that \textsc{ArcHint}, at $\sim 40\times$ smaller context than the full arc, is a candidate inference variant for the untrained-model regime: long-context-constrained or multi-character deployments built on a general-purpose model.

\section{Memorization Robustness\label{app:memorization}}

A first concern with any benchmark over popular novels is that the headline gain reflects pretraining memorization of the source text rather than the targeted construct. We check this two ways. \emph{Within subset gradient.} The Arc-vs-best-baseline gap on the validated slice is smallest on \emph{Harry Potter} ($+3.0$), the only non-Gutenberg title and the one with the largest non-corpus footprint (fan wikis, online discussion, derivative fiction), and largest on \emph{Anna Karenina} ($+5.6$) and \emph{Don Quixote} ($+5.1$). Among the four public-domain titles, the gap ordering roughly reverses Project Gutenberg download count, and the same ordering reappears in the \ours{}-vs-Qwen3 lift under Arc. A pure memorization account would predict shrinking-to-zero or negative gaps at the high-popularity end. Arc still wins $29$ of $30$ (model, novel) cells on the validated slice and all $12$ (model, novel) cells on the two low-popularity novels that are disjoint from both training pools (\S\ref{sec:additional}).

\section{Central vs.\ Supporting Characters\label{app:character-class}}

\begin{figure}[!t]
\centering
\includegraphics[width=\linewidth]{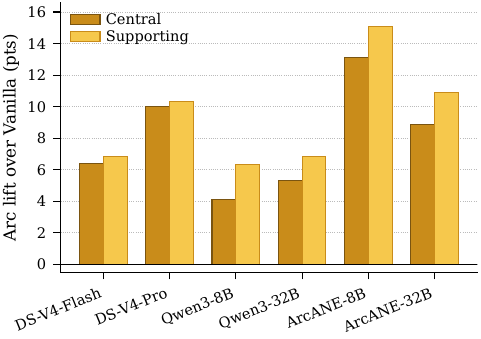}
\caption{Arc-over-Vanilla Overall lift, central ($N{=}6$) vs.\ supporting ($N{=}19$) characters.}
\label{fig:per-char-class}
\end{figure}

A second variant of the memorization concern is that the Arc lift concentrates on a handful of protagonists the base models already know well. We split the $25$ principal characters in the validated slice into a \emph{central} tier (each novel's title character, plus Levin as \emph{Anna Karenina}'s co-protagonist; $N_{\mathrm{c}}{=}6$) and a \emph{supporting} tier ($N_{\mathrm{s}}{=}19$), and report the Arc-over-Vanilla Overall lift on each (Figure~\ref{fig:per-char-class}). Vanilla supplies no external evidence, so the gap measures the marginal value of arc grounding above what the model already knows.

Every model lifts both tiers (central $+4.1$ to $+13.1$, supporting $+6.3$ to $+15.1$), and the supporting-tier lift exceeds the central-tier lift uniformly across the six models (by $+0.3$ to $+2.2$ points). A pure memorization account predicts the reverse, since central characters dominate pretraining corpora. \ours{}-8B records the largest gain in either tier ($+13.1$, $+15.1$): training on \ours{} concentrates the lift where memorization explains least. The character-popularity axis mirrors the source-distance axis of \S\ref{sec:main}: Arc helps where the model needs help.

\section{PTF Metric Validity\label{app:ptf-validity}}

PTF aggregates a single judge call across $N$ phase-keyed (reference, response) pairs (\S\ref{sec:evalproto}), so it is exposed to two failure modes a per-phase score is not: it might be a redundant re-summary of per-phase fidelity, and it might be driven by surface listing order rather than trajectory content. We test both. The main-text version of these checks is in \S\ref{sec:judge-agreement}, and this appendix gives the full breakdown.

\paragraph{Convergent-but-not-redundant correlation.} For every (model, mode) cell on the validated slice we compute Pearson and Spearman correlations between PTF and the per-phase average $\frac{1}{3}(\text{APF}+\text{RPF}+\text{RAE})$. Across $72$ cells (six base models $\times$ six modes, plus four \ours{} models minus two cells without trajectory coverage, with $n\ge 745$ probes per cell), Pearson is $0.51$ on average (median $0.52$, range $0.37$ to $0.64$), and Spearman is $0.48$ on average. The implied $r^2$ is $0.26$, so per-phase scoring explains roughly a quarter of PTF variance: PTF is convergent with per-phase fidelity but not a re-summary of it. The cells with the highest correlation are exactly the cells where the model is doing the most phase tracking (\ours{}-32B-DPO under Arc tops at $r{=}0.64$), consistent with PTF and per-phase agreeing about \emph{whether} the model anchors each phase while disagreeing about \emph{which} of the two views matters at the trajectory level.

\paragraph{Targeted perturbations.} We re-call the trajectory judge on perturbed copies of each (probe, model, mode) trajectory under four conditions:
\begin{itemize}
\setlength{\itemsep}{1pt}
\item \emph{orig}: test-retest noise.
\item \emph{shuffle\_response}: model responses permuted across phase blocks, with references fixed.
\item \emph{reverse\_response}: model responses reversed, with references fixed.
\item \emph{block\_shuffle}: whole (reference, response) blocks shuffled together so each pair stays intact but listing order changes.
\end{itemize}
$75$ probes per model are stratified across the five validated novels, $N{\geq}3$ phases. Results are in Table~\ref{tab:ptf-validity}.

\begin{table}[!t]
\centering
\small
\setlength{\tabcolsep}{4pt}
\resizebox{\columnwidth}{!}{%
\begin{tabular}{l rrr rrr}
\toprule
 & \multicolumn{3}{c}{\ours{}-32B-DPO} & \multicolumn{3}{c}{DeepSeek-V4-Pro} \\
\cmidrule(lr){2-4} \cmidrule(lr){5-7}
Condition & Align & Dir & Avg & Align & Dir & Avg \\
\midrule
orig                  & 56.0 & 54.8 & 53.5  & 57.9 & 55.3 & 54.5 \\
shuffle\_response     & 47.7 & 45.3 & 44.7  & 56.2 & 54.4 & 53.2 \\
reverse\_response     & 34.1 & 30.0 & 30.5  & 52.2 & 48.3 & 48.2 \\
block\_shuffle        & 43.3 & 42.7 & 41.0  & 53.7 & 50.2 & 49.8 \\
\midrule
$\Delta$ shuffle      & $-8.3$ & $-9.4$ & $-8.8$  & $-1.7$ & $-0.9$ & $-1.3$ \\
$\Delta$ reverse      & $-21.9$ & $-24.7$ & $-23.0$  & $-5.7$ & $-7.0$ & $-6.3$ \\
$\Delta$ block-shuffle& $-12.8$ & $-12.1$ & $-12.5$  & $-4.2$ & $-5.1$ & $-4.7$ \\
\bottomrule
\end{tabular}}
\caption{PTF under perturbation, $N{=}75$ probes per model under Arc context. Paired deltas vs.\ \emph{orig}.}
\label{tab:ptf-validity}
\end{table}

Three reads. (i) PTF is not invariant: shuffling responses costs $-8.8$ Overall on \ours{}-32B-DPO and reversing them costs $-23.0$, with the reversal hit concentrated on \texttt{ptf\_direction} ($-24.7$) as the rubric predicts. (ii) The perturbation drop is itself a behavioral signature of phase tracking: on DeepSeek-V4-Pro, a strong baseline whose Arc-context responses are less phase-anchored to begin with, the same shuffles cost only $-1.3$ and $-6.3$, because there is less per-phase structure for shuffling to destroy. (iii) The \emph{block\_shuffle} row does not isolate block position, and the correction is given below.

\paragraph{Order control for block-shuffle.}
The \emph{block\_shuffle} prompt retained each \texttt{[PHASE k]} identifier while still describing the displayed blocks as chronological, so the shuffle put physical position and the stated instruction in conflict. We re-ran the condition on the same probes under a matched order-agnostic instruction that directs the judge to reconstruct chronology from the retained identifiers, covering $150$ trajectories. Under the original prompt the plain shuffle reproduces Table~\ref{tab:ptf-validity} ($-12.76$ [$-16.31$, $-8.74$] on \ours{}-32B-DPO and $-6.76$ [$-10.10$, $-3.21$] on DeepSeek-V4-Pro, arc-clustered $95\%$ intervals containing the Table~\ref{tab:ptf-validity} point values). Swapping only the instruction, with no shuffle, moves PTF by $-0.80$ [$-2.40$, $+0.92$] and $-1.27$ [$-4.25$, $+1.96$], intervals covering zero. Once the instruction is matched, the shuffle effect disappears and turns slightly positive ($+3.71$ [$+1.23$, $+5.85$] and $+6.78$ [$+3.80$, $+9.88$]). The original degradation therefore traces to the contradictory instruction rather than to order-invariant trajectory sensitivity. We bound the small positive term rather than build on it: its sign is unexpected, we have no mechanism for it, and it arises only under a counterfactual prompt used for no reported score, since every scored trajectory in this paper is rendered from \texttt{phase\_idx}-sorted blocks and the shuffled regime never reaches the scorer. We therefore withdraw the block-shuffle claim about signal beyond positional reading and retain the response-shuffle and response-reverse reads, which corrupt trajectory content rather than listing order.

\section{Judge Validation\label{app:judge-crossval}}

This appendix documents the three checks summarized in \S\ref{sec:judge-agreement}: a $70$-sample human-annotation study with three annotators (Appendix~\ref{app:judge-crossval-human}) that produces both plausibility verdicts and re-scored APF/RPF/RAE values, a blind reference study (Appendix~\ref{app:judge-crossval-blind}), and a $300$-cell cross-judge replication on the validated slice (Appendix~\ref{app:judge-crossval-cross}) that triangulates DeepSeek-V4-Flash (DS-V4-Flash) against three independent strong judges from different model families. The first and third check the judge; only the blind study checks the references, which the judge and the Arc context both inherit from the same pipeline.

\subsection{Human-anchored validation\label{app:judge-crossval-human}}

\paragraph{Sample.} Three annotators each scored $70$ probes drawn from across all $5$ validated novels and $2$ further corpus novels available at the time of the study (Anna~Karenina, Benjamin~Franklin, Don~Quixote, Harry~Potter, Monte~Cristo, East~Lynne, Hung~Lou~Meng), covering both eval models we focus on (Qwen3-32B and \ours{}-32B-DPO) under all six context modes (Vanilla, Summary, RAG, LifeChoice, TimeCHARA, Arc). Of the $70$ samples, $50$ are eval\_1 cells (per-phase APF/RPF/RAE judge calls) and $20$ are eval\_2 cells (trajectory PTF judge calls). Each annotator returned (i) a binary \texttt{is\_reasonable} verdict on the DS-V4-Flash judge decision and (ii) an optional per-dimension re-score on the same $1$--$100$ scale.

\paragraph{Plausibility.} Of the $210$ binary verdicts, $172$ ($81.9\%$) rate the judge decision as plausible. Aggregated to a sample-level $2$-of-$3$ majority, $61 / 70$ samples ($87.1\%$) pass with $0$ ties. Plausibility is broadly uniform across categories: In-Scenario $84.6\%$, In-World $82.3\%$, Out-of-World $95.7\%$ (the latter actually higher) -- all within $10$ points of the $87.1\%$ overall figure reported in the main text. The eval\_2 trajectory judge passes at $95.0\%$, slightly above the per-phase rubric ($84.0\%$). The remaining $9$ of $70$ samples fail the two-of-three majority. On the binary decision, raw agreement is $72.4\%$, Fleiss' $\kappa$ is $0.068$, Krippendorff's $\alpha$ is $0.073$, and Gwet's AC1 is $0.608$. As with axis validation, the spread follows from the $81.9\%$ positive prevalence, under which chance-corrected coefficients collapse while AC1 stays robust.

\begin{figure}[!t]
    \centering
    \includegraphics[width=\columnwidth]{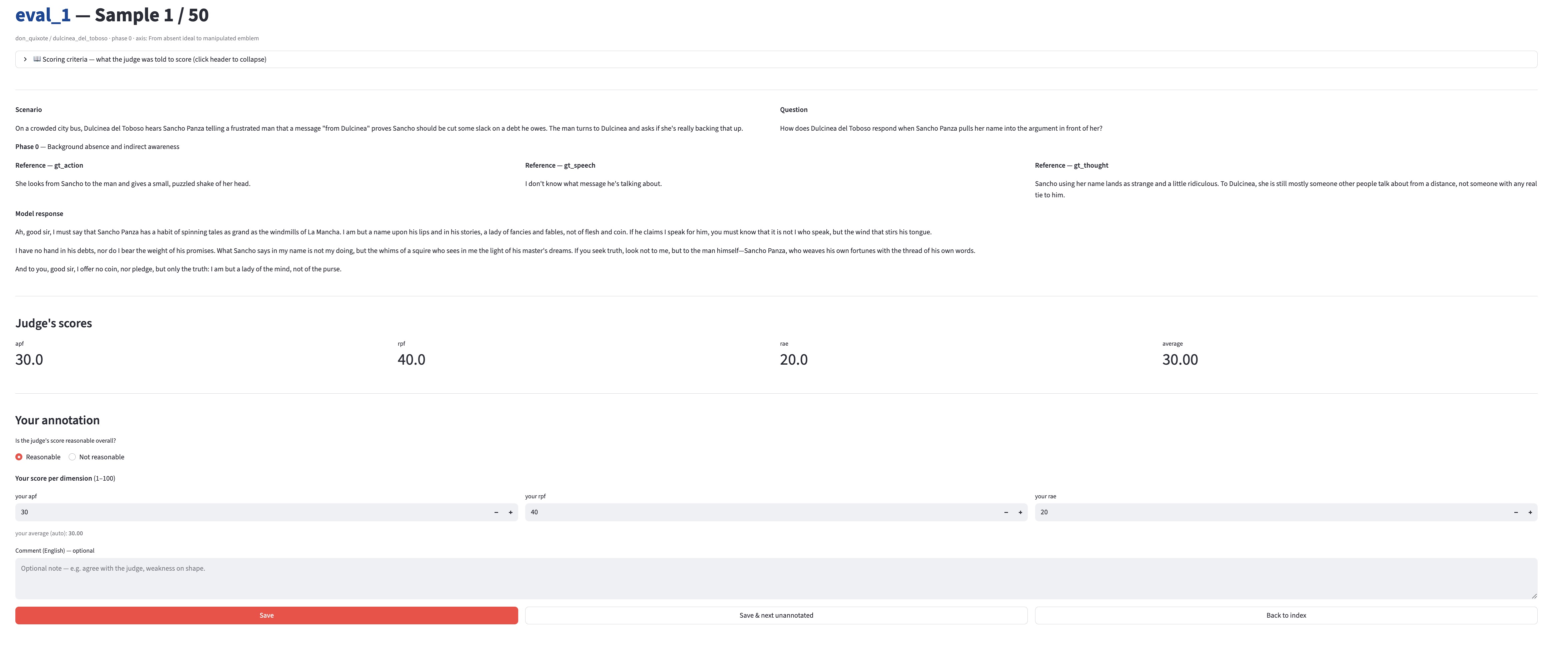}
    \caption{Annotation Page for LLM judges}
    \label{fig:annotation_2}
\end{figure}

\paragraph{Judge agreement with the human re-score reference.} On the $50$ eval\_1 cells where annotators also re-scored APF/RPF/RAE, the human re-scoring was anchored: annotators saw the DS-V4-Flash judge's verdict before adjusting, so this is a calibration check (``where humans corrected the judge, by how much?''), not an independent ground truth. Table~\ref{tab:judge-vs-human-ds} reports DS-V4-Flash's per-dimension agreement with the human-average re-score. The judge matches the human at Pearson $r{\geq}0.92$ on every dimension, with mean absolute adjustment under $7.1$ points on $0$--$100$. The largest absolute bias is on APF ($\Delta{=}+6.1$), with RPF and RAE running within $3$ points of the human average. RAE, the entailment dimension the rubric weights most heavily, shows $r{=}0.97$, $\alpha{=}0.97$, and a $\Delta$ of only $+0.2$. The judge-score verification annotation did not constitute human subjects research under our institution's IRB definitions. The task asked annotators to review scores produced by the LLM judge and rate whether each judgment was plausible. No personally identifiable information, sensitive data, or behavioral data about the annotators themselves was collected. Annotators were colleagues who contributed on a non-compensated basis with informed consent. (see Figure~\ref{fig:annotation_2})

\begin{table}[t!]
\centering
\small
\setlength{\tabcolsep}{4pt}
\resizebox{\columnwidth}{!}{%
\begin{tabular}{lrrrrr}
\toprule
Dim & Pearson & Spearman & $\alpha$-int & MAD & $\Delta_{J{-}H}$ \\
\midrule
APF & 0.920 & 0.915 & 0.877 & 7.1 & $+6.1$ \\
RPF & 0.961 & 0.957 & 0.944 & 4.6 & $+3.0$ \\
RAE & 0.967 & 0.962 & 0.965 & 4.2 & $+0.2$ \\
\midrule
\textbf{Overall} & \textbf{0.962} & \textbf{0.958} & \textbf{0.947} & \textbf{4.7} & $\mathbf{+3.1}$ \\
\bottomrule
\end{tabular}}
\caption{DS-V4-Flash judge vs.\ human re-score reference ($n{=}50$ eval\_1 cells, human average over $3$). The re-score is anchored to the DS-V4-Flash verdict, so this measures how much humans \emph{adjust} the judge, not whether the judge beats a blind human baseline. Differences are small ($\leq 7$ MAD) and the rank ordering is preserved (Spearman~$\geq 0.92$).}
\label{tab:judge-vs-human-ds}
\end{table}

\subsection{Blind reference validation\label{app:judge-crossval-blind}}

The previous study was anchored because annotators saw the DS-V4-Flash verdict before scoring, meaning it evaluated the judge rather than the reference answers. Furthermore, many of our references are AI-generated projections rather than direct quotes from the source text. To ensure these generated references are objectively valid, we conducted a blind study on one character, Ron Weasley (\emph{Harry Potter}), using Qwen3-32B with $15$ items per task. Annotators were three readers familiar with the novel who had completed both tasks.

\paragraph{Protocol.} Both tasks draw a stratified sample of $15$ items. In the reference task, annotators see the scenario, the question, and the stated narrative point, and return a ternary verdict (defensible / not defensible / uncertain). In the pairwise task they compare the Arc output with a non-Arc output of the same model, presented in randomized order with mode labels stripped and ties allowed. Neither task exposes the generating arc, the context mode, or the judge verdict. Items are stratified by probe type, by grounding (source-backed anchor against LLM-projected), and by Q-Discrim status, and the item-level majority is the unit of analysis.

\paragraph{Results.} Every reference is defensible by majority ($15/15$), and the strata least insulated from construction match the source-grounded controls exactly: LLM-projected $12/12$ against grounded $3/3$, and $6/6$, $4/4$, $5/5$ across In-Scenario, In-World, and Out-of-World, with $9/9$ separated and $6/6$ weak-pair by Q-Discrim status. The pairwise task is directional rather than conclusive: Arc wins $7$ of the $11$ items that reach a majority ($63.6\%$, $95\%$ Wilson CI $[35\%,85\%]$), with $4$ items split. The advantage sits on Out-of-World probes, where Arc takes all $5$ decisive items, and is absent on In-Scenario and In-World; by Q-Discrim status Arc wins $3/6$ separated and $4/5$ weak-pair decisive items.

\paragraph{Agreement.} Observed agreement is $0.64$ on the reference task and $0.44$ on the pairwise task. As with axis validation, the high prevalence of ``defensible'' drives chance agreement to $P_e{=}0.70$, so Fleiss' $\kappa$ is deflated to $-0.19$ while prevalence-robust coefficients stay positive (Gwet AC1 $+0.58$, Brennan--Prediger $+0.47$). On the pairwise task the three-way verdict gives $\kappa{=}+0.10$, AC1 $+0.20$, and BP $+0.17$, with ties the main source of disagreement.

\paragraph{Scope.} The study covers a single character, and we report it at that scale. It supports the reference-defensibility claim, which the anchored judge study of Appendix~\ref{app:judge-crossval-human} cannot address, and it does not establish the pairwise preference at this sample size. For $12$ of the $15$ pairwise items the reference is itself an arc-derived projection, so on those items the comparison measures agreement with the \ours{} phase representation rather than source authenticity. Scaling the study across multiple novels and characters is the main outstanding validation step.

\subsection{Cross-judge replication on the validated slice\label{app:judge-crossval-cross}}

\paragraph{Sample.} The human-annotated pool is $70$ cells. To stress-test cross-judge agreement on a broader sample, we run the same three cross-judges on a stratified $300$-cell sample drawn from the validated $5$-novel slice. The stratification is $5$ novels~$\times$~$4$ (model, mode) configs~$\times$~$15$ cells. The $4$ configs are chosen to span the DS-V4-Flash score range: (Qwen3-32B, Vanilla) for the low-score regime, (Qwen3-32B, Arc) and (DeepSeek-V4-Flash, Arc) for the mid range, and (\ours{}-32B-DPO, Arc) for the high range. Within each (novel, config) bucket we round-robin across the three probe categories to keep category coverage balanced ($100$ cells per category). The DS-V4-Flash score on the resulting sample ranges from $15$ to $96$ (mean $55.6$), covering the full benchmark distribution.

\paragraph{Cross-judge agreement.} Table~\ref{tab:judge-crossval-overall} reports per-dimension cell-level agreement between DS-V4-Flash and each cross-judge on the $300$-cell sample. Opus 4.5 and GPT-5.5 reach Krippendorff $\alpha{\geq}0.80$ on the per-cell average, the standard threshold for substantial agreement. Sonnet 4.5 is slightly lower at $\alpha{=}0.60$ but Spearman remains $0.78$, so it preserves the rank ordering even as it differs in absolute level. The pairwise correlation between the two strongest non-DS judges, Opus 4.5 and GPT-5.5, is the highest in the matrix ($r{=}0.92$, $\alpha{=}0.89$), and DS-V4-Flash correlates with each at $r{\geq}0.81$.

\begin{table*}[t!]
\centering
\small
\setlength{\tabcolsep}{5pt}
\renewcommand{\arraystretch}{0.95}
\begin{tabular}{llrrrrrrr}
\toprule
Judge & Dim & Pearson & Spearman & $\alpha$-int & MAD & DS mean & J mean & $\Delta$ \\
\midrule
\multirow{4}{*}{Claude Sonnet 4.5} & APF & 0.726 & 0.776 & 0.543 & 18.5 & 58.7 & 42.3 & $+16.4$ \\
                                   & RPF & 0.716 & 0.730 & 0.550 & 17.0 & 57.9 & 43.9 & $+14.0$ \\
                                   & RAE & 0.720 & 0.755 & 0.645 & 15.8 & 50.2 & 39.7 & $+10.6$ \\
                                   & \textbf{avg} & \textbf{0.739} & \textbf{0.777} & \textbf{0.600} & \textbf{16.5} & 55.6 & 42.0 & $\mathbf{+13.7}$ \\
\midrule
\multirow{4}{*}{Claude Opus 4.5}   & APF & 0.831 & 0.820 & 0.818 & 10.7 & 58.7 & 55.5 & $+3.2$ \\
                                   & RPF & 0.793 & 0.785 & 0.735 & 12.6 & 57.9 & 50.0 & $+7.9$ \\
                                   & RAE & 0.803 & 0.811 & 0.792 & 12.0 & 50.2 & 48.1 & $+2.1$ \\
                                   & \textbf{avg} & \textbf{0.825} & \textbf{0.823} & \textbf{0.806} & \textbf{10.9} & 55.6 & 51.2 & $\mathbf{+4.4}$ \\
\midrule
\multirow{4}{*}{GPT-5.5}           & APF & 0.808 & 0.798 & 0.807 & 11.5 & 58.7 & 59.2 & $-0.5$ \\
                                   & RPF & 0.775 & 0.767 & 0.770 & 12.0 & 57.9 & 56.7 & $+1.3$ \\
                                   & RAE & 0.793 & 0.800 & 0.785 & 13.4 & 50.2 & 54.0 & $-3.7$ \\
                                   & \textbf{avg} & \textbf{0.811} & \textbf{0.808} & \textbf{0.811} & \textbf{11.5} & 55.6 & 56.6 & $\mathbf{-1.0}$ \\
\bottomrule
\end{tabular}
\caption{Cross-judge agreement with DS-V4-Flash on the $300$-cell validated subset sample. $\Delta = $ DS mean $-$ J mean (positive means DS is more lenient).}
\label{tab:judge-crossval-overall}
\end{table*}

\paragraph{System ranking.} The most consequential check is whether the four-config ranking induced by DS-V4-Flash survives a judge swap. Table~\ref{tab:judge-crossval-rank} reports the mean per-cell Overall under each judge. All four judges agree on the top (\ours{}-32B-DPO under Arc) and the bottom (Qwen3-32B under Vanilla). The second and third positions reverse between DS-V4-Flash and Opus (which give DS-V4-Flash/Arc the silver) versus GPT-5.5 and Sonnet (which rank Qwen3-32B/Arc and DS-V4-Flash/Arc within $1.1$ points). The reversal is on tightly scored competitors and does not affect the headline finding of Table~\ref{tab:main_results}.

\begin{table*}[t!]
\centering
\small
\setlength{\tabcolsep}{6pt}
\renewcommand{\arraystretch}{0.95}
\resizebox{\textwidth}{!}{
\begin{tabular}{rllll}
\toprule
Rank & DS-V4-Flash & Opus 4.5 & GPT-5.5 & Sonnet 4.5 \\
\midrule
1 & \ours{}-32B-DPO/Arc (66.0) & \ours{}-32B-DPO/Arc (62.2) & \ours{}-32B-DPO/Arc (69.9) & \ours{}-32B-DPO/Arc (57.5) \\
2 & DS-V4-Flash/Arc (58.3)     & DS-V4-Flash/Arc (51.9)     & Qwen3-32B/Arc (55.3)       & Qwen3-32B/Arc (39.5) \\
3 & Qwen3-32B/Arc (53.7)       & Qwen3-32B/Arc (48.7)       & DS-V4-Flash/Arc (54.2)     & DS-V4-Flash/Arc (39.0) \\
4 & Qwen3-32B/Vanilla (44.5)   & Qwen3-32B/Vanilla (41.9)   & Qwen3-32B/Vanilla (47.1)   & Qwen3-32B/Vanilla (31.8) \\
\bottomrule
\end{tabular}
}
\caption{Configuration ranking by each judge on the $300$-cell sample. Mean per-cell Overall in parentheses. Ranks~$1$ and~$4$ are unanimous across all four judges.}
\label{tab:judge-crossval-rank}
\end{table*}

\paragraph{Reading.} Two complementary anchors converge. (i)~On the $70$-sample human study, the DS-V4-Flash judge passes a $2$-of-$3$ plausibility majority on $87\%$ of probes, uniform across categories. On the $50$ cells with human re-scoring, the human and judge averages move together at Pearson $r{=}0.96$ with MAD~$4.7$, so where annotators adjusted the judge, they adjusted by only a few points. The re-score is anchored to the DS-V4-Flash verdict, so this is a calibration check rather than an independent ground truth, but it shows the judge is not far from where humans land. (ii)~On the $300$-cell sample with no human anchor, DS-V4-Flash sits inside the three-judge consensus ($r{\geq}0.81$, $\alpha{\geq}0.80$ against each of Opus and GPT-5.5), and all four judges return the same top-ranked configuration. Together, these ablations address the two most credible reviewer concerns about an LLM-as-judge benchmark, namely that the judge is unreliable in absolute terms and that the judge is self-preferring in its rankings, with quantitative evidence rather than appeal to authority.

\section{Prompts\label{app:prompts}}

We list the verbatim prompts used throughout the pipeline. Placeholders in angle brackets (e.g.\ \texttt{<character>}) are substituted at runtime. Each table shows the \texttt{[System]} message followed by the \texttt{[User]} template that the API call uses, and rendered output JSON schemas are kept in full so any field that downstream code reads is visible. Where multiple stages share a sub-block we factor it out (Tables~\ref{tab:prompt-style} and~\ref{tab:prompt-phase-instructions}) and reference it from each caller.

\clearpage
\subsection{Arc construction}
\begin{table*}[htbp]
\centering
\begin{lstlisting}[basicstyle=\ttfamily\scriptsize]
[System]
You are a literary analyst specializing in character psychology. You are analyzing <NOVEL> chapter by chapter to track psychologically impactful events for a specific character.

[User]
TARGET CHARACTER: <character>

RUNNING CONTEXT (events extracted from previous chapters):
<running_context>

CURRENT CHAPTER (<chapter_num>):
<chapter_text>

Extract events from this chapter that have psychological impact on <character>. Include events where:
- <character>'s emotions, beliefs, or self-perception shift
- <character>'s relationship with another character changes
- <character> makes a significant decision or judgment
- Something happens that will later affect <character>'s development

If <character> does not appear in this chapter or nothing psychologically relevant happens to them, return an empty events array.

IMPORTANT: Be selective. Only include events with genuine psychological weight, not every minor interaction.

Return JSON only, no other text:
{
  "chapter": <chapter_num>,
  "character": "<character>",
  "events": [
    {
      "event_id": "ch<chapter_num>_evt01",
      "description": "Brief description of what happens",
      "characters_involved": ["list", "of", "characters"],
      "psychological_impact": {
        "emotional_state": "What <character> feels",
        "belief_change": "How their beliefs/views shift (or 'none')",
        "self_perception_change": "How they see themselves differently (or 'none')",
        "sequence_type": "redemptive (bad->good outcome) / contaminating (good->bad outcome) / neutral",
        "meaning_made": "Does <character> extract a lesson or update their self-narrative from this event? Describe briefly, or 'none'"
      },
      "relationship_changes": [
        {
          "target": "OtherCharacterName",
          "direction": "negative/positive/neutral/ambivalent",
          "change_description": "How the relationship changes",
          "dimensions_affected": ["trust", "esteem", "intimacy",
                                  "antagonism"]
        }
      ],
      "intensity": "low/medium/high"
    }
  ],
  "updated_running_summary": "Updated 2-3 sentence summary of <character>'s psychological trajectory so far, incorporating this chapter's events"
}
\end{lstlisting}
\caption{Stage~i event-stream prompt.}
\label{tab:prompt-event-stream}
\end{table*}

\clearpage

\begin{table*}[htbp]
\centering
\begin{lstlisting}[basicstyle=\ttfamily\scriptsize]
[System]
You are a literary psychologist analyzing character development in <NOVEL>. You track psychological states, belief shifts, desires, intentions, and relationship dynamics chapter by chapter.

[User]
TARGET CHARACTER: <character>

RUNNING PSYCHOLOGICAL PROFILE (accumulated from previous chapters):
<running_profile>

CURRENT CHAPTER (<chapter_num>):
<chapter_text>

Analyze this chapter for <character>'s psychological state and any changes. Describe:

1. PRESENCE: Is <character> present or mentioned in this chapter? If not, return minimal response.
2. EMOTIONAL STATE: What is <character> feeling in this chapter? How does it compare to the previous chapter?
3. BELIEFS & VALUES: Any shifts in what <character> believes about themselves, others, society, morality?
4. DESIRES: What does <character> want to be true about the world or their situation? How have their desires shifted from the previous chapter?
5. INTENTIONS: What is <character> actively trying to bring about? What plans or goals are they pursuing?
6. KEY RELATIONSHIPS: How does <character> view/feel about other characters in this chapter? Any shifts?
7. SELF-AWARENESS: Does <character> show any new self-knowledge or remain unaware of something about themselves?
8. CHANGE MAGNITUDE: Rate overall psychological change in this chapter as: none / subtle / moderate / significant / transformative

Return JSON only:
{
  "chapter": <chapter_num>,
  "character": "<character>",
  "present_in_chapter": true,
  "emotional_state": "Description",
  "belief_shifts": "Description or 'none'",
  "current_desires": "What <character> wants to be true about the world or their situation, or 'none'",
  "current_intentions": "What <character> is actively trying to bring about, or 'none'",
  "relationship_states": [
    {
      "target": "OtherCharacterName",
      "current_attitude": "Description of current attitude/feeling",
      "direction_of_change": "warming/cooling/stable/conflicted",
      "dimensions": {
        "trust": "low/medium/high",
        "esteem": "low/medium/high",
        "intimacy": "low/medium/high"
      }
    }
  ],
  "self_awareness_notes": "Description or 'none'",
  "change_magnitude": "none/subtle/moderate/significant/transformative",
  "chapter_summary": "1-2 sentence summary of psychological state at end of this chapter",
  "updated_running_profile": "Updated comprehensive psychological profile of <character> as of this chapter, max 300 words. This should capture their current emotional state, active beliefs, desires, intentions, relationship stances, and ongoing internal conflicts."
}
\end{lstlisting}
\caption{Stage~i state-stream prompt.}
\label{tab:prompt-state-stream}
\end{table*}

\begin{table*}[htbp]
\centering
\begin{lstlisting}[basicstyle=\ttfamily\scriptsize]
[System]
You are a literary analyst specializing in character psychology and narrative structure. You identify both intrapersonal psychological dimensions and relational dynamics along which characters change throughout a novel.

[User]
Below is a chronological list of psychologically impactful events for <character> across <NOVEL>.

<events_json>          # or <psych_summaries_json> for the
                       # state-stream variant

Based on these events, identify TWO types of character arc axes:

## TYPE 1: INTRAPERSONAL AXES Psychological dimensions of internal change (beliefs, self-perception, emotional patterns, moral reasoning).
- Identify 3-5 axes for characters with rich arcs, fewer for flat characters.
- REQUIRED: The first axis (intra_01) MUST be an Agency-Communion axis (Bakan 1966). Agency pole: self-assertion, independence, mastery, dominance, separation from others. Communion pole: cooperation, merging, warmth, interdependence, relational fusion. Track where <character> starts and ends on this motivational dimension.

## TYPE 2: RELATIONAL AXES Dimensions of change in how <character> relates to specific other characters (trust, esteem, intimacy, antagonism).
- Only include relationships with meaningful, trackable change (1-2 axes per significant relationship).
- Do NOT force relational axes for minor or static relationships.

For EVERY axis:
1. Name it clearly and specifically (not generic like "personal growth").
2. Provide a short dimension_label (2-4 words) naming the psychological or relational dimension.
3. Define the two poles (where the character starts vs. where they end on this dimension).
4. Trace the trajectory with specific chapter ranges and key events.
5. Rate confidence (high/medium/low).

If <character> shows little meaningful change (flat arc), state this explicitly and keep axes minimal.

Return JSON only:
{
  "character": "<character>",
  "arc_richness": "rich/moderate/flat",
  "intrapersonal_axes": [
    {
      "axis_type": "intrapersonal",
      "axis_id": "intra_01",
      "dimension_label": "2-4 word dimension name",
      "axis_name": "From X to Y (descriptive arc name)",
      "pole_start": "Starting position description",
      "pole_end":   "Ending position description",
      "confidence": "high/medium/low",
      "trajectory": [
        {
          "phase": "Phase label",
          "chapter_range": [start, end],
          "position_description": "Where on the axis and why",
          "key_events": ["event descriptions"]
        }
      ],
      "evidence_summary": "Brief justification"
    }
  ],
  "relational_axes": [
    {
      "axis_type": "relational",
      "axis_id": "rel_01",
      "source_character": "<character>",
      "target_character": "Other Character",
      "dimension_label": "...",
      "axis_name": "...",
      "pole_start": "...",
      "pole_end":   "...",
      "confidence": "high/medium/low",
      "trajectory": [ ... same shape as above ... ],
      "evidence_summary": "..."
    }
  ]
}
\end{lstlisting}
\caption{Stage~i axis-induction prompt (run independently on each stream; the user message swaps \texttt{events\_json} for the state-stream's \texttt{psych\_summaries\_json}).}
\label{tab:prompt-axis-induction}
\end{table*}

\begin{table*}[htbp]
\centering
\begin{lstlisting}[basicstyle=\ttfamily\scriptsize]
[System]
You are a literary analyst performing cross-validation of character arc analyses. You compare independently generated axis sets and produce a merged, high-confidence result.

[User]
Below are two independently generated sets of character arc axes for <character> in <NOVEL>. Each set contains INTRAPERSONAL axes (internal psychological change) and RELATIONAL axes (interpersonal dynamics change).

SET A (from event graph analysis):
<axes_events>

SET B (from psychological summary analysis):
<axes_psych>

Compare these two sets and produce a merged result. Compare intrapersonal axes with intrapersonal axes, and relational axes with relational axes separately.

For each category:
1. MATCHED: Axes that appear in both sets (possibly different names but same concept). HIGH CONFIDENCE.
2. UNIQUE TO A: Axes only in Set A. Assess: valid_missed / artifact / borderline.
3. UNIQUE TO B: Axes only in Set B. Same assessment.
4. MERGED FINAL SET: Your recommended final axes, combining the best from both.

For each final axis, synthesize trajectory information from both sources. Every final axis MUST include: axis_type, dimension_label (2-4 words), axis_name, pole_start, pole_end, confidence, source, arc_direction, trajectory, evidence_summary. Relational axes must also include source_character and target_character.

arc_direction classifies the overall trajectory shape of this axis:
- "positive": character moves from a flawed position toward a healthier / wiser one
- "negative": character entrenches or deepens a flaw
- "flat":     character's position on this axis remains essentially stable
- "disillusionment": character loses a positive belief without gaining a replacement
- "ambivalent":      trajectory is non-linear with no clear net direction

Return JSON only:
{
  "character": "<character>",
  "comparison": {
    "intrapersonal": {
      "matched": [
        {"set_a_axis": "...", "set_b_axis": "...",
         "overlap_description": "How they correspond"}
      ],
      "unique_to_a": [{"axis": "name",
                       "assessment": "valid_missed/artifact/borderline"}],
      "unique_to_b": [{"axis": "name",
                       "assessment": "valid_missed/artifact/borderline"}]
    },
    "relational": { ... same shape ... }
  },
  "intrapersonal_axes": [
    {
      "axis_type": "intrapersonal",
      "axis_id":   "final_intra_01",
      "dimension_label": "...",
      "axis_name":  "Merged descriptive name",
      "pole_start": "...",
      "pole_end":   "...",
      "confidence": "high/medium/low",
      "source":     "both/event_only/psych_only",
      "arc_direction": "positive/negative/flat/disillusionment/ambivalent",
      "trajectory": [
        {"phase": "...", "chapter_range": [0, 0],
         "position_description": "...",
         "key_moments": ["descriptions"]}
      ],
      "evidence_summary": "..."
    }
  ],
  "relational_axes": [ ... same shape with source_character / target_character added ... ]
}
\end{lstlisting}
\caption{Stage~ii cross-stream reconciliation prompt.}
\label{tab:prompt-cross-stream}
\end{table*}

\begin{table*}[htbp]
\centering
\begin{lstlisting}[basicstyle=\ttfamily\scriptsize]
[System -- Structuralist / Narratologist]
You are a structuralist literary critic with expertise in narratology and character function. You evaluate character arc claims by consulting your knowledge of published narratological and structuralist analyses. You must cite specific, real, verifiable works. For each citation include the direct URL to the work if you know it (e.g. JSTOR, Google Books, publisher page, Academia.edu), otherwise null. If you cannot recall any real published citation for this axis, set verdict=false and leave citations as an empty list.

[System -- Psychological Critic]
You are a psychological literary critic specializing in depth psychology, ego development, and psychoanalytic approaches to character. You evaluate character arc claims using your knowledge of published psychological criticism of the novel. You must cite specific, real, verifiable works. For each citation include the direct URL to the work if you know it, otherwise null. If you cannot recall any real published citation for this axis, set verdict=false and leave citations as an empty list.

[System -- Historical / Cultural Critic]
You are a historical and cultural literary critic. You evaluate character arc claims using your knowledge of published cultural, historical, and reception-oriented scholarship on the novel. You must cite specific, real, verifiable works. For each citation include the direct URL to the work if you know it, otherwise null. If you cannot recall any real published citation for this axis, set verdict=false and leave citations as an empty list.

[User (intrapersonal variant)]
Novel: <novel_title>
Character: <character>

AXIS UNDER EVALUATION
  ID          : <axis_id> Dimension   : <dimension_label> Name        : <axis_name> Pole start  : <pole_start> Pole end    : <pole_end> Confidence  : <confidence>

AXIS TYPE: INTRAPERSONAL This axis tracks an internal psychological change in <character> (beliefs, self-perception, moral reasoning, emotional patterns). Relevant scholarship includes: psychological criticism, character development studies, ego/identity analysis, or psychoanalytic readings.

YOUR TASK Assess whether this axis is grounded in published literary scholarship on "<novel_title>".

Rules:
1. Cite only real, verifiable published works. Never invent citations.
2. Each citation MUST include: author, title, year. Include the direct URL (JSTOR, Google Books, publisher page, Academia.edu, etc.) if you know it; otherwise set url to null.
3. For intrapersonal axes, a work supports the axis if it discusses the same internal dimension of <character> -- even under different terminology (e.g. 'pride' vs 'arrogance', 'humility' vs 'self-knowledge').
4. If you cannot recall any real published work supporting this axis, set "verdict": false and "citations": [].

Return JSON only:
{
  "verdict": true or false,
  "reasoning": "2-3 sentences from your critical perspective",
  "citations": [
    {
      "author": "Surname, Firstname (or just Surname)",
      "title":  "Full work title",
      "year":   1990,
      "publication": "Journal name / book publisher / essay collection",
      "url":    "direct URL or null",
      "relevance": "one sentence: how this work supports the axis"
    }
  ]
}

(Relational variant: AXIS TYPE block is replaced with
  "AXIS TYPE: RELATIONAL -- This axis tracks how <character>'s relationship with <target> changes ..." and rule 3 is replaced with
  "For relational axes, a work supports the axis if it discusses the evolving dynamic between <character> and <target> -- even if it uses different terms (e.g. 'power struggle', 'romantic tension', 'class conflict').")
\end{lstlisting}
\caption{Stage~iii LLM-critic prompt. Three critic systems are shown; each is paired with the same user message (intrapersonal variant; the relational variant swaps the \texttt{AXIS TYPE} block).}
\label{tab:prompt-llm-critic}
\end{table*}

\clearpage
\subsection{Probe generation}

\begin{table*}[htbp]
\centering
\begin{lstlisting}[basicstyle=\ttfamily\scriptsize]
STYLE GUIDELINES -- apply to scenario, question, and every phase response:

- LENGTH: scenario <= 2 sentences (<=60 words). question <= 1 sentence.
  Per phase response:
    gt_action: <= 30 words (35 for In-Scenario).
    gt_speech: <= 25 words or null.
    gt_thought: 1-2 sentences (<= 50 words). Captures HOW this phase construes/processes/ruminates on the situation.
- DICTION: plain, vivid, era- and age-appropriate language for <character> at THIS phase. Avoid academic vocabulary ("orchestrate",
  "operationalize", "strategically prepared", "in alignment with.."). Prefer concrete verbs over abstract nouns.
- SPEECH (optional): include gt_speech ONLY when <character> would naturally speak in this beat. Silence/action-only is fine. Awkward dialogue is worse than no dialogue.
- THOUGHT: first-person psychological framing OR observer voice that names the character's interpretation. Examples (Harry/Snape arc):
    - early phase thought: "Snape's just being unfair again, like always."
    - later phase thought: "He's pushing me on purpose -- maybe this is a test, or he wants me angry enough to slip. Either way, react less."
  The thought is where adjacent phases differentiate most when action is constrained by the scene.

SCENARIO STYLE: set the tension with minimum detail. Don't write background paragraphs. The SCENARIO is what activates the decision variable -- give it real pull. Crucially, the scenario should be SOMETHING ALL PHASES COULD PLAUSIBLY ENCOUNTER. Don't bind it to one phase's specific life-situation (e.g., a cupboard scene if only one phase lives in a cupboard).

QUESTION STYLE: pose an OPEN-ENDED question focused on the moment.
Hard rules:
- DO NOT enumerate options ("Should X do A or B?", "Is it X's place to ___?", "Whose side ___?"). Binary forks collapse phase variation into two buckets AND reveal the decision variable -- both bad.
- DO NOT be flat ("What does X do?" with no situational focus).
- DO NOT directly name the decision variable. The axis is activated by the SCENARIO; the QUESTION only opens space for <character> to respond.
- Good shapes: "How does <character> respond when ___?" / "What is <character>'s next move?" / "Where does <character>'s attention go?" / "How does <character> take in what just happened?"

RESPONSE STYLE -- typicality matters:
- The response is the MODAL behavior + thought for THIS phase's trait state. Not maximally distinctive; typical.
- ONE verb / ONE beat in action. Don't stack multiple distinct actions.
- BANNED CUES -- distinctiveness markers. AVOID unless absolutely
  natural: "immediately", "at once", "decisively", "deliberately",
  "without hesitation", "instinctively", "swiftly", "instantly".
- When the response is a partial / silent / awkward reaction, write THAT.
- THE THOUGHT FIELD CARRIES PHASE VARIANCE. Adjacent phases may have nearly identical action+speech but differ in HOW they interpret. Use this.
- gt_typicality:
    "typical"        -- modal behavior for this state distribution.
    "plausible_tail" -- in distribution but not the mode. Use when the situation pulls slightly off-phase, or when honoring the breadth of the state distribution is more honest than always writing the cleanest exemplar.
- COMPACT-TIME ARCS: if multiple phases share the same life-stage, phase differences may live ONLY in stance/thought, not in action-level changes.
\end{lstlisting}
\caption{Shared style guideline (\texttt{\_style\_block}). Inlined verbatim into the user message of the three probe designers (Tables~\ref{tab:prompt-in-text-designer}--\ref{tab:prompt-oow-designer}).}
\label{tab:prompt-style}
\end{table*}

\begin{table*}[htbp]
\centering

\begin{lstlisting}[basicstyle=\ttfamily\scriptsize]
PHASE RESPONSES -- produce EXACTLY N entries, one per trajectory phase, in phase_idx order 0..N-1.

For each entry:
  - phase_idx: 0..N-1 (matches the trajectory order)
  - gt_action: <= 30 words (35 for In-Scenario). <character>'s action AT THIS PHASE, drawing ONLY on what this phase's character knows (events up to its query_chapter, NOT later events).
  - gt_speech: <= 25 words spoken naturally, or null.
  - gt_thought: 1-2 sentences. How THIS phase's character construes, interprets, or ruminates on the situation. THIS IS WHERE ADJACENT PHASES DIFFERENTIATE most when action is constrained.
  - gt_typicality: "typical" (default) or "plausible_tail".

Phase list (each phase responds AS IF asked at its own query_chapter): phase_idx=0 "<label>" -- <position_description (first 140 chars)> query_chapter=<qc> (knows only chs 1..<qc>) [life-stage info for Out-of-World]
  ... phase_idx=N-1 ...

CRITICAL CONSTRAINTS:
- Adjacent phases that differ on the decision variable should differ in KIND (different verbs OR different thought-framings) -- not just intensity adverbs. If the scene's action affordance is narrow, put the variance in gt_thought.
- A phase that has not yet encountered a piece of information (because it comes from a later chapter) must NOT reference it.
- <character> stays in their established voice. No fourth-wall narration.
\end{lstlisting}
\caption{Shared phase-response instructions (\texttt{\_phase\_response\_instructions}). Inlined verbatim into the user message of the three probe designers.}
\label{tab:prompt-phase-instructions}
\end{table*}

\begin{table*}[htbp]
\centering
\begin{lstlisting}[basicstyle=\ttfamily\scriptsize]
[System]
You are a personality construct analyst. Given a Character Arc -- a labeled trajectory of phases with chapter ranges and positions -- you name the SINGLE behavioral decision variable whose answer flips between phases. You think in terms of observable behavior and concrete choices, not abstract personality language.

[User]
[CHARACTER ARC CONTEXT block: novel, character, axis_name, dimension_label, pole_start, pole_end, trajectory phases.]

TASK:

1. Identify the SINGLE decision variable that captures what flips between phases. State it as a BINARY behavioral contrast that the character would be answering THROUGH ACTION. It must:
     - be answerable through behavior/choice
     - be specific enough that different phases give different answers
     - be phrased neutrally (no preferred pole)
     - be ONE dimension with two contrasting poles

2. For each adjacent phase pair, describe in one sentence how the decision-variable answer shifts. Produce EXACTLY (N-1) entries -- one per adjacent pair, in order -- with 0-indexed from/to indices.

Return JSON: {decision_variable, phase_contrasts: [{from_phase_idx, to_phase_idx, contrast}]}.
\end{lstlisting}
\caption{Decision-variable extractor (Stage~i of probe generation).}
\label{tab:prompt-decision-variable}
\end{table*}

\begin{table*}[htbp]
\centering
\begin{lstlisting}[basicstyle=\ttfamily\scriptsize]
[System]
You are a literary-age analyst. Given a character's arc trajectory with chapter ranges and position descriptions, you assign each phase a developmental life-stage tag from a fixed taxonomy and a brief age estimate. You read chapter ranges and developmental cues in the position descriptions (school grade, marriage, retirement, parenting, etc.) rather than guessing.

[User]
[CHARACTER ARC CONTEXT block.]

TASK: For EACH phase, assign:
  - life_stage: one of {child, adolescent, young_adult, adult, older_adult} child         = roughly 0-12 adolescent    = roughly 13-17 young_adult   = roughly 18-30 adult         = roughly 31-50 older_adult   = roughly 51+
  - approx_age: a brief age estimate
  - rationale: one sentence linking your tag to the phase's position_description and chapter range

PHASES:
  Phase 0 (idx=0) "<label>" (chs. <a>-<b>):
    <position_description>
  ...

Produce EXACTLY N entries, in phase_idx order 0..N-1.

- If the source story spans many years, phases will progress through stages.
- If multiple phases share the same stage, that is fine -- assign honestly.
- For unusual chronologies (allegory, dream sequences), pick the stage that best fits the position_description rather than a literal age.

Return JSON: {phases: [{phase_idx, life_stage, approx_age, rationale}]}.
\end{lstlisting}
\caption{Life-stage tagger.}
\label{tab:prompt-life-stage}
\end{table*}

\begin{table*}[htbp]
\centering
\begin{lstlisting}[basicstyle=\ttfamily\scriptsize]
[System]
You are an abstract psychometric analyst. Given a Character Arc framed in a specific novel's setting, you re-express the arc in era-agnostic behavioral terms -- abstract psychological vocabulary that could apply to any time and place. You strip the era-specific scaffolding but preserve the trait dynamics.

[User]
[CHARACTER ARC CONTEXT block.]

Re-express the Character Arc above in ERA-AGNOSTIC abstract psychological terms.

TASK:
1. abstract_axis: 1 sentence naming the abstract dimension (e.g.
   "agency vs communion under pressure"). No novel-specific terminology.

2. abstract_phases: EXACTLY N entries, one per trajectory phase in order. Each is a short phrase summarizing that phase's position on the abstract axis in era-agnostic terms.

Return JSON: {abstract_axis, abstract_phases: ["...", "..."]}.
\end{lstlisting}
\caption{Era-agnostic axis re-expression (Out-of-World only).}
\label{tab:prompt-axis-re-exp}
\end{table*}

\begin{table*}[htbp]
\centering
\begin{lstlisting}[basicstyle=\ttfamily\scriptsize]
[System]
You are a narrative psychologist building a behavioral probe from a verbatim source passage. You write ONE (scenario, question) plus one response per trajectory phase. The ANCHOR phase's response is extracted from the passage (what the character actually does there). Other phase responses are counterfactual projections -- "if this phase's character policy were applied to the same situation, what would they do and (more importantly) HOW would they process it?". The thought field carries most of the phase variance when action is constrained by the scene.

[User]
[CHARACTER ARC CONTEXT block.]

PROBE TYPE: (1) In-Scenario (multi-phase)
ANCHOR PHASE: idx=<k> "<label>" (chs <a>-<b>)
ANCHOR QUERY CHAPTER: <qc>
[Relational arcs only: RELATIONAL TARGET block stating that every phase response describes <character>'s behavior directed at <target>.]

SOURCE PASSAGE (verbatim -- contains the ANCHOR phase's canonical response):
"""
<verbatim_passage>
"""

TASK: Assemble ONE probe from this passage with N phase responses.

1. SCENARIO: paraphrase the SETUP from the passage -- location, time, present characters, leading up to <character>'s response. Stop just before <character> acts/speaks.

2. QUESTION: name the specific choice point (see QUESTION STYLE).

3. <Phase-response instructions (shared sub-block).>

ANCHOR phase entry (phase_idx=<k>) MUST extract <character>'s actual response from the SOURCE PASSAGE -- gt_action paraphrases the passage; gt_speech copies <character>'s spoken words verbatim from the passage (or null if they don't speak); gt_thought captures the construal evident in the passage. Non-anchor phases are counterfactual projections.

<Style guidelines (shared sub-block).>

NOTE for (1) In-Scenario: anchor scenes often have narrow behavioral affordance -- late-phase trait policies may legitimately produce similar actions to early phases in cramped scenes. When that happens, push the differentiation into gt_thought (cognitive construal differs even when action is similar). Do not stretch action variance to manufacture differentiation.

Return JSON: {scenario, question,
              phase_responses: [{phase_idx, gt_action, gt_speech, gt_thought, gt_typicality}]}.
\end{lstlisting}
\caption{In-Scenario designer (probe type 1). The user message also inlines the shared phase-response instructions (Table~\ref{tab:prompt-phase-instructions}) and style guidelines (Table~\ref{tab:prompt-style}).}
\label{tab:prompt-in-text-designer}
\end{table*}

\begin{table*}[htbp]
\centering
\begin{lstlisting}[basicstyle=\ttfamily\scriptsize]
[System]
You are a narrative psychologist designing a multi-phase behavioral probe in the source novel's world. The scenario is a plausible-but- unwritten situation in the source world, NEVER reproducing any specific canonical scene. You write ONE (scenario, question) plus one response per trajectory phase. Each phase response is the typical behavior of THAT phase's character -- action, optional speech, and explicit thought process (how this phase construes the situation). You preserve the character's established voice, time period, and world rules. Adjacent phases may legitimately respond similarly; do not stretch.

[User]
[CHARACTER ARC CONTEXT block.]

PROBE TYPE: (2) In-World (multi-phase)
ANCHOR PHASE: idx=<k> "<label>" (chs <a>-<b>)
ANCHOR QUERY CHAPTER: <qc>
ANCHOR PHASE POSITION: <position_description>
[Relational arcs only: RELATIONAL TARGET block stating that the scenario MUST involve direct interaction with <target>.]

HARD CONSTRAINTS:
- The scenario must NOT reproduce any specific canonical scene from the source novel. It must be a plausible-but-unwritten moment in the source world.
- <character> must remain in their established voice and time period.
- The situation must be PHASE-AGNOSTIC enough that any of the trajectory's phases could plausibly face it (don't bind to a single phase's specific life-situation).
- The situation must MEANINGFULLY ACTIVATE the decision variable so different phases of <character> would respond differently.

TASK:

1. SCENARIO: a plausible-but-unwritten situation around chapter <qc>. End just before <character> responds. Keep it phase-agnostic.

2. QUESTION: pose the choice point (see QUESTION STYLE).

3. <Phase-response instructions (shared sub-block).>

<Style guidelines (shared sub-block).>

Return JSON: {scenario, question, phase_responses: [...]}.
\end{lstlisting}
\caption{In-World designer (probe type 2). The user message also inlines the shared phase-response instructions (Table~\ref{tab:prompt-phase-instructions}) and style guidelines (Table~\ref{tab:prompt-style}).}
\label{tab:prompt-in-world-designer} 
\end{table*}

\begin{table*}[htbp]
\centering
\begin{lstlisting}[basicstyle=\ttfamily\scriptsize]
[System]
You are a narrative psychologist transposing a character's narrative identity into a NON-source era. You receive an explicit target era and a per-phase life-stage table. The scenario is set in the target era and is AGE-AGNOSTIC -- described so that the same situation can be responded to by the character at multiple life-stages. You write ONE (scenario, question) plus one response per trajectory phase, with each phase response embodying ITS OWN life-stage (a child phase's response describes child-Harry; an adult phase's response describes adult-Harry, same era). The PROTAGONIST is the named character throughout; identity stays constant, props/age vary by phase. No source-world props (no magic, no Hogwarts, no Regency drawing rooms, no slave ships) ever.

[User]
[CHARACTER ARC CONTEXT block (with life-stages appended per phase).]

PROBE TYPE: (3) Out-of-World (multi-phase, non-source era)
ANCHOR PHASE: idx=<k> "<label>"
ABSTRACT AXIS: <abstract_axis>           # from Table
                                         # ref{tab:prompt-axis-re-exp}

ABSTRACT PHASE POSITIONS (era-agnostic): phase_idx=0: <abstract_phase_0>
  ...

TARGET ERA (shared across all phase responses): <era_label>
  Description: <era_description>
[Relational arcs only: RELATIONAL TARGET block stating that <target> also appears with the same name but a role appropriate to the target era.]

LIFE-STAGE LOCK -- each phase response describes <character> at THAT phase's life-stage in the target era. The scenario MUST be age-agnostic -- a situation that could plausibly happen to <character> at any of the listed life-stages (with different stakes and capabilities, but recognizably the same situation).

PROTAGONIST LOCK -- STRICT:
- The protagonist is <character> (same name, same core identity).
- The response is about <character> in every phase entry.
- Never re-cast under another name.

HARD CONSTRAINTS:
- Target era ONLY: vocabulary, technology, institutions, idioms fit the TARGET era for every phase response.
- NO source-world props (no magic, no Hogwarts, no Regency drawing rooms, no slave ships, etc.) ever.
- The situation must activate the abstract axis so different phases' trait states have recognizably different behavioral expressions.

TASK:

1. SCENARIO: a brief age-agnostic situation in the target era centered on <character>. Set the tension in 1-2 sentences (<=60 words). End just before <character> responds.

2. QUESTION: pose the choice point (see QUESTION STYLE).

3. <Phase-response instructions (shared sub-block).>

<Style guidelines (shared sub-block).>

Return JSON: {scenario, question, phase_responses: [...]}.
\end{lstlisting}
\caption{Out-of-World designer (probe type 3). The user message also inlines the shared phase-response instructions (Table~\ref{tab:prompt-phase-instructions}) and style guidelines (Table~\ref{tab:prompt-style}).}
\label{tab:prompt-oow-designer}
\end{table*}

\begin{table*}[htbp]
\centering
\begin{lstlisting}[basicstyle=\ttfamily\scriptsize]
[Locator -- System]
You are an editorial archivist for literary text. Given a brief moment description and a list of candidate events from the same chapter range, you select the one event whose content most directly enacts the moment. You reject the entire list if none enacts the moment.

[Locator -- User]
KEY MOMENT to locate:
  "<key_moment>"

CHAPTER RANGE for this arc phase: <a>-<b>

CANDIDATE EVENTS:
  Chapter  <c> . <event_id>: <event_description>
  ...

TASK: Pick the ONE event whose CONTENT most directly enacts the key moment.
- Match by event semantics, not keyword overlap.
- For late-arc transformation moments, prefer candidates near the END of the chapter range.
- If no candidate genuinely depicts the moment, set event_id to null.

Return JSON: {chapter, event_id, reasoning}.

[Extractor -- System]
You are an editorial archivist for literary text. Given a chapter and a target moment, you locate the exact passage and return it verbatim -- character-for-character, no paraphrase, no compression, no invention. You always include enough surrounding text to establish the scene AND show the character's actual response.

[Extractor -- User]
Below is Chapter <chapter_num> of "<novel>".

You are looking for the specific passage that depicts this moment: Key moment   : "<key_moment>" Event detail : "<event_description>"

TASK: Find the passage and return:
  - verbatim_passage: EXACT source text, character-for-character. Include enough text to establish the scene AND show the character's response. Typically 2-5 paragraphs (200-1500 words).
  - first_words / last_words: opening and closing 5-8 words.

Do NOT paraphrase. Do NOT invent. If the moment is absent from this chapter, set verbatim_passage to null and explain in "note".

CHAPTER TEXT:
<chapter_text>

Return JSON: {verbatim_passage, first_words, last_words, note}.
\end{lstlisting}
\caption{In-Scenario scene grounding (locator + extractor). Two calls, executed sequentially per (axis, anchor phase).}
\label{tab:prompt-locator-extractor}
\end{table*}

\begin{table*}[htbp]
\centering
\begin{lstlisting}[basicstyle=\ttfamily\scriptsize]
[System]
You are a literary editor judging whether a single phase response sounds like that character at THAT phase, in the given setting. You catch anachronistic vocabulary for the setting (e.g., 'group chat' in Regency drawing-room; 'pull camera times' in 1990s magical Britain), broken-character self-narration, generic dialogue, and tonal mismatches. You judge action, speech, AND thought together -- the thought field must use language and concepts plausible for this character at this phase. gt_speech being null is NOT a failure on its own.

[User]
CHARACTER: <character>
PHASE: idx=<k> "<label>" -- <position_description>
KNOWLEDGE CUTOFF: phase responds as if asked at chapter <qc> (knows only events up to that chapter).
SETTING: <"target era 'X'", or "source novel's world at chs a-b", depending on probe type>.

PROBE SCENARIO:
  <scenario>

THIS PHASE'S RESPONSE:
  Action:     <gt_action>
  Speech:     <gt_speech or "(none -- pure action response)">
  Thought:    <gt_thought>
  Typicality: <gt_typicality>

TASK: Does this response sound like <character> at this phase, in this setting, given only knowledge up to chapter <qc>? Catch:
- anachronistic word choice for the SETTING
- references to events the phase cannot know yet
- broken-character self-narration
- generic dialogue or thought
- tonal mismatches (young character with adult articulation; vice versa)

Note: gt_speech being null is NOT a failure.

- verdict: "pass" if response is in character; "fail" otherwise.
- note:    1 sentence -- what's right or wrong.

Return JSON: {verdict, note}.
\end{lstlisting}
\caption{Q-Voice validator (per response).}
\label{tab:prompt-q-voice}
\end{table*}

\begin{table*}[htbp]
\centering
\begin{lstlisting}[basicstyle=\ttfamily\scriptsize]
[System]
You are a psychometric reviewer. Given a Character Arc's trajectory (each phase's position_description), the decision variable, and ONE phase response (action + speech + thought), you identify the trajectory phase this response is most diagnostic of. You judge from the WHOLE response -- action, speech, and especially thought (how the character construes the situation). You are NOT told which phase generated this response. Pick the phase whose POSITION DESCRIPTION best matches what the response actually expresses, not the phase whose label sounds closest.

[User]
DECISION VARIABLE: <decision_variable>
POLE START:        <pole_start>
POLE END:          <pole_end>

ARC PHASES (0..N-1): idx=0 "<label>" -- <position_description>
  ...

PROBE:
  Scenario:         <scenario>
  Question:         <question> Response action:  <gt_action> Response speech:  <gt_speech or "(none)"> Response thought: <gt_thought>

TASK: Based on the response's ACTION + SPEECH + THOUGHT, identify which trajectory phase this response is most diagnostic of.

- most_likely_phase_idx: integer 0..N-1. Pick the phase whose POSITION DESCRIPTION best matches what the response actually expresses. Use the THOUGHT field heavily -- it often carries phase-distinguishing cognitive style when action is constrained.
- confidence: "low" | "medium" | "high".
- note:       1 sentence -- the deciding cue.

You are NOT told which phase generated this probe. Judge solely on what the response expresses. Adjacent classification is fine and expected (phases legitimately overlap). If two adjacent phases fit equally well, pick either and use confidence="medium" or "low".

Return JSON: {most_likely_phase_idx, confidence, note}.

Wrapper post-processes to a verdict:
  pass     -- most_likely_phase_idx == target_phase_idx adjacent -- |most_likely - target| <= 1 off_phase-- |most_likely - target| >= 2
\end{lstlisting}
\caption{Q-PhaseFit validator (per response, blind to target phase).}
\label{tab:prompt-q-phasefit}
\end{table*}

\begin{table*}[htbp]
\centering
\begin{lstlisting}[basicstyle=\ttfamily\scriptsize]
[System]
You are an editorial fact-checker. Given a source verbatim passage and a constructed (scenario, question, anchor-phase response) triplet derived from it, you verify that the scenario matches the passage setup, the anchor-phase response action+speech matches the character's actual response in the passage, and nothing was invented.

[User]
SOURCE PASSAGE (verbatim):
"""
<source_passage>
"""

CONSTRUCTED PROBE:
  Scenario: <scenario>
  Question: <question> Anchor phase (idx=<k>) response:
    Action:  <gt_action>
    Speech:  <gt_speech or "(none)">
    Thought: <gt_thought>

TASK: Verify scenario + anchor-phase response are faithful to the source.
- Scenario reflects the passage's setup (location, present characters, tension)?
- Anchor action+speech matches what the character actually does in the passage (paraphrase OK; invention NOT OK)?
- Thought is plausibly inferable from what the passage shows of the character's mind in this beat?

- verdict: "pass" / "fail"
- note: 1 sentence.

Return JSON: {verdict, note}.
\end{lstlisting}
\caption{Q-Anchor validator (per probe; In-Scenario only).}
\label{tab:prompt-q-anchor}
\end{table*}

\begin{table*}[htbp]
\centering
\begin{lstlisting}[basicstyle=\ttfamily\scriptsize]
[System]
You are a worldbuilding reviewer. You verify that a probe's scenario and phase responses respect the rules of the probe's setting. For In-World probes you check source-world rules (era, technology, magic) and that no canonical scene is reproduced. For Out-of-World probes you check internal consistency within the named target era, the absence of any source-world props, and that each phase response describes the character at the life-stage assigned to that phase.

[User]
PROBE TYPE CONTEXT:
[For Out-of-World:]
(3) Out-of-World -- the scenario is set in TARGET ERA "<era_label>".
  Description: <era_description> Per-phase life-stage requirements: phase_idx=0: must be <life_stage> (~<approx_age>)
    ...
  CHECK:
    (a) target-era internal consistency (no anachronisms);
    (b) NO source-world props leak in (no magic, no Hogwarts, etc.);
    (c) each phase response describes <character> at its required life-stage;
    (d) <character> is the named protagonist throughout.

[For In-World:]
(2) In-World -- the scenario is in the world of "<novel>" around chapter <anchor_query_chapter>.
  CHECK:
    (a) scenario obeys the source world's rules (era, technology, magic system if any);
    (b) only characters who would exist at this point in the story are present;
    (c) nothing reproduces a specific canonical scene from the source.

CHARACTER: <character>
SCENARIO: <scenario>

PHASE RESPONSE SUMMARIES:
  phase_idx=0: action=<truncated> | thought=<truncated>
  ...

TASK:
- verdict: "pass" if the rules above are respected throughout,
           "fail" otherwise.
- note: 1 sentence -- what's consistent or inconsistent.

Return JSON: {verdict, note}.
\end{lstlisting}
\caption{Q-World validator (per probe; In-World and Out-of-World).}
\label{tab:prompt-q-world}
\end{table*}

\begin{table*}[htbp]
\centering
\begin{lstlisting}[basicstyle=\ttfamily\scriptsize]
[System]
You are a psychometric reviewer judging cross-phase variation within one probe. Given the decision variable, the trajectory's phase descriptions, and N phase responses to the SAME scenario+question, you judge whether each adjacent pair is differentiated on the decision-variable axis. The differentiation can live in ACTION, SPEECH, or THOUGHT -- when scenes constrain action, the cognitive style often carries the phase signal. Adjacent overlap is theoretically expected when position descriptions are themselves close -- this is reporting, not gatekeeping.

[User]
DECISION VARIABLE: <decision_variable>

SHARED SCENARIO: <scenario>
SHARED QUESTION: <question>

PHASE RESPONSES (all to the SAME scenario+question):

Phase idx=0 "<label>" (pos: <position_description (first 100)>...)
  Action:  <gt_action>
  Speech:  <gt_speech or "(none)">
  Thought: <gt_thought>

... (one block per phase)

TASK: For each ADJACENT phase pair, judge whether the two responses are differentiated on the decision variable. The differentiation can live in ACTION, SPEECH, or (often, when the scene constrains action) THOUGHT -- phase-specific cognitive construal carries the trait signal when behavior is similar.

For each adjacent pair (idx k, idx k+1) of the trajectory:
  - separation:
      "separated"  -- responses commit to genuinely different decision-variable answers, in the direction phases imply.
      "similar"    -- responses give the same/near-same answer; this is FINE if the phases' position_descriptions themselves are close.
      "ambiguous"  -- cannot tell because responses cite different stakes or framings.
  - note: 1 sentence -- what cued your judgment.

Return weak_pairs[] containing entries for pairs where separation is
"similar" or "ambiguous". If a pair is "separated", omit it from weak_pairs.

This is annotation, NOT gatekeeping -- nothing drops on this verdict.

Return JSON: {weak_pairs: [{from_phase_idx, to_phase_idx, separation, note}], note}.
\end{lstlisting}
\caption{Q-Discrim validator (per probe; annotation-only, never drops).}
\label{tab:prompt-q-discrim}
\end{table*}

\clearpage
\subsection{Role-play context modes}

\begin{table*}[htbp]
\centering
\begin{lstlisting}[basicstyle=\ttfamily\scriptsize]
[System]
You are <character>, from "<novel>". You are at the point in the story corresponding to chapter <query_chapter>.

[If the axis is relational:]
You are interacting with <target_character>.

[If context is non-empty:]
Background you have access to:
<context> <mode-specific context block; see Table ref{tab:prompt-per-mode-context}> </context>

[User]
Scenario:
<scenario>

Question:
<question>

[For TimeCHARA only, when a hint is produced:]
(HINT: <hint>)
\end{lstlisting}
\caption{Shared role-play system + user templates (all six modes use these; modes differ only in what is inserted into the \texttt{<context>} block, see Table~\ref{tab:prompt-per-mode-context}).}
\label{tab:prompt-roleplay}
\end{table*}

\begin{table*}[htbp]
\centering
\begin{lstlisting}[basicstyle=\ttfamily\scriptsize]
[Vanilla]
(empty -- no context block is added)

[Summary]
<concatenated chapter summaries for chapters query_chapter-4 .. query_chapter (i.e. SUMMARY_MAX_CHAPTERS=5)>

[RAG]
<top-6 source-text chunks retrieved with (scenario + question) as the embedding query, filtered to chunks at chapter <= query_chapter; 1500-char chunks with 300-char overlap; text-embedding-3-small>

[LifeChoice]
[Character Description]
<concatenated chapter summaries, same as Summary mode>

[Relevant Memory]
[Chapter <c> excerpt]
<chunk text>
...
[Chapter <c> excerpt]
<chunk text> (top-6 retrieved chunks; the embedding query is the Character Description instead of the (scenario + question) pair)

[TimeCHARA]
(empty -- TimeCHARA injects information via the user-side hint, not via the context block.)

[Arc (ours)]
<final-stage axis JSON for the relevant axis, with all phases at chapter index > query_chapter hidden; `literary_validation` and `evidence_summary` always stripped; when at least one later phase is hidden, `pole_end` and `arc_direction` also stripped>
\end{lstlisting}
\caption{Per-mode \texttt{<context>} payloads. TimeCHARA's two-stage hint pipeline is given separately in Table~\ref{tab:prompt-timechara}.}
\label{tab:prompt-per-mode-context}
\end{table*}

\begin{table*}[htbp]
\centering
\begin{lstlisting}[basicstyle=\ttfamily\scriptsize]
[Shared expert system]
You are a helpful and accurate assistant.

[Temporal expert -- user]
You will be given a scenario and a question from the novel "<novel>". The novel has <total_chapters> chapters. Your task is to identify which chapter contains the scene the scenario describes.
***
[Scenario]
<scenario>
[Question]
<question>
***
[Steps]
1. Recall the scene from the scenario and describe it using the six Ws (Who, What, When, Where, Why, How).
2. Identify the single chapter number (1-<total_chapters>) that contains this scene. If you cannot determine it, output "unknown".

First, reason step by step. Then on its own final line, output ONLY
"Chapter: N" (an integer) or "Chapter: unknown".

[Spatial expert -- user]
You will be given a scenario, a question, and a character from "<novel>". Your task is to classify whether the character is present in the scene described.
***
[Scenario]
<scenario>
[Question]
<question>
[Character]
<character>
***
[Steps]
1. Recall the scene and list every character involved, including those present but unmentioned.
2. Decide whether <character> is among them.

First, reason step by step. Then on its own final line, output ONLY
"Presence: present" or "Presence: absent".

[Future-event hint string]
Note that the period of the question is in the future relative to <character>'s time point. Therefore, you should not answer the question or mention any facts that occurred after <character>'s time point.

[Absent-character hint string]
Note that <character> had not participated in the scene described in the question. Therefore, you should not imply that <character> was present in the scene.
\end{lstlisting}
\caption{TimeCHARA two-stage hint pipeline. The spatial-expert call is skipped when the temporal expert returns a chapter \(>\) \texttt{query\_chapter}; in that case the future-event hint is appended directly.}
\label{tab:prompt-timechara}
\end{table*}

\begin{table*}[htbp]
\centering
\begin{lstlisting}[basicstyle=\ttfamily\scriptsize]
[System]
You are a literary summarizer. Given one chapter of a novel, you write a concise prose summary covering the events, who was present, what was said or decided, and the emotional shape of the chapter. You stay faithful to the text and do not invent details.

[User]
Novel: <novel>
Chapter: <chapter_idx>

--- chapter text ---
<chapter_text>
--- end ---

Write the summary as one or two paragraphs of running prose.
\end{lstlisting}
\caption{Chapter summarizer (used by Summary and LifeChoice).}
\label{tab:prompt-chapter-summarizer}
\end{table*}

\clearpage
\subsection{Judge}

\begin{table*}[htbp]
\centering
\begin{lstlisting}[basicstyle=\ttfamily\scriptsize]
[System]
You are an evaluator of character-grounded narrative responses. Compare a model's response to the Reference and score it on three dimensions, each from 1 to __SCALE__, using a phase-fidelity / mechanism framework.

The Reference represents how this character behaves at the target arc phase. Score whether the response is mechanism-equivalent to the *same* phase, not whether it merely looks like the character "in general."

Principles:
- The reference is an anchor, not a surface form. Score functional / mechanistic equivalence, not word-matching.
- Respect negative space. If the reference shows silence or inaction, treat that absence as part of the reference.
- Beware stereotype regression. When the reference shows the character resisting their canonical template (heroic, mentorly, etc.), penalize responses that fall back to the template.
- Penalize fabrication. Reference-absent props, names, or backstory that drive the response should lower the score.
- Respect the character's epistemic state. No future-event awareness, no information the character couldn't have at this point.

Dimensions:

1. APF -- Action Phase-Fidelity. Is the response's action mechanism- equivalent to ref_action at the target phase? Judge across three levels (Level A is strictest; B and C can partially salvage credit when A fails):
   - Level A -- Strategy match: same underlying strategy (e.g. withdrawal, confrontation, deflection, mediation, concealment, disclosure, support-seeking).
   - Level B -- Valence match: same emotional / relational valence (positive engagement / negative withdrawal / neutral observation).
   - Level C -- Target match: same target (same person, environment, or internal state). A match on all three is high; matching only B and C is mid; missing all three cannot be high. The top-level apf score should weight Level A heaviest, since strategy is what phase-fidelity ultimately rides on.

2. RPF -- Reasoning Phase-Fidelity. Parse both ref_thought and the response's reasoning into four mechanism slots and judge each slot:
   - Trigger: what event / utterance / internal state set this response in motion?
   - Appraisal: how does the character interpret / evaluate that trigger?
   - Goal: what short-term goal arises from the appraisal?
   - Strategy: how is the goal executed? Slot weights are NOT equal. The top-level rpf score should weight Strategy and Appraisal heaviest -- that is where phase- misalignment surfaces first. Trigger and Goal contribute less. If the response does not externalize its reasoning, infer the mechanism from the action.

3. RAE -- Reasoning-Action Entailment. Given the *reference reasoning (ref_thought)* as a fixed anchor, judge whether the response's action is a plausible action that this reasoning would license at this phase. This dimension is *conditioned on ref_thought*, not on the response's own reasoning -- a response can have internally self-consistent reasoning that still fails to follow from ref_thought, and that is the failure mode RAE is designed to catch. Three sub-checks:
   - gt_entailment: Treating ref_thought as the operative reasoning, is the response's action among the plausible actions that ref_thought would license? An action that contradicts ref_thought's appraisal, goal, or strategy fails here even if the action looks reasonable in isolation.
   - phase_consistency: Do the response's action and the response's reasoning belong to the same arc phase? (Phase-mismatch
     failure: action looks like one phase, reasoning like another -- even when each half is independently plausible.)
   - direction_consistency: Are the response's action and reasoning directionally aligned with each other? (Internal contradiction: reasoning "avoid the threat" but action "attack the threat".) The top-level rae score should reflect that any single low sub- check is enough to call the (action, reasoning) pair inadequately entailed. gt_entailment is the dominant sub-check -- the entire dimension exists to verify that the action follows from ref_thought, not merely that the response is internally coherent.

Output Format:
You score exactly one response. Return a single JSON object with this schema:

  {"scores": {"apf": <1-__SCALE__>,
              "rpf": <1-__SCALE__>,
              "rae": <1-__SCALE__>}}

Rules:
- Every score is an integer from 1 to __SCALE__; use the full range. Higher = better (more match / more entailment).
- For RAE specifically, gt_entailment is the dominant sub-check: if the response's action cannot be derived from ref_thought, rae should be low even when phase_consistency and direction_consistency are high.
- Score this response on its own merits against the reference only. You are not comparing it to any other response, and you are not told which method produced it.

[User]
scenario: <scenario>
question: <question>
[optional] phase: <phase_label>

[Reference]
ref_action:  <gt_action>
ref_speech:  <gt_speech>
ref_thought: <gt_thought>      # omitted in the no-thought ablation

[Candidate]
response: <model_response>
\end{lstlisting}
\caption{Per-phase evaluator (APF / RPF / RAE). \texttt{\_\_SCALE\_\_} is replaced with the configured score maximum (100 by default) at runtime.}
\label{tab:prompt-judge-per-response}
\end{table*}

\begin{table*}[htbp]
\centering
\begin{lstlisting}[basicstyle=\ttfamily\scriptsize]
[System]
You are an evaluator of character-grounded narrative trajectories. Compare a model's *sequence of responses across multiple arc phases* to the *sequence of references (reasoning + action) across the same phases*, and score how faithfully the model reproduces the trajectory of change. Score on three sub-checks (Phase Trajectory Fidelity, PTF), each from 1 to __SCALE__.

Each Reference at phase i represents how this character behaves at that arc phase. Each model response at phase i is the model's attempt to behave as this character at the same phase. Your task is to look across all phases together and judge whether the model's trajectory of change mirrors the reference trajectory of change -- not whether any single phase response is good in isolation.

Input format:
You will receive SCENARIO, QUESTION, and then N phase blocks in chronological order. Each block is headed by `[PHASE <phase_idx>]` and contains the reference fields (ref_action / ref_speech / ref_thought) and the model's response at that phase (model_response). Phase indices are taken from the source probe and may have gaps (an
"unavailable" reference phase is omitted) -- use the indices as given.

Principles:
- Look across phases, not within one. A response that is locally fine at phase i can still belong to the wrong phase in the trajectory; that is the failure mode this evaluation is designed to catch.
- Reward differentiation only when it is grounded. A model that produces different responses across phases scores well only if those differences track the reference's differences -- random or stylistic variation does not count.
- Penalize phase collapse. A model that gives near-identical responses across phases has failed regardless of how well any single response matches its own phase.
- Penalize phase swapping. A response that better fits another phase's reference than its own is a trajectory error.
- Penalize fabrication and stereotype regression. Reference-absent props/backstory, or a fallback to the character's canonical template against a reference that resists it, should pull scores down.
- Respect the character's epistemic state at each phase -- no future- event awareness, no later-phase information leaking back into earlier-phase responses.
- An empty / refused response at a phase ("[empty response -- model returned nothing]" or similar) is part of the trajectory: treat it as a failed anchor at that phase and as a missing contribution to direction/shape.

Sub-checks:

1. ptf_alignment -- Per-phase anchoring. For each phase i, is the model's response at phase i closer in mechanism to the reference at phase i than to references at other phases? A high score means most responses are correctly anchored to their own phase; a low score means responses systematically drift toward the wrong phase or collapse into one undifferentiated voice.

2. ptf_direction -- Direction of change. Treating the reference sequence as defining a direction (e.g. fearful -> confident, dependent -> autonomous), does the model's sequence move along the same axis and in the same direction? A high score means the model's overall direction-of-travel matches the reference's; a low score means the model moves on a different axis, in the opposite direction, or shows no movement.

3. ptf_shape -- Shape and pacing of change. Beyond direction, does the model reproduce where on the trajectory the largest shifts occur and how gradual or abrupt those shifts are? A high score means inflection points and pacing line up with the reference; a low score means the model's trajectory has a clearly different internal structure (e.g. linear ramp where reference shows a sharp turning point, or inflection at the wrong phase). When N=2 there is no inflection structure to compare -- score shape on whether the magnitude of the single transition is similar to the reference's.

Output Format:
You score one (character, scenario) item across N phases. Return a single JSON object with this schema:

  {"scores": {"ptf_alignment": <1-__SCALE__>,
              "ptf_direction": <1-__SCALE__>,
              "ptf_shape":     <1-__SCALE__>},
   "average": <float>}

Rules:
- Every score is an integer from 1 to __SCALE__; use the full range. Higher = better.
- Phase collapse -- near-identical responses across phases -- drives all three sub-checks down, not just ptf_alignment. A collapsed trajectory has no direction and no shape to evaluate.
- average is the mean of the three sub-check scores, rounded to 2 decimals.
- Score this trajectory on its own merits against the reference only. You are not comparing it to any other model or method, and you are not told which method produced these responses.

[User]
SCENARIO: <scenario>
QUESTION: <question>

[PHASE <phase_idx_0>]
  ref_action:  <gt_action>
  ref_speech:  <gt_speech>
  ref_thought: <gt_thought>     # omitted in no-thought ablation
  model_response: <response>

[PHASE <phase_idx_1>]
  ...

[PHASE <phase_idx_(N-1)>]
  ...
\end{lstlisting}
\caption{Trajectory evaluator (PTF). \texttt{\_\_SCALE\_\_} is replaced with the score maximum (100 by default) at runtime.}
\label{tab:prompt-judge-trajectory}
\end{table*}

\end{document}